%% file: main.tex
\documentclass[10pt,twocolumn,letterpaper]{article}

\usepackage[pagenumbers]{iccv} 
\usepackage{times}
\usepackage{epsfig}
\usepackage{graphicx}
\usepackage{amsmath}
\usepackage{amssymb}
\usepackage{multirow}
\usepackage{booktabs}
\usepackage{caption}
\usepackage{xcolor}
\definecolor{red}{rgb}{0.9, 0, 0}
\definecolor{deepred}{rgb}{0.6, 0, 0}
\definecolor{blue}{rgb}{0, 0, 0.7}
\usepackage{cuted}
\usepackage[export]{adjustbox}
\usepackage{enumitem}

\usepackage{subcaption}
\usepackage{tabularx}

\definecolor{slightblue}{rgb}{0.21,0.49,0.74}
\usepackage[pagebackref,breaklinks,colorlinks,allcolors=slightblue]{hyperref}

\title{ACMo: Attribute Controllable Motion Generation}

\author{
	\begin{tabular}[h]{ccc}
		Mingjie Wei$^{1}$ \quad\quad Xuemei Xie$^{1\dagger}$ \quad\quad 
		Guangming Shi$^{1,2}$\\
	\end{tabular}\\
	\begin{tabular}[h]{cc}
		$^{1}$Xidian University \quad\quad  $^{2}$Peng Cheng Laboratory\\
	\end{tabular}\\
	{\tt\small mjwei@stu.xidian.edu.cn}, {\tt\small xmxie@mail.xidian.edu.cn}, {\tt\small gmshi@xidian.edu.cn}
}

\begin{document}
\maketitle
\begin{strip}
	\vspace{-40pt}
	\centering
	\includegraphics[width=\textwidth]{fig/ACMo_idea.pdf}
	\captionof{figure}{ACMo handles motions beyond dataset representation, using motion prompts for stylize multimodal generation and multi-attribute control, with LLM Planner mapping zero-shot unseen attributes to dataset texts. Employ rapid fine-tuning to enable the model to recognize new motion patterns. The bracketed text enhances the stability of style and trajectory. Control your motions as you wish!}
	\label{fig:ACMo_idea}
\end{strip}
{
	\renewcommand{\thefootnote}
	{\fnsymbol{footnote}}
	\footnotetext[2]{Corresponding author}
}
\input{sec/0_abstract}
\input{sec/1_introduction}
\input{sec/2_relatedwork}
\input{sec/3_method}
\input{sec/4_experiments}
\input{sec/5_conclusion}
{\small
	\bibliographystyle{ieeenat_fullname}
	\bibliography{RefMain}
}

\end{document}


\maketitle
\tableofcontents 

\section{Visualization}
We have tried different types of visualization methods in this section. The following are the visualizations of our three core contributions:\par
\begin{figure*}[htbp]
	\centering
	\caption{Visual experiment of Reasoning motion generation. The proposed LLM Planner can convert unseen attributes to a dataset-specific text. The BVH production process in Momask\cite{momask} was used for visualization.}
	\label{fig:VisualizationSupRMG}
	\begin{tabular}{*{4}{c}}
		\toprule
		\multicolumn{2}{c}{ADM} & \multicolumn{2}{c}{ADM + LLM Planner} \\
		\midrule
		\multicolumn{2}{c}{A person tiptoes across creaky floorboards} &
		\multicolumn{2}{c}{A person tiptoes carefully} \\
		
		\multicolumn{2}{c}{to avoid waking others.} &
		\multicolumn{2}{c}{and quietly forward.} \\
		
		\multicolumn{2}{c}{\includegraphics[width=0.25\linewidth]{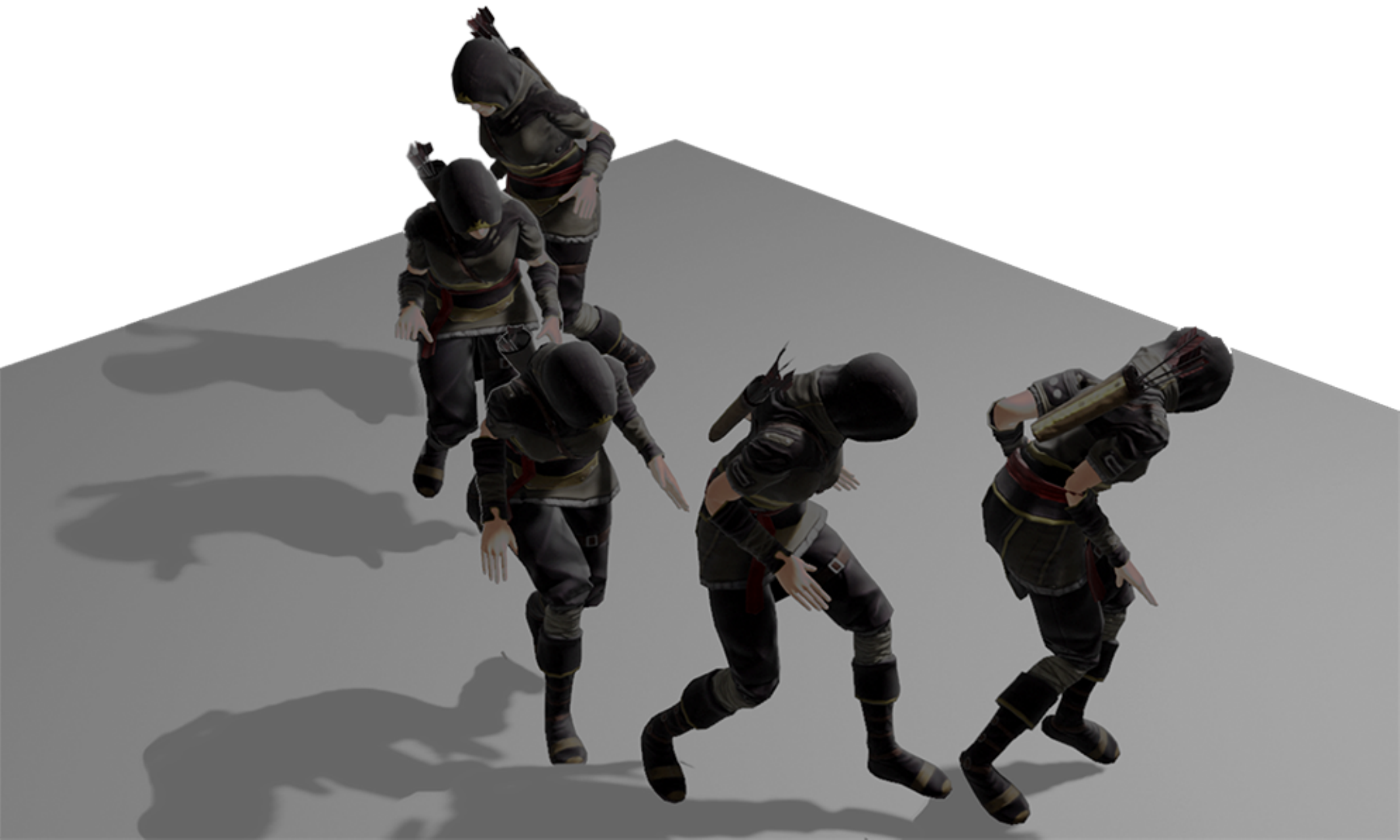}} &
		\multicolumn{2}{c}{\includegraphics[width=0.35\linewidth]{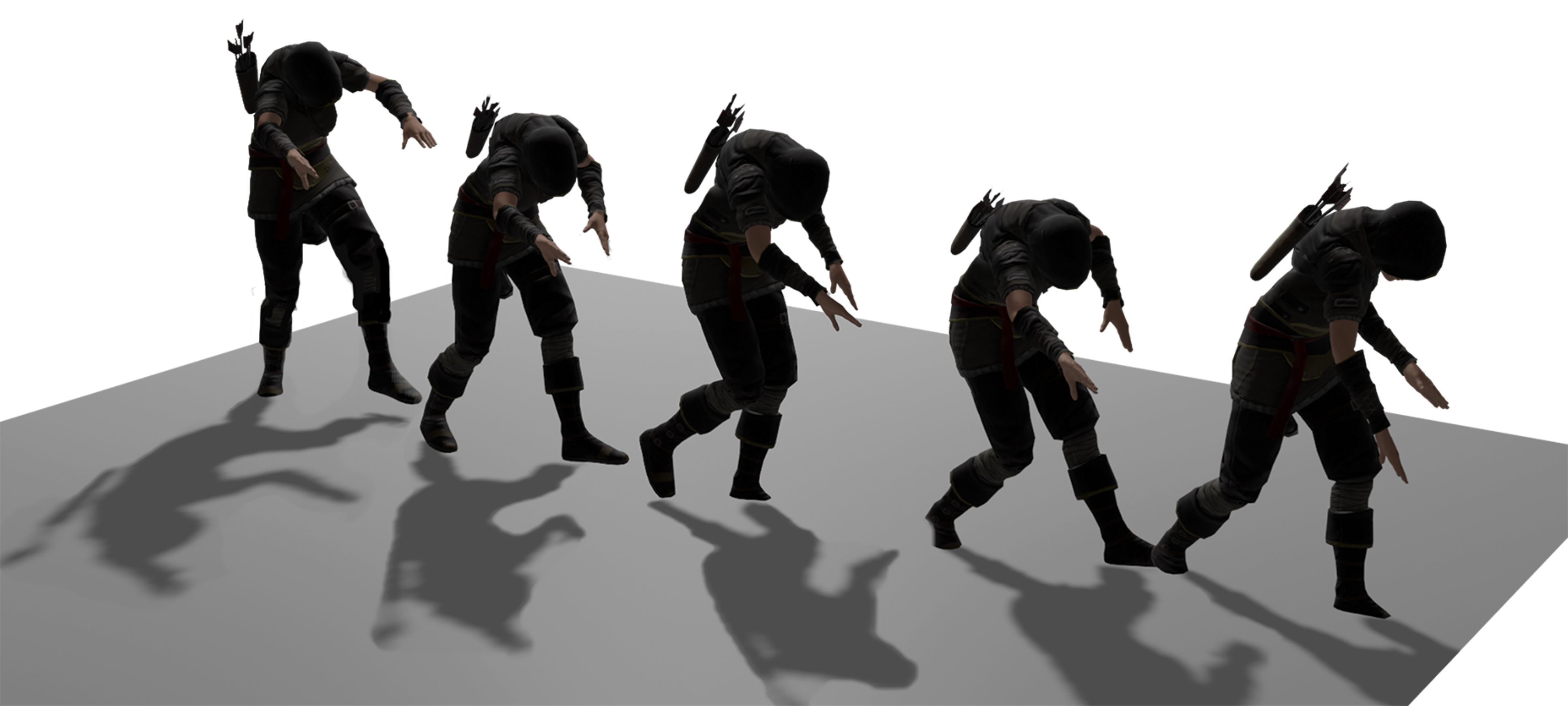}} \\
		\midrule
		ADM & Momask & \multicolumn{2}{c}{ADM + LLM Planner} \\
		\midrule
		\multicolumn{2}{c}{A person jogs through an empty park at sunrise,} &
		\multicolumn{2}{c}{A person jogs happily with swinging arms.}\\
		\multicolumn{2}{c}{enjoying the peaceful atmosphere.} & & \\
		\includegraphics[width=0.24\linewidth]{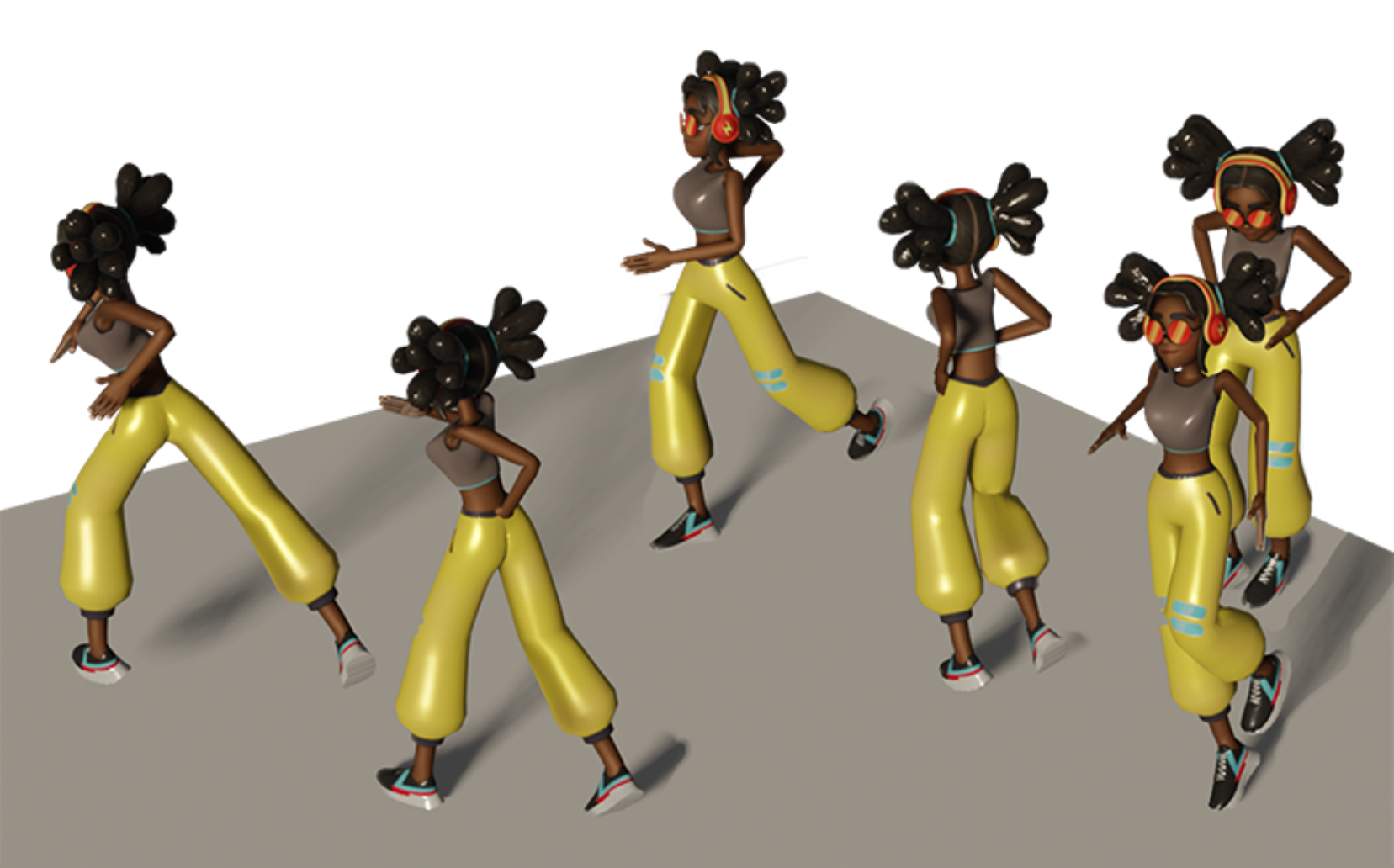}
		& 
		\includegraphics[width=0.18\linewidth]{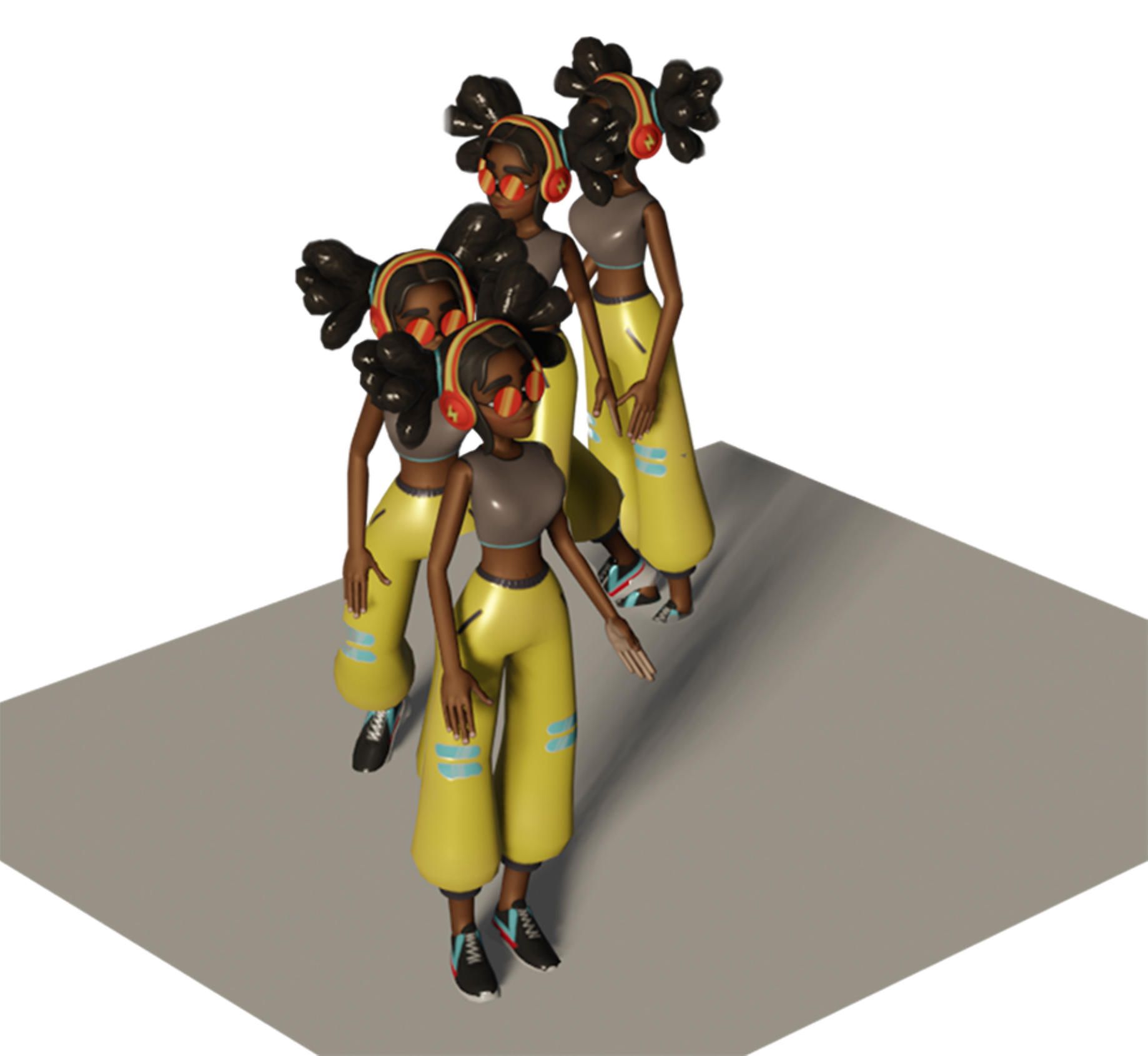}
		&
		\multicolumn{2}{c}{\includegraphics[width=0.31\linewidth]{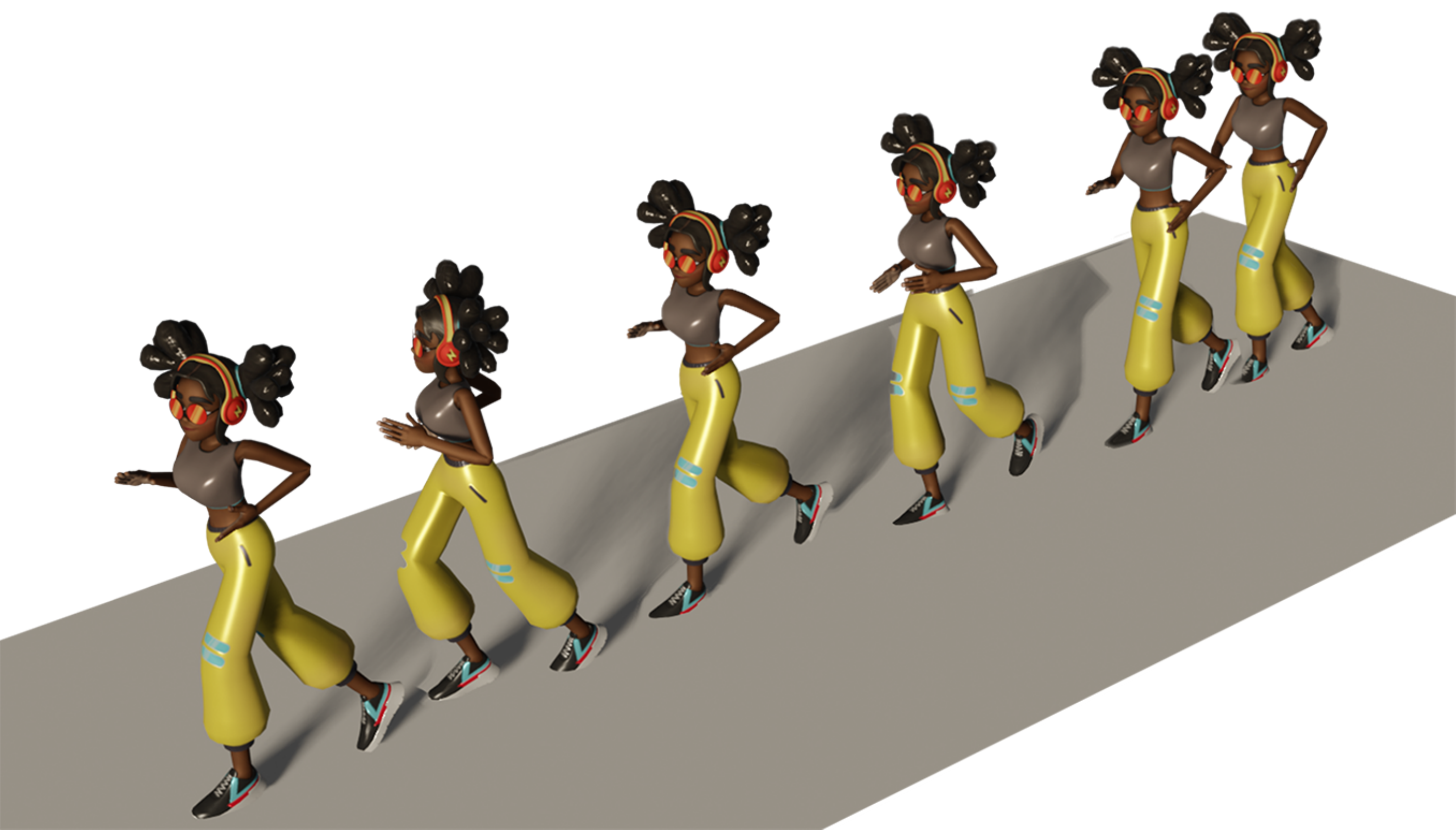}} 
		\\
		\bottomrule
	\end{tabular}
\end{figure*}

{\bf Reasoning motion generation.} We supplemented some reasoning generation visualization with Attribute Diffusion model and LLM Planner in Fig. \ref{fig:VisualizationSupRMG}.\par
\begin{figure*}[htbp]
	\centering
	\caption{Visual experiment of stylized text-to-motion generation.}
	\label{fig:VisualizationSupSTG}
	\begin{tabular}{*{2}{c}}
		\toprule
		Motion Prompt & Ours \\
		\midrule
		Aeroplane & a person is walking with his {\it hands out} in front of him\\
		\includegraphics[width=0.25\linewidth]{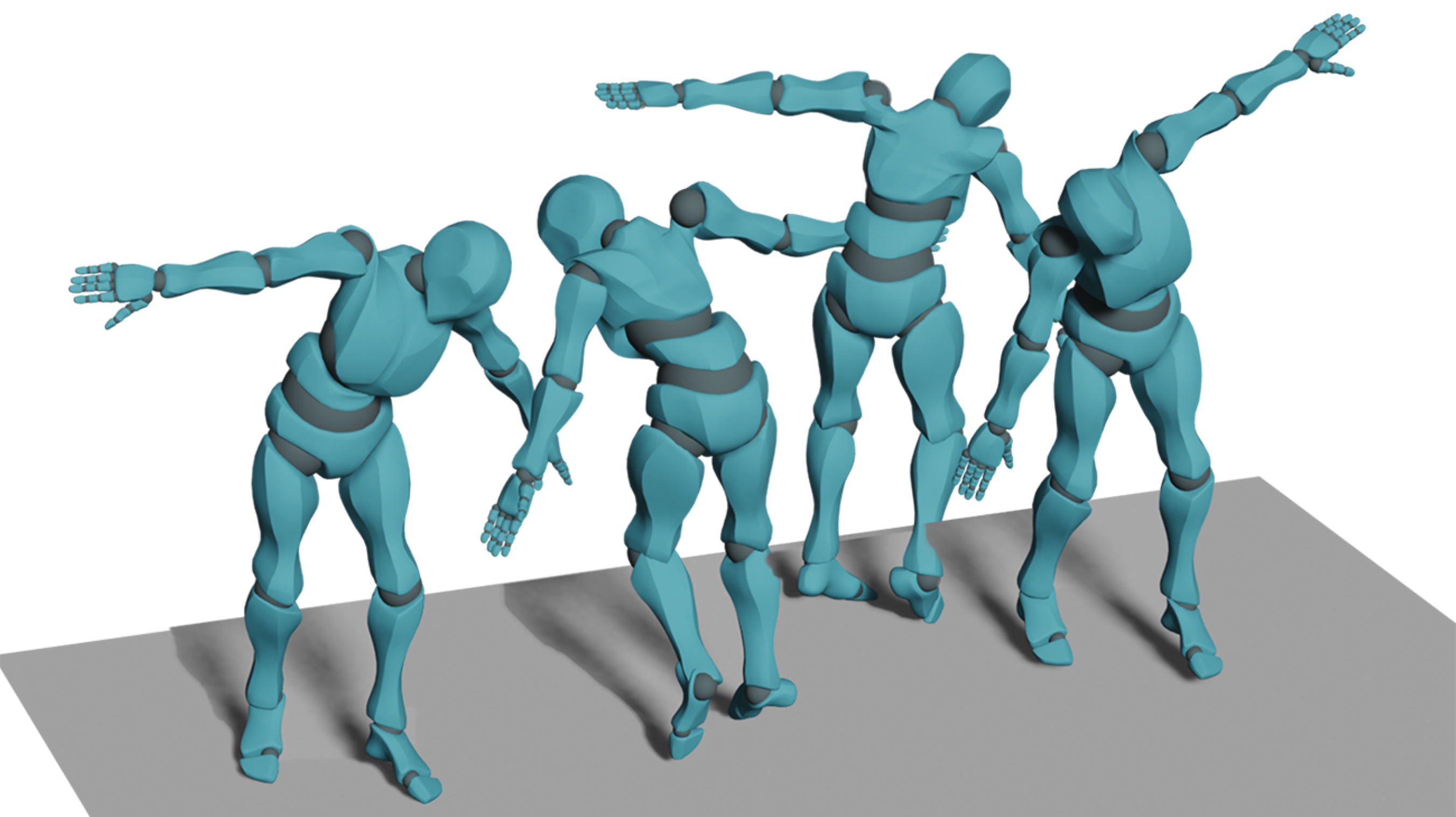} &
		\includegraphics[width=0.4\linewidth]{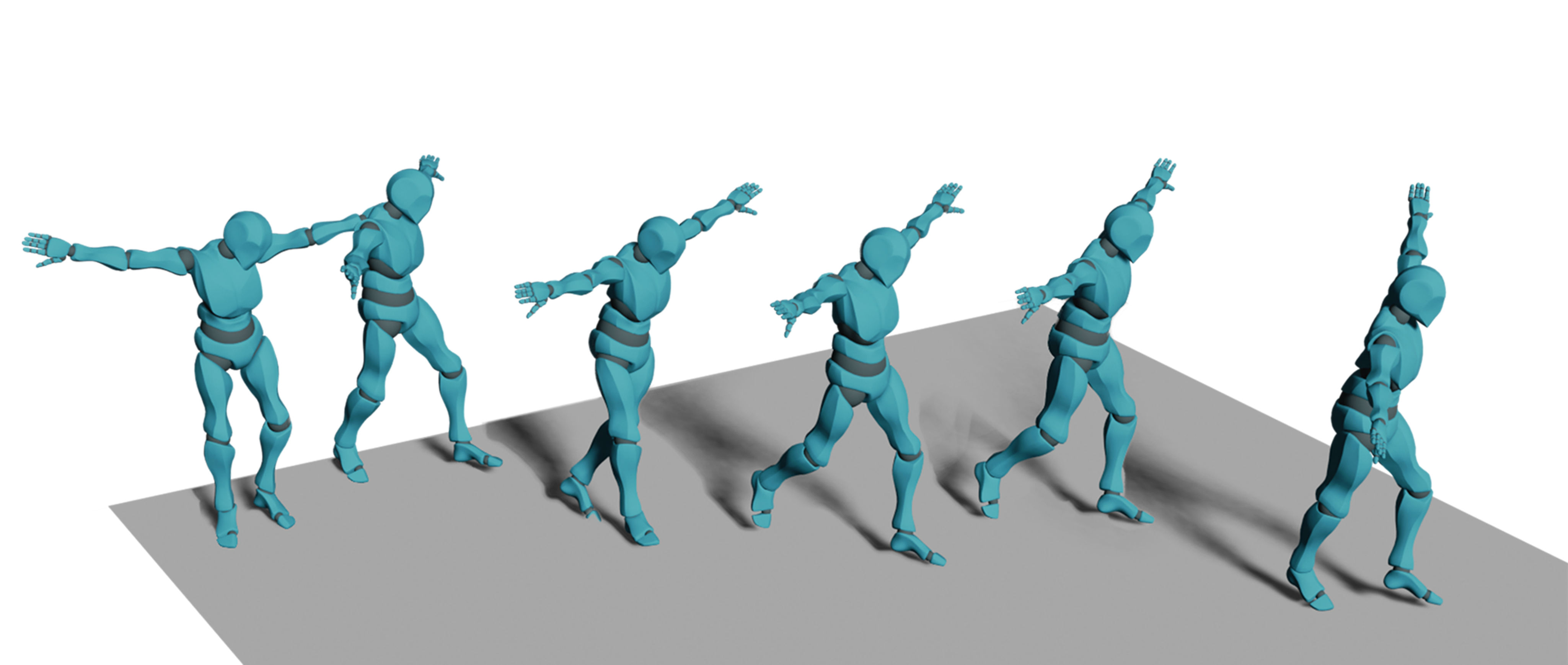} \\
		Chicken & a person {\it bent down walks} around with their arms \\
		& bent backwards, imitating a chicken\\
		\includegraphics[width=0.25\linewidth]{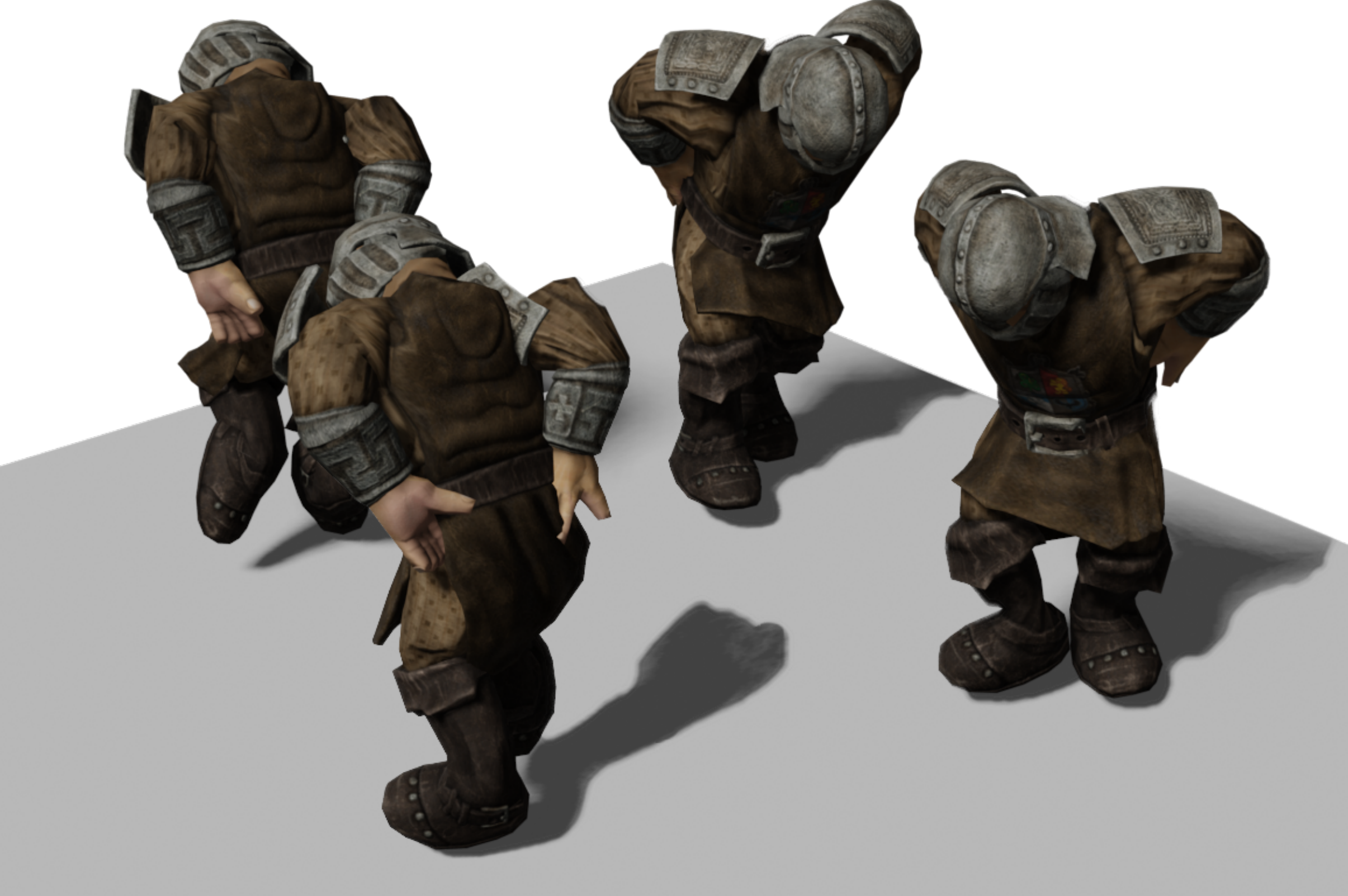} &
		\includegraphics[width=0.3\linewidth]{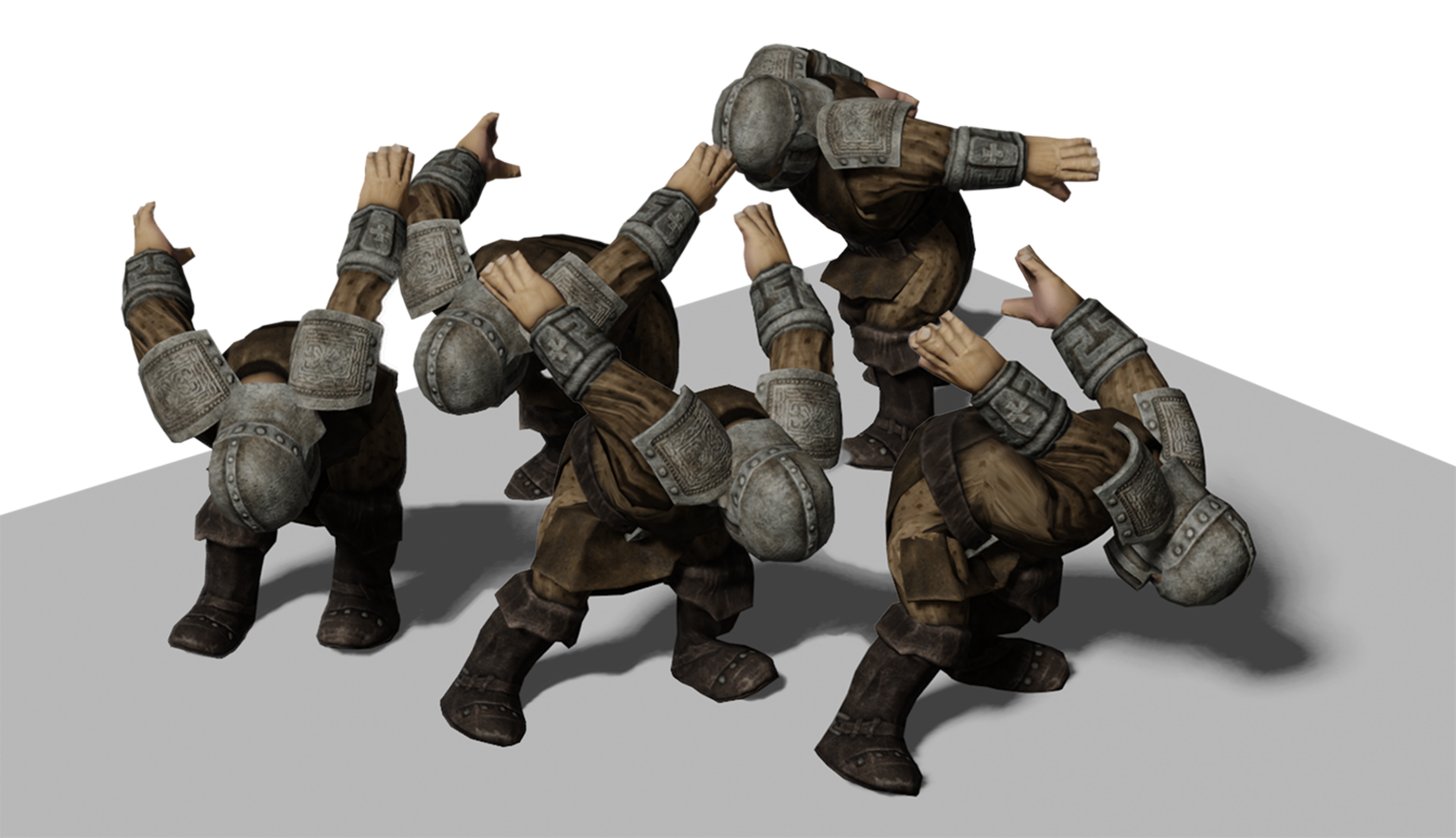} \\
		\bottomrule
	\end{tabular}
\end{figure*}

{\bf Stylized motion generation.} We supplemented some stylized generation visualization via Motion Adapter in Fig. \ref{fig:VisualizationSupSTG}.\par
\begin{figure*}[htb]
	\centering
	\caption{Visual experiment of text-to-motion generation. The proposed Attribute Diffusion Model can achieve word-level granularity generation.}
	\label{fig:VisualizationSupTTM}
	\begin{tabularx}{\textwidth}{>{\centering\arraybackslash}X
			>{\centering\arraybackslash}X
			>{\centering\arraybackslash}X
			>{\centering\arraybackslash}X
		}
		\toprule
		a person squats down a little and then jumps
		&
		a person raises their right hand slightly then sets it back down.
		&
		a man plays a violin with his left hand.
		&
		a person swings a golf club\\
		\includegraphics[width=0.75\linewidth]{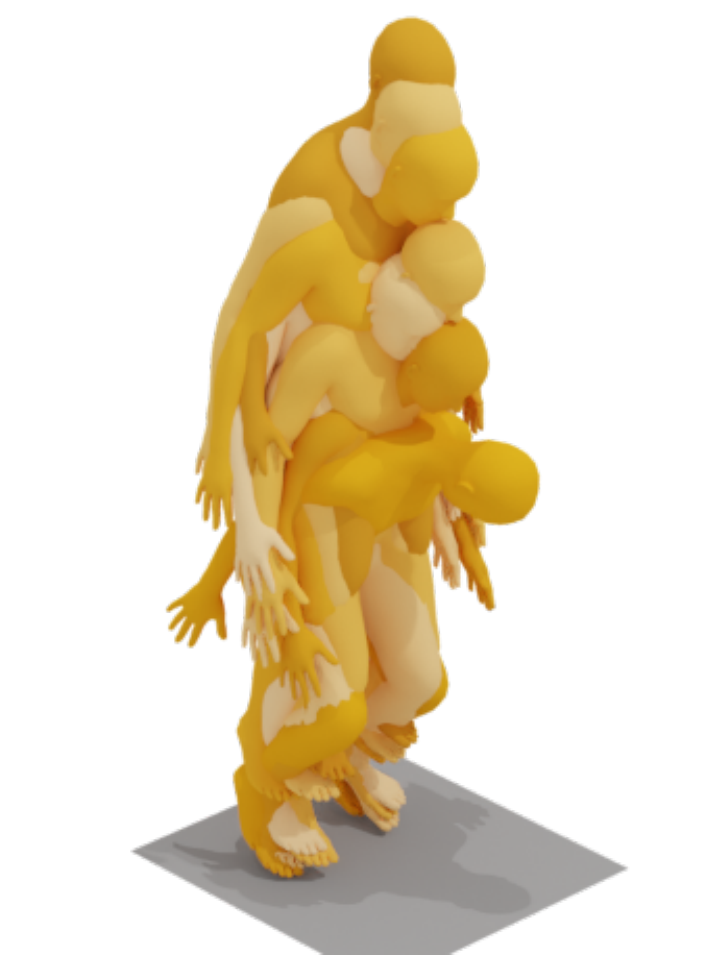} &
		\includegraphics[width=0.6\linewidth]{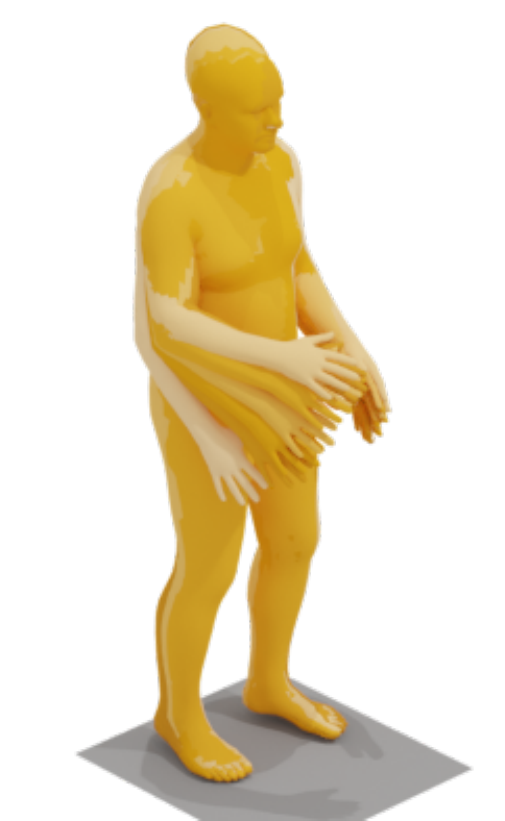} &
		\includegraphics[width=0.65\linewidth]{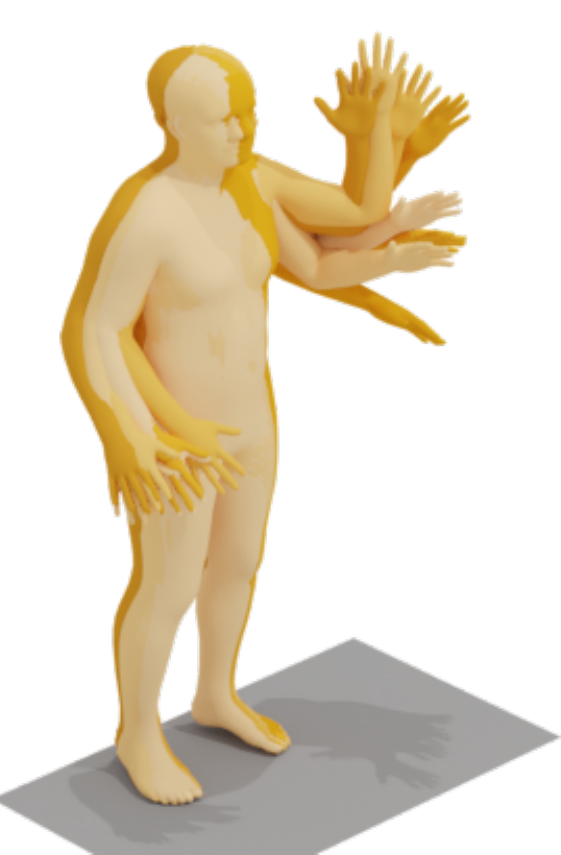} &
		\includegraphics[width=0.87\linewidth]{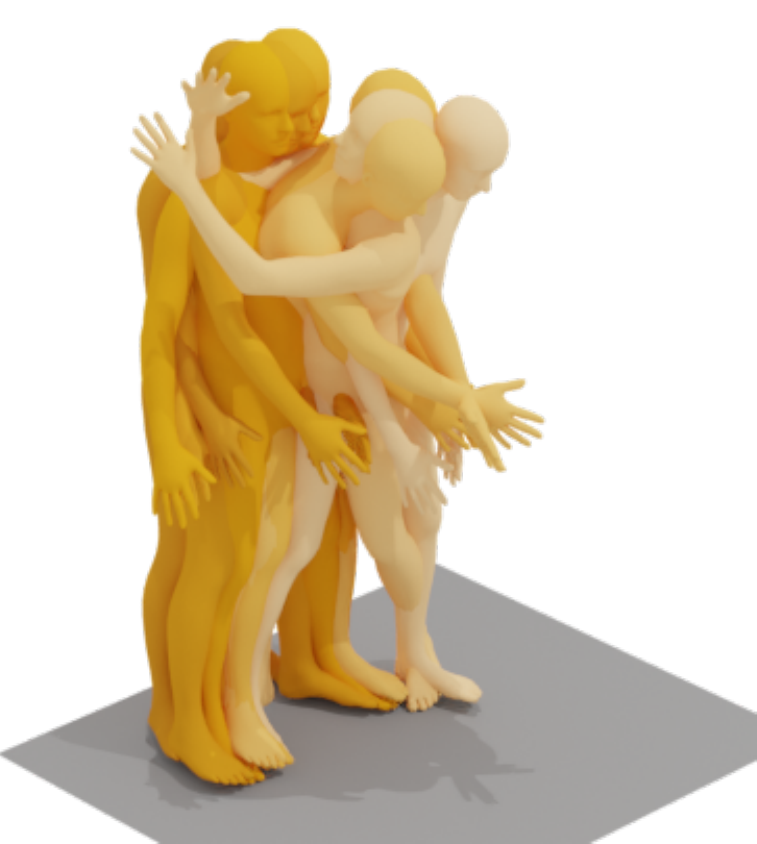}\\
		\hline
		the man is doing jumping jacks
		&
		a person kicking something like in football
		&
		\multicolumn{2}{c}{this person runs lightly forward then stops.} \\
		\includegraphics[width=0.8\linewidth]{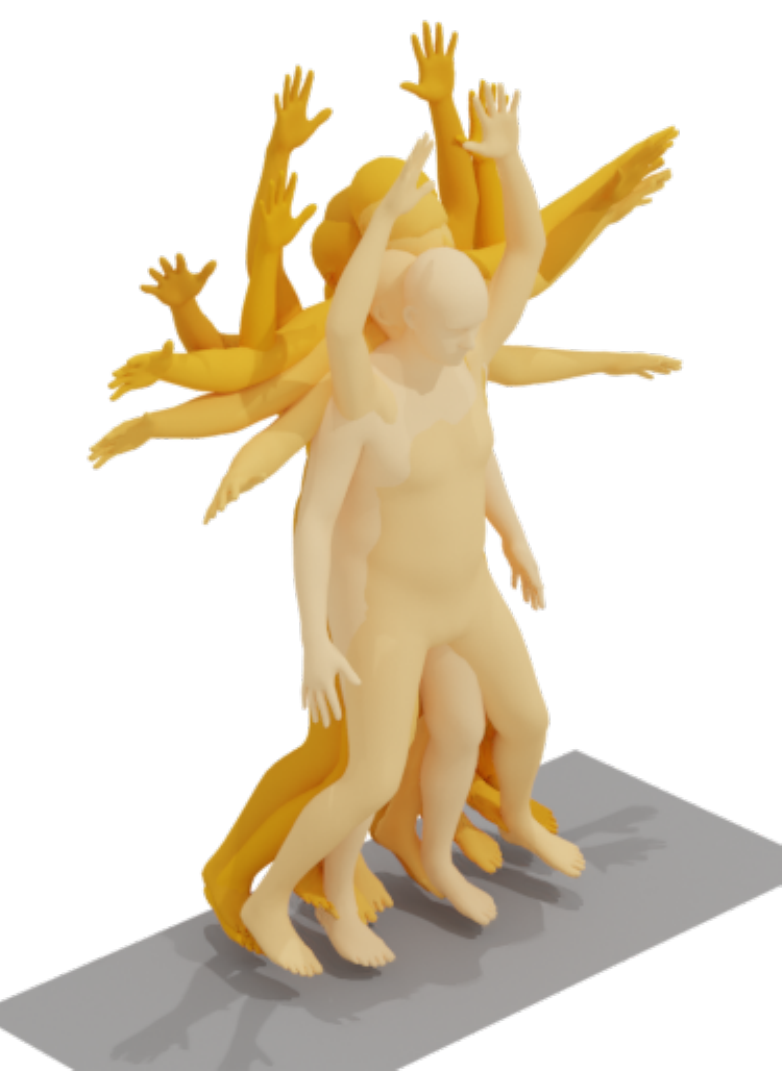} &
		\includegraphics[width=0.9\linewidth]{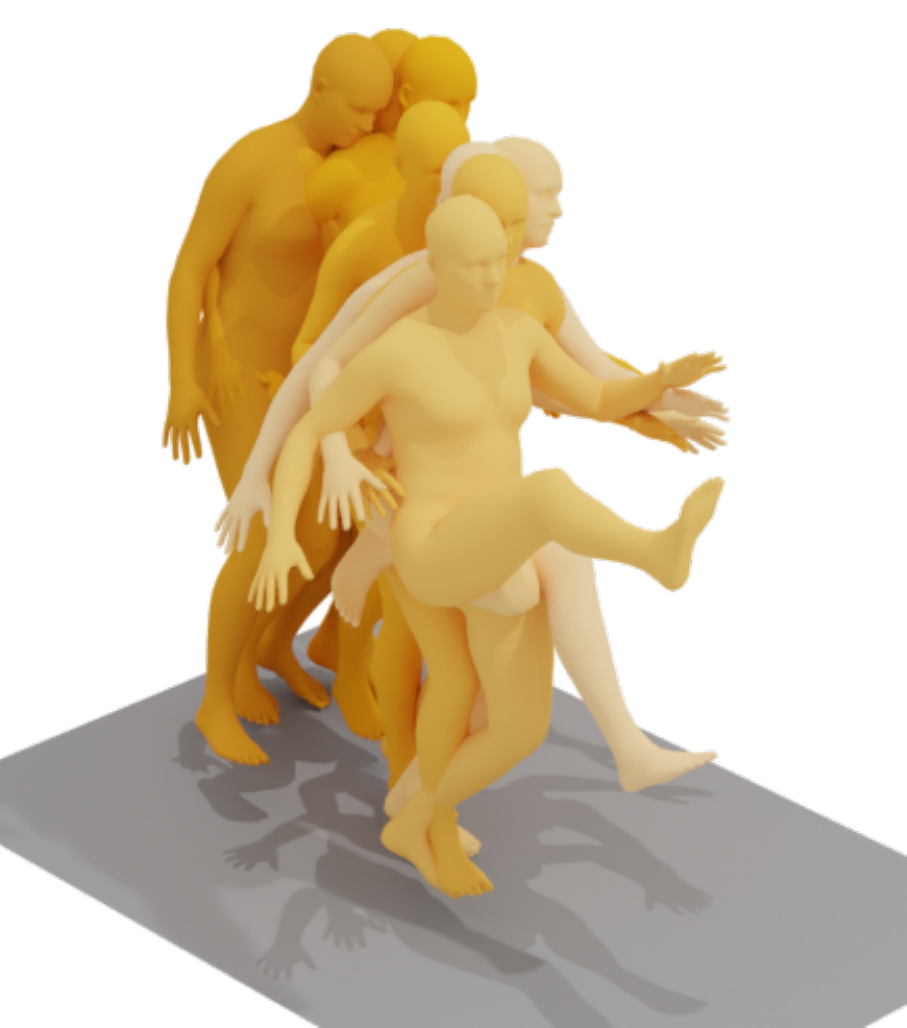} &
		\includegraphics[width=1.8\linewidth]{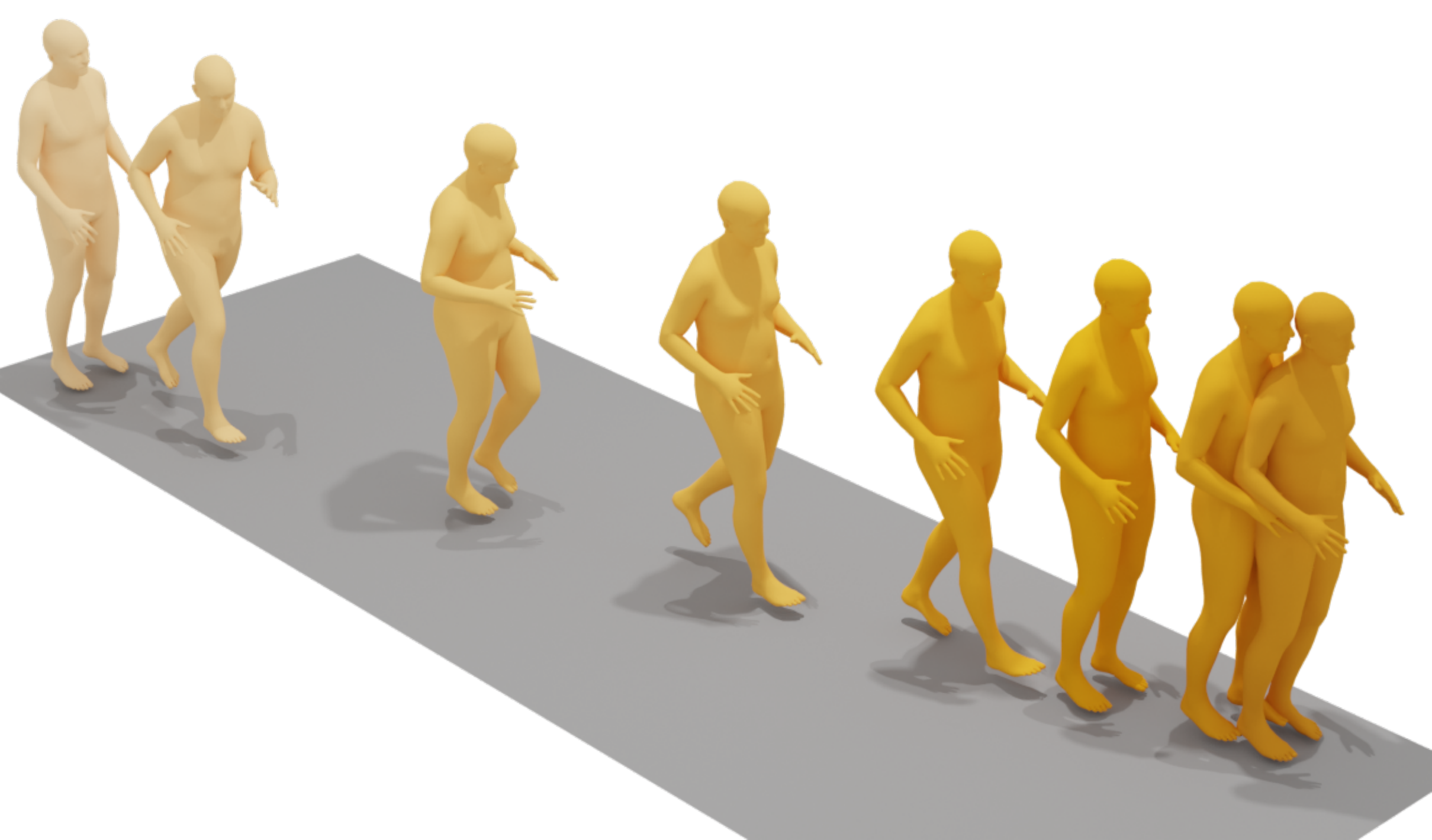} \\
		\bottomrule
	\end{tabularx}
\end{figure*}

{\bf Text-to-motion generation.} We supplemented extensive text-to-motion generation visualization with Attribute Diffusion Model in Fig. \ref{fig:VisualizationSupTTM}.
\section{Experimental Design Details}
\subsection{Research motivation}
We are hoping to create an end-to-end controllable generation application, meaning that anyone can get the motions they wish for. This faces the challenges of being difficult to control and having a generalization problem that goes beyond the dataset. So the core idea is to quickly fine-tune the model to 'see' new patterns or convert unseen descriptions to dataset-specific text.\par
The origin of the LLM Planner was an application where we wanted to "jump into the well," but we found that the model didn't understand the context ({\it i.e.} well). We thought and analyzed this precision paradox. The text in HumanML3D is so clean that it is difficult for users to interact with. Higher and higher aligned SOTA will only limit the training sample distribution to clean datasets.\par
The reason for the research of Motion Adapter is that we found that many motions are actually not in the HumanML3D dataset, especially in some games and movies, there are a lot of missing actions (e.g., fly) and unique character styles (e.g., superman). So, we wanted to get some stylized generated motions. Fortunately, 100STYLE has some new styles and actions that can be used as data for research samples. Finally, we are compatible with Trajectory ControlNet to implement control of multiple attributes.\par 
\subsection{Training details}
\begin{figure}[htbp]
	\centering
	\includegraphics[width=0.48\textwidth]{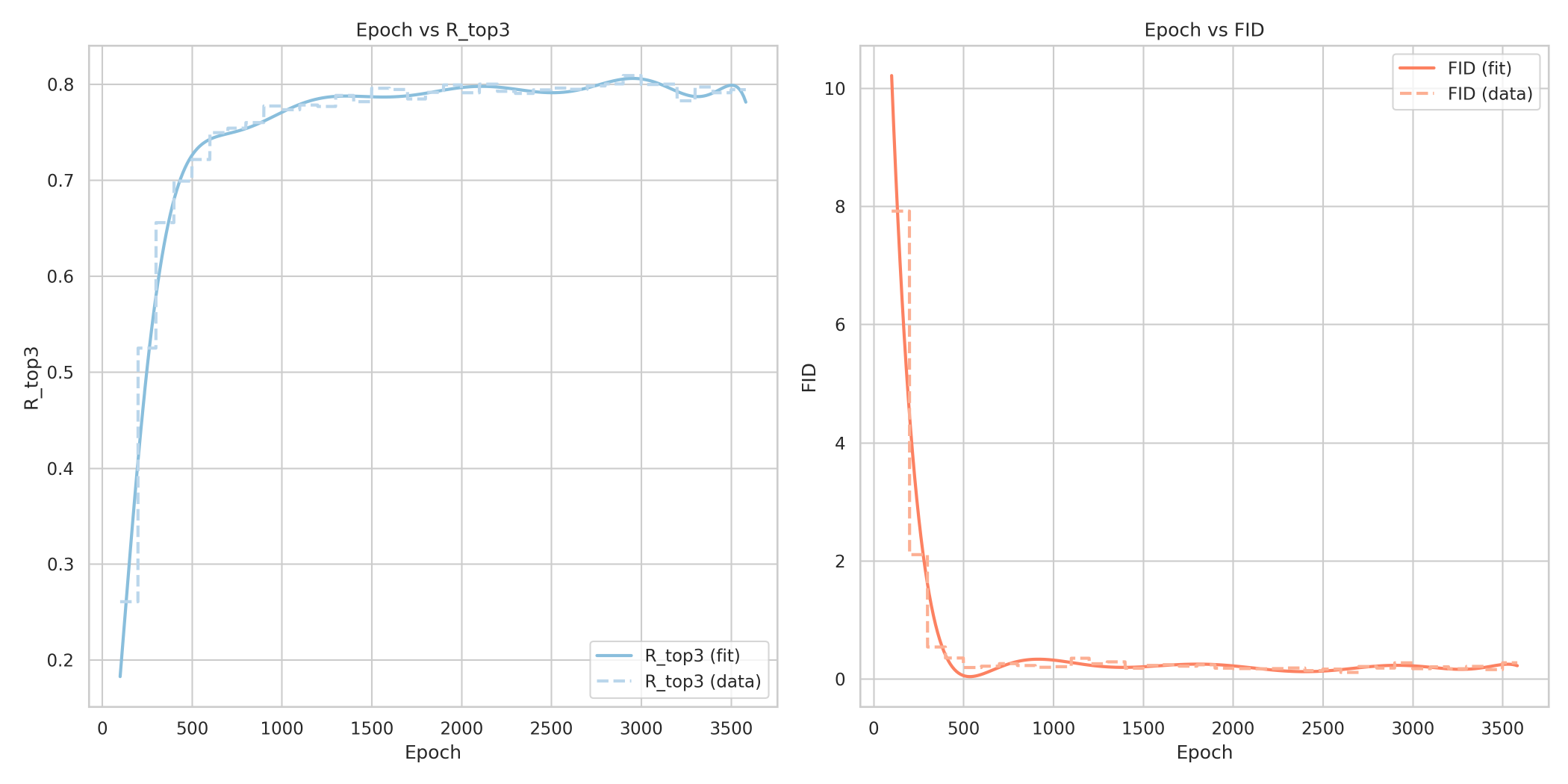}
	\caption{Training ADM detail diagram. }
	\label{fig:admepoch}
\end{figure}

\begin{figure}[t]
	\centering
	\includegraphics[width=0.48\textwidth]{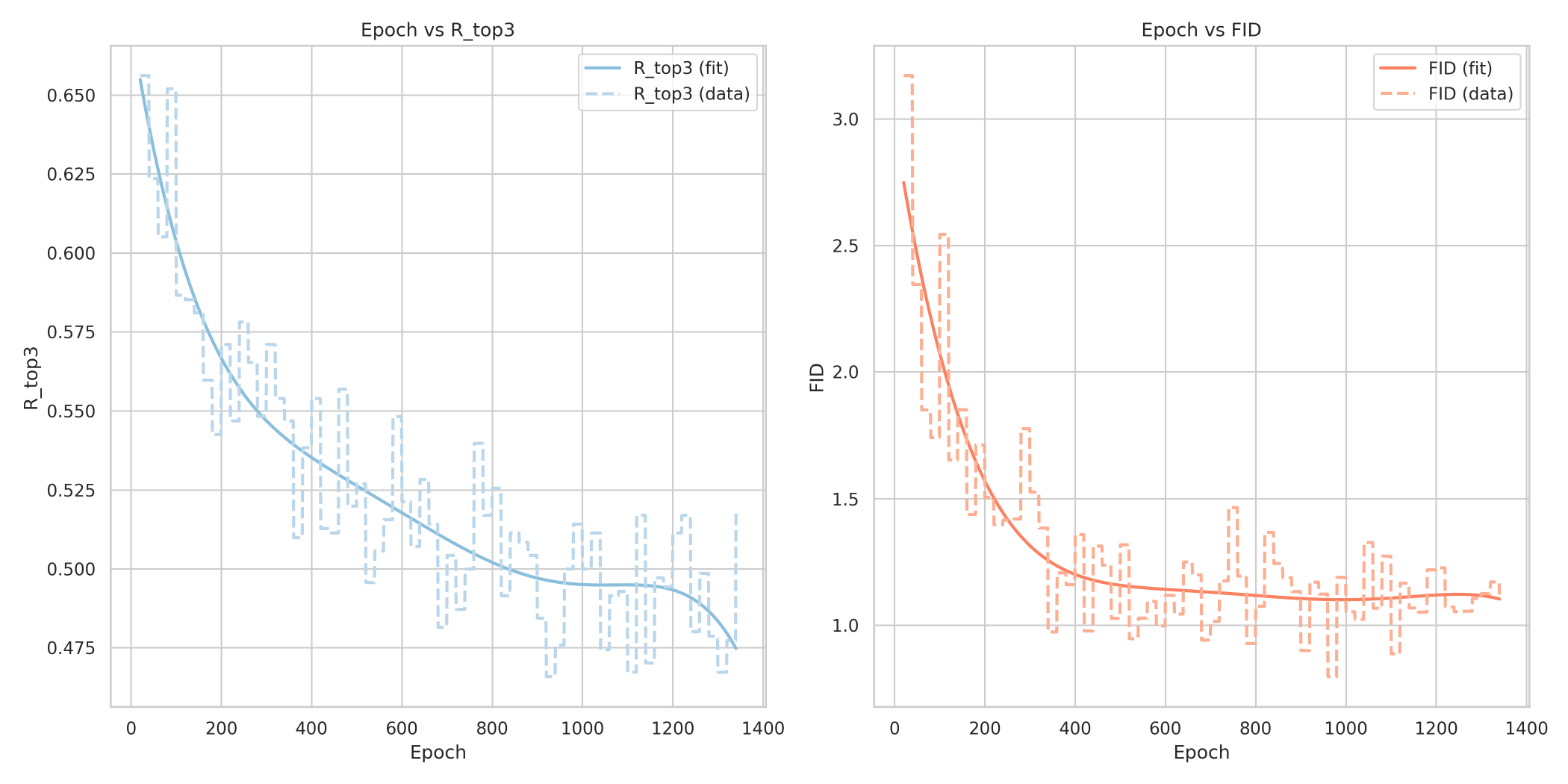}
	\caption{Finetuning self-attention detail diagram. This illustrates the overfitting phenomenon that results as the epoch learns. The pre-trained alignment knowledge of motion and text is severely lost.}
	\label{fig:motionadapterepoch}
\end{figure}

\indent \indent {\bf Training diagram.} We provided validation data for FID and R Precision when we trained the ADM, as shown in the Fig. \ref{fig:admepoch}. Then, we provided validation data for FID and R Precision from 1 to 1000 epochs with only self attention when we trained the Motion Adapter, as shown in the Fig. \ref{fig:motionadapterepoch}. \par
{\bf Parameters of each module.} The number of model parameters for our entire network is shown in the Table \ref{tab:module_params}. Because Controlnet-based models require an initial duplication of the denoiser, followed by the incorporation of control information via zero convolutions. This highlights the superiority of our motion adapter, which directly integrates into the denoiser and achieves control conditions through network parameter reuse and the addition of a few attention layers. And our approach doesn't require training an style encoder. We used the trained VAE directly as the style encoder because they are both motion mode.\par
\begin{table}[htbp]
	\centering
	\resizebox{\columnwidth}{!}{
	\begin{tabular}{c | c c c c c}
		\hline
		\textbf{Modular} & \textbf{text encoder} & \textbf{vae} & \textbf{denoiser} & \textbf{controlnet} & \textbf{traj\_encoder} \\
		\hline
		\textbf{Param.} & 335 M & 18.0 M & 29.8 M & 30.7 M & 7.8 M \\
		\hline
	\end{tabular}
	}
	\caption{Model parameters for different modules.}
	\label{tab:module_params}
\end{table}
{\bf Inference Tips.} To better replicate our results in each table, we offer some tips. The random number seed of all experiments was set to 1234, and then the batchsize during test was 32, the guidance scale was adjusted according to the tables, and two A6000 were used. The above parameters are the main factors that affect the different effects of the same training weight.\par 
Therefore, we recommend that when using the demo, generate multiple identical text inputs (e.g. eight same texts) at the same time, so as to construct a batch to produce better results. Meanwhile, the descriptions of the text, style, and trajectory should be as consistent as possible, without any conflict (e.g., 'bent' in the text and 'old' for the style complement each other well).

\subsection{Trajectory ControlNet}
\begin{figure}[htb]
	\centering
	\includegraphics[width=0.4\textwidth]{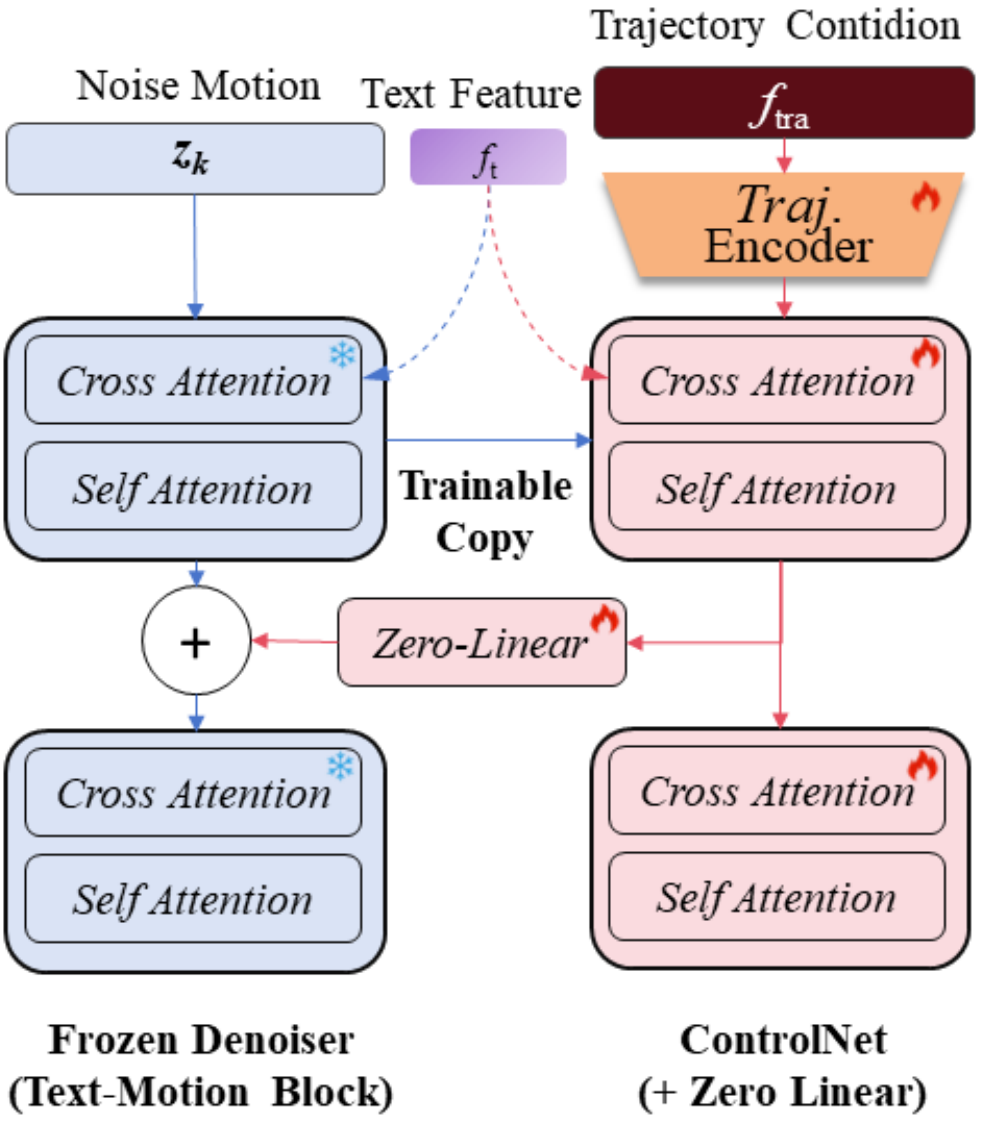}
	\caption{Finetuning self-attention detail diagram. This illustrates the overfitting phenomenon that results as the epoch learns. The pre-trained alignment knowledge of motion and text is severely lost.}
	\label{fig:trajectoryNet}
\end{figure}

In our Trajectory ControlNet, we only design a simple ControlNet to verify the compatibility of the Motion Adapter and ControlNet to achieve multi-mode control. We have found that guided diffusion model control can be very time consuming, even if it works well. This is the opposite of the lightweight and fast design of our motion adapter. Specifically, our ControlNet design is shown in the Figure \ref{fig:trajectoryNet}. Training the Trajectory ControlNet for 1000 epochs takes approximately 24 hours without Control loss, and roughly 48 hours with control loss.

\subsection{Training Loss}
The train loss of our work follows the MLD \cite{mld} implementation. We only trained with text conditions in Stage 1. In Stage 2, the model are finetuned through other datasets, is still trained with text as a condition. In Stage 3, the condition consists of text and trajectories. The proposed Motion Adapter and ControlNet aligns with the training loss $L_{ADM}$ of diffusion model.\par
However, we found that when training ControlNet, the loss constraint only in the latent space is not strong enough. This can provide a trajectory guidance rather than a mandatory trajectory. Although as a verification of compatibility, it is enough. Following \cite{motionlcm}, we tried to set a control loss, a $l2$ loss in original space after vae decode, but the model did not converge well, we are trying to solve it.

\subsection{Stack Layers Design}
\begin{figure}[htb]
	\centering
	\includegraphics[width=0.3\textwidth]{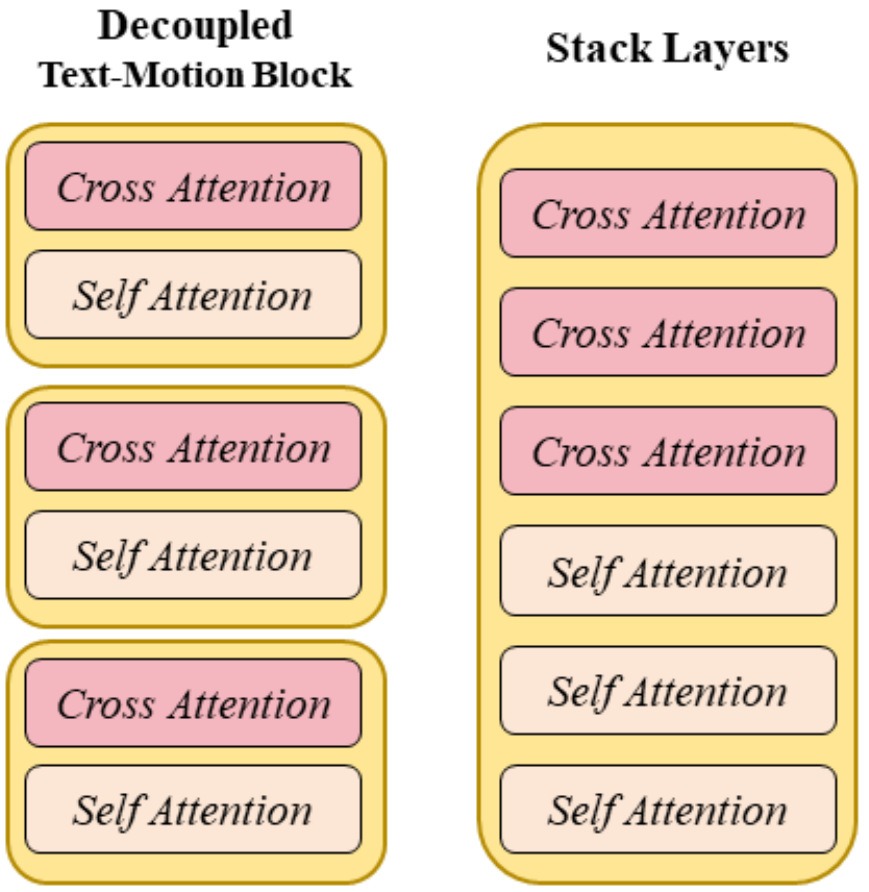}
	\caption{Finetuning self-attention detail diagram. This illustrates the overfitting phenomenon that results as the epoch learns. The pre-trained alignment knowledge of motion and text is severely lost.}
	\label{fig:stacklayers}
\end{figure}

In the proposed attribute diffusion model, we conducted ablation studies on stacked layers to explore whether the learning of text and motion should be alternated. Specifically, we designed experiments stacking motion self-attention and text cross-attention, as illustrated in the Fig. \ref{fig:stacklayers}. Our findings validate the effectiveness of decoupling the text-to-motion block.
\subsection{Classifier-free diffusion guidance}
Our denoiser $g$ is trained with classifier-free diffusion guidance, learning both the conditioned and unconditioned distributions. A linear combination of these is used as follows:
\begin{equation}
	\begin{split}
		g_{\theta,w}^{\text{cfg}}(z_t,t,c) &= wg_\theta(z_t,t,c) + (1-w) g_\theta(z_t,t,\varnothing ), \\
		\{f_t,f_m,f_{\text{tra}}\} &\in c,
	\end{split}
\end{equation}
where $c$ represents the conditions ( {\it i.e.}, the feature of texts $f_t$, motion prompt $f_m$ and trajectories $f_{\text{tra}}$), and w is the guidance scale. A value of $w>1$ strengthens the guidance effect. This approach effectively balances the trade-off between boosting sample quality and reducing diversity in conditional diffusion models.
\section{Additional Experiments}

\subsection{Comparasion of VAE}
By adjusting the parameter configuration of MLD-1, we achieved the VAE results for our VAE-7. The primary modification involved setting the size of the latent space to 7. This adjustment was based on previous experiments within the MLD framework, which demonstrated that a latent space size of 7 yields the most realistic VAE results.\par 
As shown in Table \ref{tab:com_vae}, we compared our approach with all current state-of-the-art VAE methods in terms of motion reconstruction performance, where VAE reconstruction error is evaluated by MPJPE and PAMPJPE, measuring errors in millimeters. This demonstrates that our motion representation space is the most faithful to real motion, thereby preserving more kinematic details.\par
\begin{table}[htb]
	\centering
	\caption{Evaluation of our motion encoder on the motion part of HumanML3D dataset. We follow \cite{motiongpt} to evaluate our VAE model $E$. MPJPE and PAMPJPE are measured in millimeter. Compared to other motion codecs, our method is obviously closer to the most realistic data.}
	\resizebox{\columnwidth}{!}{
		\begin{tabular}{ccccc}
			\hline
			\multirow{2}{*}{Method} & \multicolumn{4}{c}{Motion Reconstruction} \\ 
			~ & MPJPE↓ & PAMPJPE↓ & FID↓ & DIV→ \\ 
			\hline
			Real & - & - & 0.002 & 9.455 \\ 
			\hline
			TM2T \cite{tm2t} & 230.1 & - & 0.307 & - \\ 
			ACTOR \cite{petrovich2021action} & 65.3 & 41.0 & 0.341 & 9.569 \\ 
			MLD-1 \cite{mld} & 54.4 & 41.6 & 0.247 & 9.630 \\ 
			MotionGPT \cite{motiongpt} & 55.8 & 40.1 & 0.067 & 9.675 \\ 
			MoMask \cite{momask} & 29.5 & 21.9 & 0.019 & 9.609 \\ 
			\hline
			\textbf{Ours} & \textbf{{12.8}} & \textbf{{8.7}} & \textbf{{0.007}} & \textbf{{9.530}} \\ 
			\hline
		\end{tabular}
	}
	\label{tab:com_vae}
\end{table}
\subsection{Comparison experiment of Trajectory}
As shown in Table \ref{tab:trajComparsion}, our experiments validate ControlNet's effectiveness in latent space under $L_{ADM}$, demonstrating that trajectory-guided motion effectively reduces positional errors by aligning poses with trajectories. While it underperforms compared to state-of-the-art trajectory-based methods, likely due to insufficient latent space loss constraints, this outcome sufficiently verifies the compatibility between our proposed Motion Adapter and Trajectory ControlNet.
\begin{table}[htb]
	\centering
	\caption{Evaluation of our trajector controlnet.  “LC” and “MC” refer to the control supervision introduced in the latent space and motion space. It shows that our trajectory control can provide enough trajectory guidance.}
	\resizebox{\columnwidth}{!}{
		\begin{tabular}{cccc}
			\hline
			\multirow{2}{*}{Method} & Traj.err.↓ & Loc.err.↓ & Avg.err.↓ \\ 
			& (50cm) & (50cm) & (50cm)\\
			\hline
			OmniControl (MC) & 0.3362 & 0.0322 & 0.0977\\ 
			MotionLCM (LC\&MC) & 0.1960 & 0.0143 & 0.1092\\ 
			\hline
			Ours w/o Control & 0.6208 & 0.4024 & 0.8201\\
			Ours (LC) w/ Control & 0.5498 & 0.3341 & 0.5872\\
			\hline
		\end{tabular}
	}
	\label{tab:trajComparsion}
\end{table}

\section{Dataset process Details}
{\bf HumanML3D dataset.} Thanks to the data processing work of \cite{T2M}, we identified several annotation errors in HumanML3D as well as instances of misspelled words. We did not address the annotation errors, but the LLM Planner provides an excellent solution for correcting spelling mistakes. For the 'left,right,clockwise' orientation of the current model, {\it  we found that there are certain annotation errors in the dataset, especially 'left''right'. We haven't solved it yet, for a fair comparison, but we think it's the key to the puzzle model.}\par
{\bf 100STYLE dataset.} Thanks to the data processing work of \cite{smoodi}, we found that there are some retargeted actions missing in the dataset, including WhirlArms, WideLegs, WiggleHips, WildArms, WildLegs, and Zombie. We directly removed the data for these actions, approximately one thousand entries, and fine-tuned the model using the remaining 15,000 or so entries. The difference is that we re-conducted the training and testing split of the dataset. Smoodi was trained by mixing two datasets to maintain the original knowledge, while we only fine-tuned on the new dataset. The smaller reduction in our R Precision illustrates the superiority of our fine-tuning approach. Of course, this also led to a certain amount of text and motion alignment knowledge forgotten, we are also trying to mix the original data to fine-tune.\par
{\bf Motion Representations.} We unify all dataset input representations as HumanML3D via the tools provided by \cite{T2M}, then obtain latent vector representations in latent space through VAE, and finally use SMPL \cite{loper2023smpl} format and BVH for visualization and application.\par 

\section{Metrics Introduction}
{\bf Rule Consistency Score ($RCS\in\{0,1\}$)} measures strict rule compliance in generated instructions, as dataset alignment depends on this consistency. LLM as binary evaluator determines the model results, and then calculates the correct rate:
\begin{equation}
	RCS = \frac{1}{n} \sum_{i=1}^{n} \left\{f(I_{model}^i)==Ture\right\},
\end{equation}
where $n$ denotes the number of answer; $f$ is LLM as a judge function, with the value of boolean True or False; $I_{model}^i$ is the instruction of $i-th$ model. RCS results range $0-100\%$.\par
{\bf Plan Score ($PS\in[1,5]$)} measures the relative scores of five large language models. We obtain the scores of each model by ranking these methods. The closer the model is to 1, the higher the ranking is, and the better balance between real instructions and rules is achieved. Our Plan Score is the average score of different instructions:
\begin{equation}
	PS = \frac{1}{n} \sum_{i=1}^{n} f(I_{user};I_{model}^i),
\end{equation}
where $I_{user}$ denotes the origin instruction for reference; $f$ is LLM as a sort function, with the value of an integer from 1 to 5; $I_{model}^i$ is the instruction of $i-th$ model.\par
\begin{figure*}[htb]
	\centering
	\caption{An example illustrates the bad case. All auto text annotations are 'a person moves around with their arms outstretched.' This leads to the interference of the text information, making the stylized generated results random.}
	\label{fig:badcaseS}
	\begin{tabularx}{\textwidth}{>{\centering\arraybackslash}X
			>{\centering\arraybackslash}X
			>{\centering\arraybackslash}X
			>{\centering\arraybackslash}X|
			>{\centering\arraybackslash}X
		}
		\toprule
		Aeroplane
		&
		Akimbo
		&
		ArmsFolded
		&
		InTheDark
		&
		Generation\\
		\includegraphics[width=\linewidth]{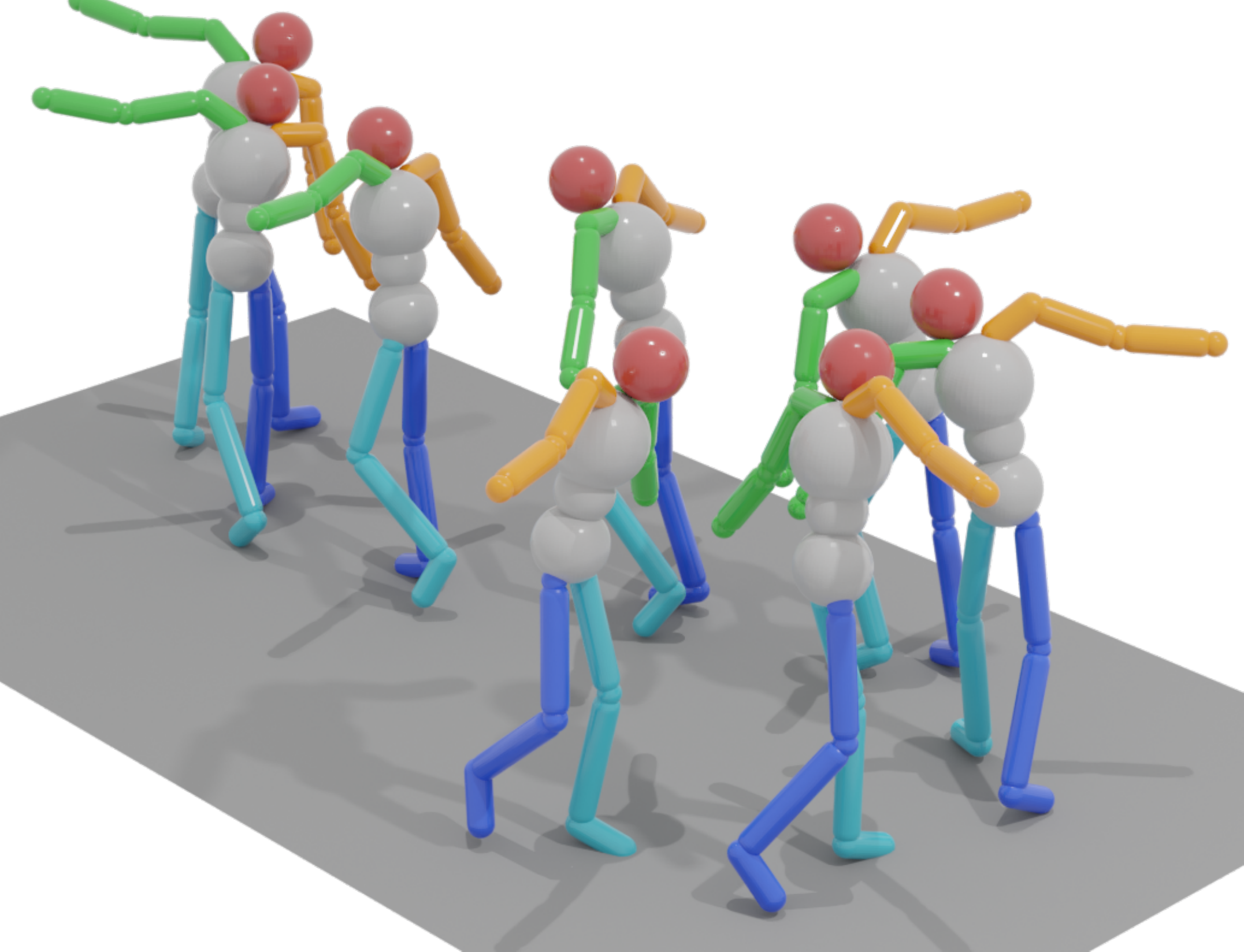} &
		\includegraphics[width=0.9\linewidth]{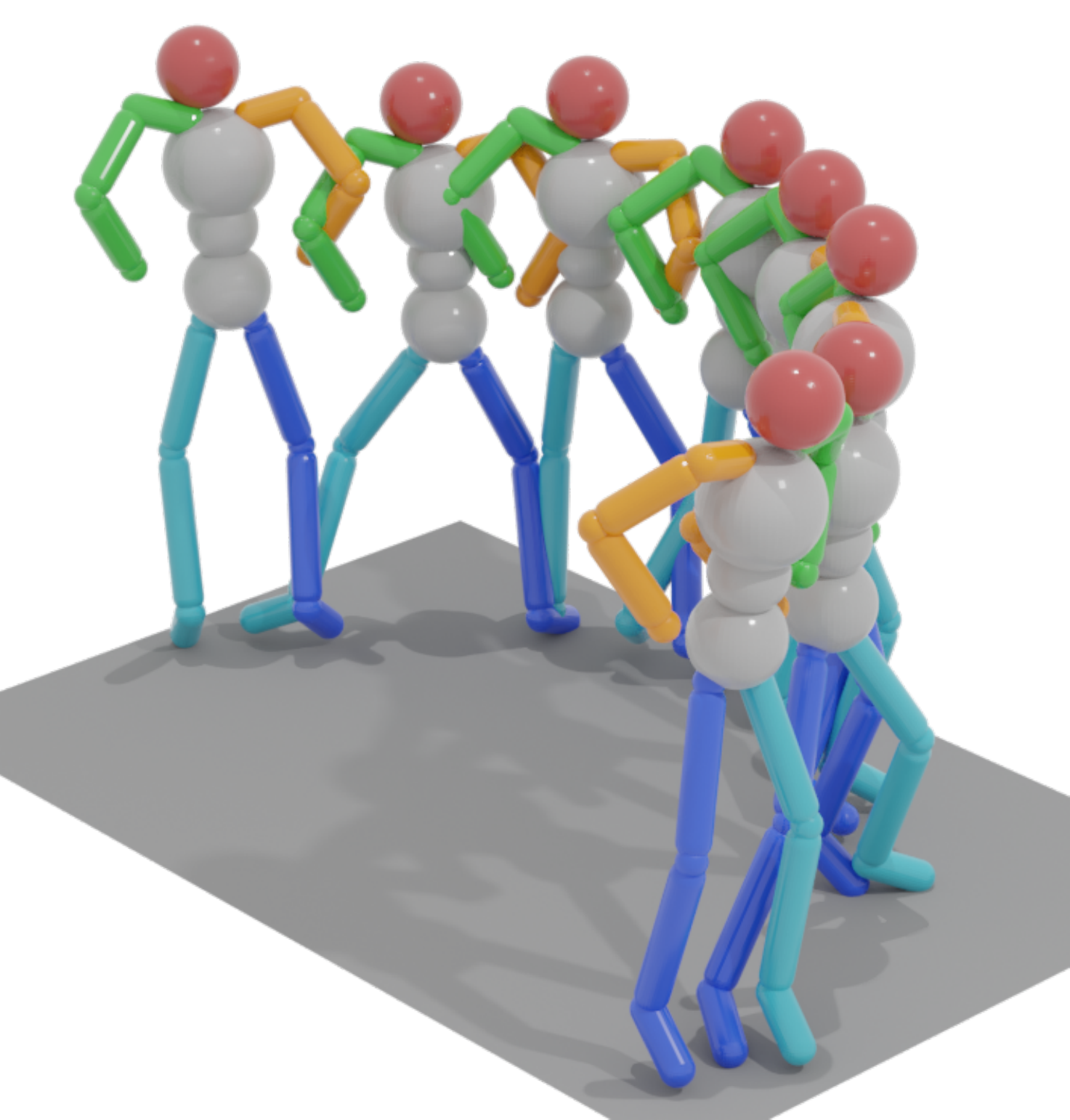} &
		\includegraphics[width=\linewidth]{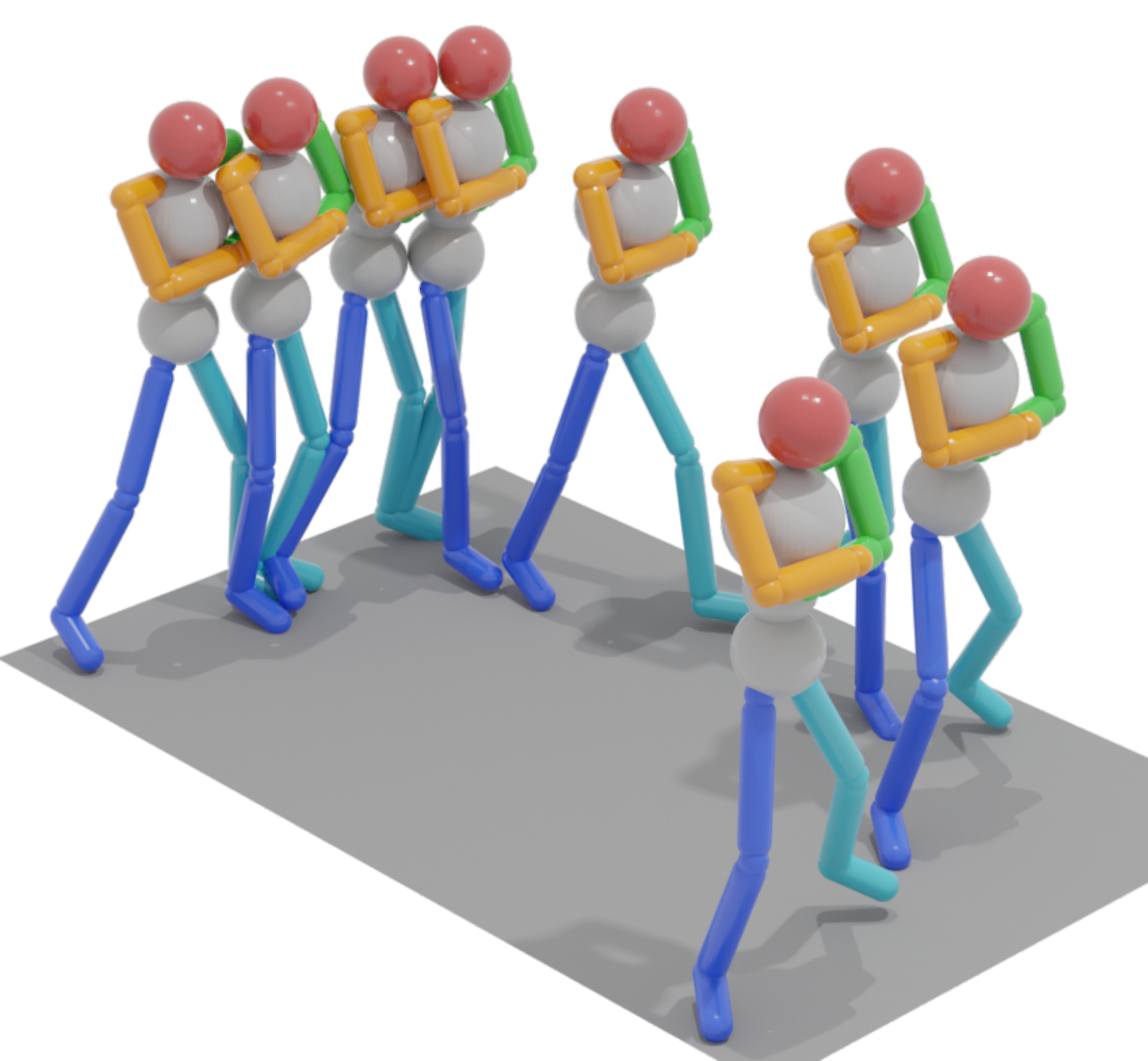} &
		\includegraphics[width=0.9\linewidth]{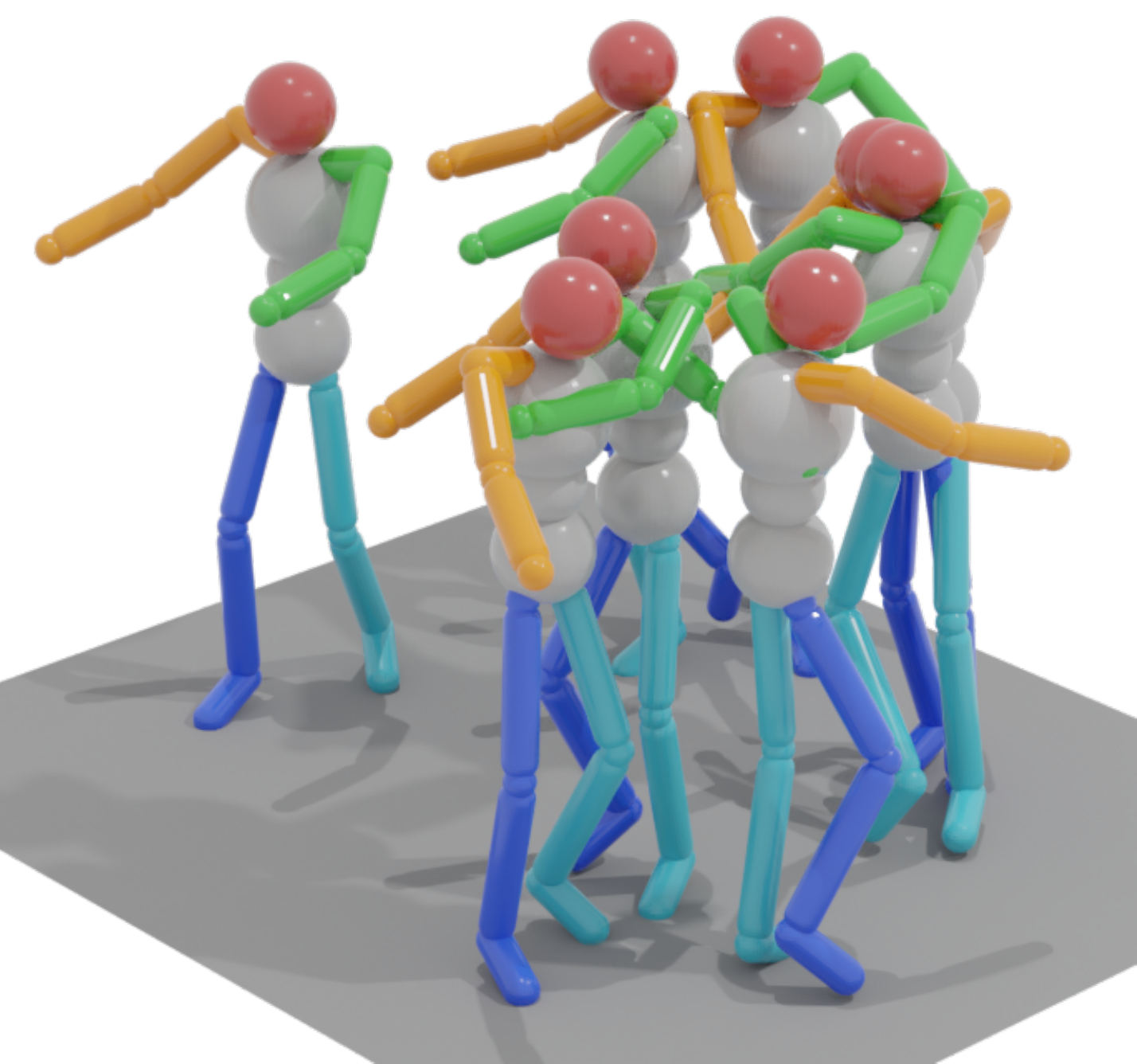} & 
		\includegraphics[width=\linewidth]{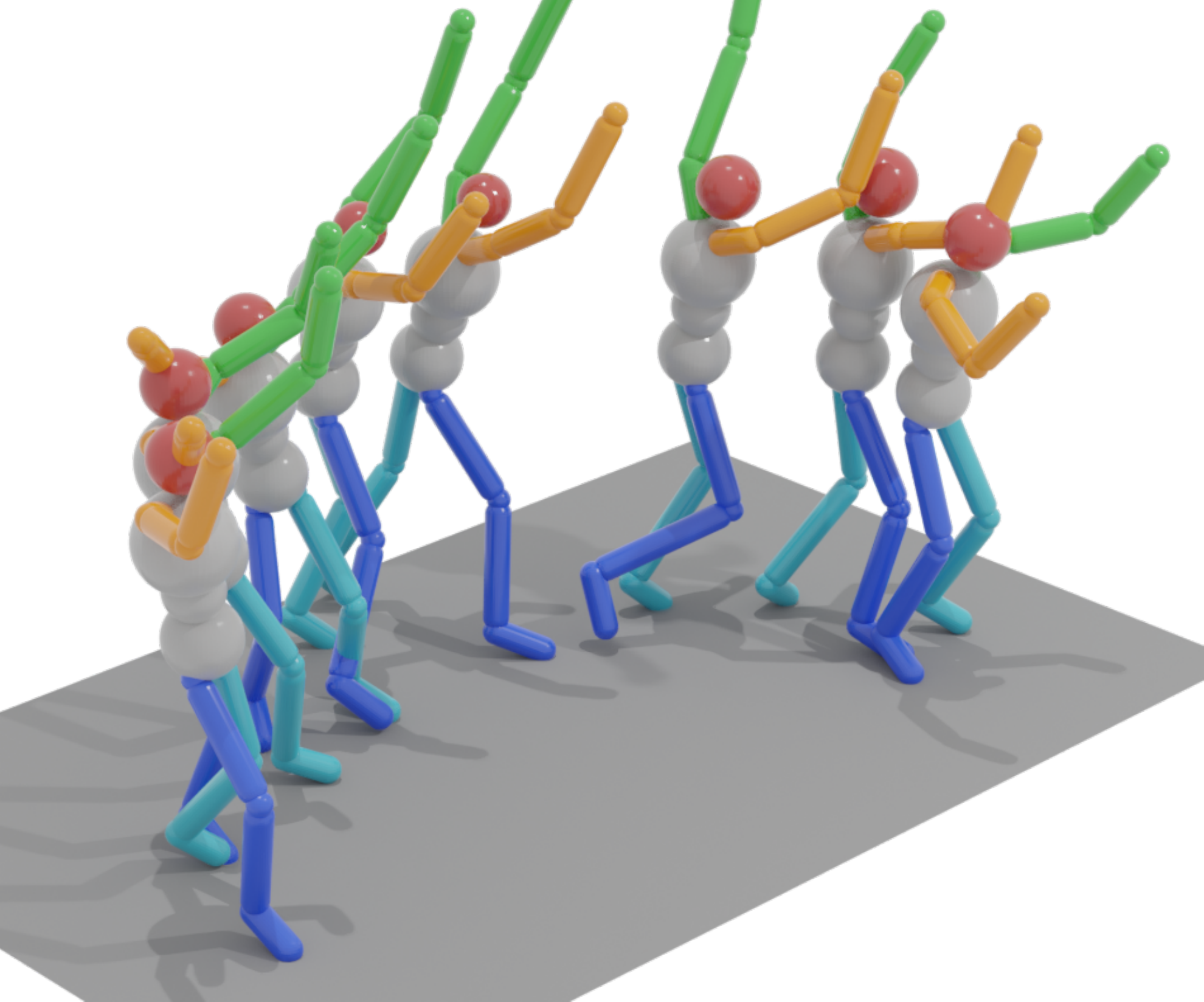}\\
		\bottomrule
	\end{tabularx}
\end{figure*}

\section{Limitations}
We honestly discuss any limitations encountered during the research and hope to discuss and resolve them in the future. We would like to call for recognition of some new limitations, which will also give more inspiration to the research field. Finally, we hope that it does not affect the innovation of our work.\par
\begin{figure}[htb]
	\centering
	\includegraphics[width=0.3\textwidth]{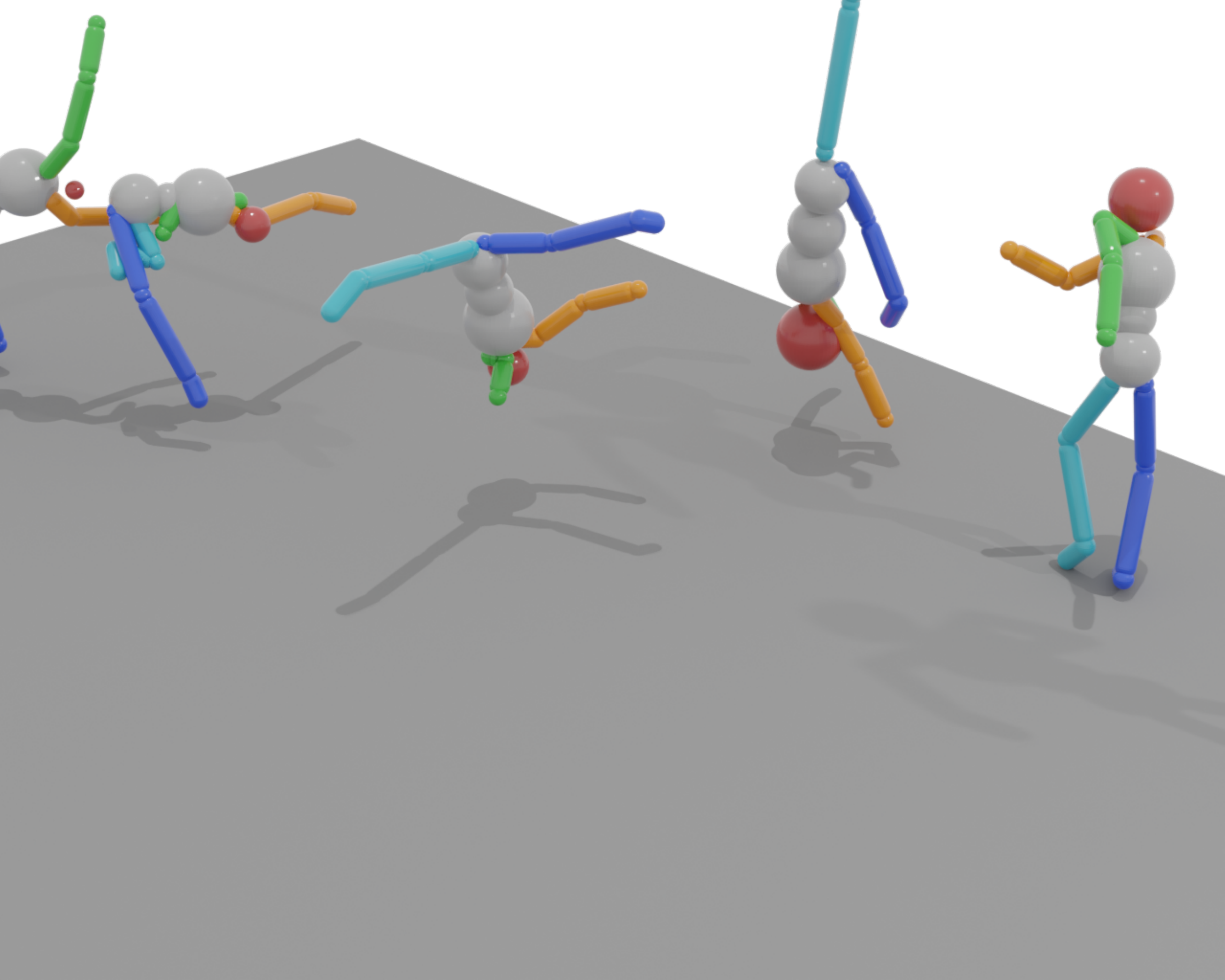}
	\caption{A bad case of 'A person parkour-flips over a sinking river ferry’s railings, dodging panicked passengers, to restart engines before currents drag it into a waterfall.' Long text, long move range, long time and zero-shot generation are very difficult.}
	\label{fig:badcase}
\end{figure}

{\bf Motion generation for long texts.} This limitation comes from our evaluation of the MotionIT dataset, where we found that existing methods do not work well for the generation of long text stories. For example: "A person parkour-flips over a sinking river ferry’s railings, dodging panicked passengers, to restart engines before currents drag it into a waterfall." The result after llm planner is: "A person flips over railings quickly, dodges passengers, and runs forward to start urgently." The bad case is shown as Fig. \ref{fig:badcase}. We find that such an motion sequence contains multiple actions that need to be generation, and inevitably exists information that is not present in the dataset. In addition, move range in the motion is wide and the generation time is long, which brings great challenges to the existing motion generation.\par 

{\bf Foot Skating.} The foot skating cause of this was the finetuning we did on the new data set. However, the mean and variance of different datasets (HumanML3D and 100STYLE) are different, which will lead to incorrect normalization, thus resulting in the visualization of foot skating. We consider that any training with multiple datasets may encounter this problem. We will solve it in the future.\par 

{\bf Multi-condition conflicts in motion synthesis.} Multiple control conditions (e.g., style, text, trajectory) may induce optimization conflicts that compromise motion stability. We found the same problem with other methods \cite{smoodi}, requires repeated parameter adjustment and multiple generation to get desired motion. Explicit contradictions between semantic (text) and kinematic (style/trajectory) constraints particularly risk yielding unpredictable outcomes.\par
Our modular architecture, while offering interpretability and selective condition application through decoupled components, inherits fundamental limitations of multi-objective optimization - specifically reduced stability under conflicting multi-condition inputs.\par
We analyze that the conflict between style and text is may caused by automatic text annotation. Because current automated text annotations make it easy to get the same text description for all types of actions, a text-style generated motion is actually derived from the supervised action category, as shown in Fig. \ref{fig:badcaseS}. This is most likely one of the difficulties that led to the current generation of multiple conditions.

\section{LLM Planner Details}
In our experiment, we utilized ChatGPT-4o and deepseek-r1 to create the MotionIT dataset and implemented LLM Planner using various large models. Finally, we used a deployed deepseek-r1-32b as the critic for evaluation. The following are our prompts and implementation details.\par
\subsection{prompt}
\begin{framed}
	{\footnotesize
	{\bf GPT-4o/Deepseek Prompt (Data Generation)}\par
	You are now a motion analysis assistant. I need you to generate descriptions of physical body actions that require reasoning. Direct actions include walking, running, jumping, etc. Actions requiring reasoning involve additional context or attributes, such as walking after a paper is accepted or running to help a child who has fallen.\par
	•  Focus solely on full-body physical actions, avoiding facial expressions or hand details.\par
	•  Begin each description with “a person.”\par
	
	Please provide 100 different descriptions of physical body actions. Each description should be about 15 words. Ensure the descriptions are concise and clear! \par
	
	Example:\par
	a person walks nervously towards a podium to deliver an important speech at a conference.
	}
\end{framed}
We ensured the diversity of the generated results through human observation and simple adjustments to the prompt, which we generated every 100 examples in a group and collected 5K datasets. This dataset will be made public.
\begin{framed}
	{\footnotesize
	{\bf Prompt (As LLM Planner)}\par
	Act as a motion analysis agent that translates complex action instructions into simplified text commands from Frequent Word Bank. You need to consider whether the context is a necessary output, as the Frequent Word Bank usually does not contain context-related words. Maintain sentence fluency while following these rules:
	
	1.Strictly start outputs with "A person" in English
	
	2.Keep sentences concise - describe actions/characteristics minimally. Focus on the action and the character of the action itself.
	
	3.The actions described need to have distinguishing features, such as happy, left hand raised, etc
	
	4.Prioritize the use of words from the Frequent Word Bank (listed below), especially verbs, when appropriate
	
	5.Output ONLY the final sentence without explanations
	
	Examples Output:\par
	a person walking backwards slowly\par
	a man raises its right arm and wiggles it and brings it back down
	
	Example:\par
	User: A person is walking when his paper is accepted\par
	Agent: A man walks happily and relaxed with swinging arms.
	
	Frequent Word Bank:
	right forward ...
	
	Now input is : ...
	}
\end{framed}
\begin{framed}
	{\footnotesize
	{\bf Deepseek-r1-32B Prompt (For Rule Judgement)}\par
	Act as a rule-based validator for motion descriptions.     Return 'True' or 'False' with JSON based on strict compliance with these 5 non-negotiable criteria:
	
	1. Initial Phrase Enforcement
	Output MUST start with exactly "A person" or "a person" (case-sensitive match)
	
	2. Structural Integrity
	MUST be a single continuous sentence
	MUST contain NO explanations, context, or examples
	
	3. Attribute Requirement
	MUST include $>=$1 explicit distinguishing attribute (adverb/adjective: e.g., 'quickly', 'left hand', 'diagonally')
	The result should focus on the action itself, not the context information.
	
	4. Lexical Compliance
	$>=$50\% of verbs/body parts MUST come from the Frequent Word Bank
	
	5. Grammatical Soundness
	MUST maintain grammatical correctness and action fluency
	
	Validation Protocol:
	
	\{
		"validation\_result": \{
			"final\_result": "False", // This should be 'True' only if all criteria are satisfied
			"final\_explanation": "Not all criteria were met. Specifically, Criterion 1 failed because [brief reason]."
		\}
	\}
	
	Examples:
	User:
	"A person jumps" 
	Output:
	\{
		"validation\_result": \{
			"final\_result": "False",
			"final\_explanation": "Violates Criteria 3 that ..."
		\}
	\}
	
	User:
	"A person runs energetically through the rain, unfazed by the downpour soaking them." 
	Output:
	\{
		"validation\_result": \{
			"final\_result": "False",
			"final\_explanation": "Not all criteria were met. Specifically, Criterion 2 failed because the sentence contains context ..."
		\}
	\}
	
	User:
	"a person raises right arm and spins clockwise" 
	Output:
	\{
		"validation\_result": \{
			"final\_result": "True",
			"final\_explanation": "All pass because ..."
		\}
	\}
	}
\end{framed}
\begin{framed}
	{\footnotesize
	{\bf Deepseek-r1-32B Prompt (For Plan Score Judgement)}\par
	Act as an expert evaluator for motion description generation models. Your task is to rank outputs from 5 different models based on their adherence to specifications and information preservation.
	The result should focus on the action itself, not the context information.
	
	Evaluation Criteria (Descending Priority):
	
	1. **Format Compliance**: 
	- Must start with "A person"
	- Must output ONLY the final sentence. MUST contain NO explanations, context, or examples
	- Must be in English
	- Keep sentences concise - describe actions/characteristics minimally, focusing on the action and the character of the action itself
	- Prioritize the use of words from the Frequent Word Bank (listed below), especially verbs, when appropriate
	
	2. **Content Quality**:
	- Contains distinguishing characteristics/attributes (e.g., body parts, emotions, directions)
	- Maintains all key information from original instruction
	- Prioritizes words from Frequent Word Bank
	
	3. **Linguistic Quality**:
	- Concise and fluent expression
	- No redundant contextual information
	- Grammatically correct structure
	
	Task Description:
	You will receive:
	1. Original user instruction
	2. 5 model outputs (labeled A-E)
	
	Required Output Format:
	Rank the models from best to worst (1-5), where 1 is the best and 5 is the worst. Provide explanations for each model’s score based on the criteria.
	YOU MUST OUTPUT THE sorted LIST! Your sorted list must match the explanation! JSON Format!
	
	Example Output:
	\{
		"sorted\_list": ["A", "D", "B", "C", "E"],
		"explanation": \{
			"A": "Explanation for rank 1",
			"D": "Explanation for rank 2",
			"B": "Explanation for rank 3",
			"C": "Explanation for rank 4",
			"E": "Explanation for rank 5"
		\}
	\}
	}
\end{framed}
We tried evaluating with different models, but we found significant variations in the evaluation results when using different models, such as various versions of Qwen2.5. Finally, thanks to the role of the thought chain, we chose Deepseek-r1 as the large evaluation model. Its output thought chain process can effectively and more credibly judge the accuracy of the results.
\subsection{Frequent Word Bank}
The word bag of the first 512 high-frequency words in HumanML3D. We adopted TF-IDF to calculate our bag of words, the core code in the listing. The answer of the wordbag is shown in Table \ref{tab:wordbag}.

\begin{lstlisting}[caption={TF-IDF codes}]
from sklearn.feature_extraction.text import TfidfVectorizer
vectorizer = TfidfVectorizer(max_features=1024, stop_words='english')
tfidf_matrix = vectorizer.fit_transform(documents)
words = vectorizer.get_feature_names_out()
# Sum the columns to get the total TF-IDF score for each word
tfidf_scores = tfidf_matrix.sum(axis=0).A1  
word_tfidf = list(zip(words, tfidf_scores))
# sort and obtain 512 important words
important_words = sorted(word_tfidf, key=lambda x: x[1], reverse=True)[:1024]
\end{lstlisting}
\begin{table}
	\caption{Frequent Word Bank. This recorded 512 high-frequency words from the HumanML3D dataset, which were used as Prompt inputs in the LLM to allow the large model to understand the text information in the dataset.}
	\resizebox{\columnwidth}{!}{
		\begin{tabular}{|lllllll|}
			\hline
			right & forward & left & hand & walk & arm & step \\ 
			slowly & leg & circle & head & turn & foot & backwards \\ 
			stand & jump & motion & place & straight & slightly & clockwise \\ 
			raise & object & sit & face & body & ground & run \\ 
			stop & quickly & wave & throw & hold & pick & air \\ 
			line & bend & lift & direction & knee & kick & shoulder \\ 
			jog & dance & start & pace & look & chest & make \\ 
			bent & balance & squat & floor & jumping & stretch & backward \\ 
			lean & knees & sides & swing & spot & push & diagonally \\ 
			reach & legs & small & running & elbow & appear & cross \\ 
			bring & hop & way & grab & hip & stair & crawl \\ 
			level & shake & waist & clap & punch & using & wide \\ 
			dancing & jack & touch & gesture & carefully & high & spin \\ 
			end & chair & stick & thing & near & pose & drink \\ 
			half & casually & extend & room & opposite & uses & sway \\ 
			crouch & big & upper & catch & mouth & raising & hips \\ 
			little & shape & come & wrist & lifting & weight & path \\ 
			apart & taking & continue & places & pushed & stance & fall \\ 
			drop & point & begin & quick & curve & try & upwards \\ 
			stopping & rotate & center & open & couple & perform & angle \\ 
			watch & table & scratch & manner & kneel & climb & exercise \\ 
			play & bow & crosses & hit & spread & pull & box \\ 
			shuffle & outward & crawls & movement & warm & twist & speed \\ 
			wipe & rail & surface & motions & short & steady & basketball \\ 
			support & shakes & facing & torso & upward & ahead & trying \\ 
			gently & kicking & cartwheel & squatting & shelf & leaning & rub \\ 
			long & dances & swaying & wall & area & elbows & forearm \\ 
			curved & middle & punches & touches & briskly & wheel & lightly \\ 
			complete & crouches & gestures & screen & degrees & act & extended \\ 
			upright & neck & sets & swimming & boxing & rest & sidestep \\ 
			bottle & playing & golf & dumbbell & suddenly & bounce & rotates \\ 
			stomach & shaking & started & outstretched & guitar & door & arc \\ 
			shuffles & check & wash & handrail & camera & moment & close \\ 
			picking & downward & railing & kneels & seat & clean & quarter \\ 
			shift & limp & dodge & eat & balancing & clapping & stay \\ 
			avoid & hopping & repeats & overhead & lunge & dog & slide \\ 
			hunch & performs & crouched & pour & steering & scratches & leap \\ 
			skip & karate & pivot & briefly & catches & thigh & grabbed \\ 
			crossing & sidesteps & stride & sprint & trip & pause & placing \\ 
			bows & toe & plane & reaching & land & alternate & drinking \\ 
			pretend & flap & lay & swim & salsa & zig & completely \\ 
			fours & crouching & roll & football & help & confidently & tilt \\ 
			hard & semi & hunched & waltz & pain & rubs & fold \\ 
			acting & stool & length & action & relaxed & recover & curl \\ 
			fight & march & moved & fist & cautiously & oval & downwards \\ 
			twists & spinning & pivots & rapidly & ramp & window & stagger \\ 
			grabbing & clock & obstacle & washing & wiping & heel & stagger \\ 
			march & clasp & slope & returning & palm & proceeds & distance \\ 
			gap & elevated & pauses & stationary & eating & rotating & bicep \\ 
			horizontally & ladder & stir & trips & crossed & uneven & toes \\ 
			zigzag & higher & jumped & edge & scratching & ankle & simultaneously \\ 
			shoe & bringing & shifts & inward & stroke & ballet & lowering \\ 
			happily & partner & pushing & food & hurt & opens & various \\ 
			lead & style & leading & carry & location & weights & keeping \\ 
			talk & space & hang & defensive & container & stumbling & follow \\ 
			drag & wiggle & aggressively & falls & horizontal & semicircle & fastly \\ 
			sweeping & beginning & slides & form & tap & brush & dribble \\ 
			toss & bounces & coming & posture & alternating & flight & practice \\ 
			lunges & tie & clasps & pacing & uppercut & stretched & balances \\ 
			rock & rubbing & chin & neutral & resting & pulls & lands \\ 
			objects & flex & swat & finger & wind & parallel & random \\ 
			extending & warming & keeps & similar & lose & cleaning & thighs \\ 
			infront & viewer & single & round & water & outwards & club \\ 
			abruptly & mix & came & block & stomp & vertically & press \\ 
			glass & light & far & jab & lap & stiff & folds \\ 
			drive & tries & movements & exercises & hello & terrain & gait \\ 
			swiftly & eye & shot & veer & signal & type & wobble \\ 
			wait & zombie & leaps & woman & straighten & rolls & sneak \\ 
			mid & continuously & duck & landing & work & backs & tilts \\ 
			imitate & bowl & pours & handstand & catching & row & tip \\ 
			cover & tight & pass & steadily & acts & pretending & crowd \\ 
			dry & salute & exercising & width & talking & centre & ends \\ 
			vigorously & brisk & belly & shrug & strides & flip & recovers \\ 
			overhand &  &  &  &  &  &  \\ 
			\hline
		\end{tabular}
	}
	\label{tab:wordbag}
\end{table}

{\small
	\bibliographystyle{ieee_fullname}
	\bibliography{RefMain}
}

%% file: sec/0_abstract.tex
\begin{abstract}
	Attributes such as style, fine-grained text, and trajectory are specific conditions for describing motion.  However, existing methods often lack precise user control over motion attributes and suffer from limited generalizability to unseen motions. This work introduces an Attribute Controllable Motion generation architecture, to address these challenges via decouple any conditions and control them separately.
	Firstly, we explored the Attribute Diffusion Model to imporve text-to-motion performance via decouple text and motion learning, as the controllable model relies heavily on the pre-trained model.
	Then, we introduce Motion Adpater to quickly finetune previously unseen motion patterns. Its motion prompts inputs achieve multimodal text-to-motion generation that captures user-specified styles. 
	Finally, we propose a LLM Planner to bridge the gap between unseen attributes and dataset-specific texts via local knowledage for user-friendly interaction. 
	Our approach introduces the capability for motion prompts for stylize generation, enabling fine-grained and user-friendly attribute control while providing performance comparable to state-of-the-art methods. Project page: \url{https://mjwei3d.github.io/ACMo/}
\end{abstract}

%% file: sec/1_introduction.tex
\section{Introduction}
\indent \indent Motion generation from text and controllable signals offers a wide range of applications such as video games, animation production, virtual reality, and embodied robots. However, several key challenges hinder broader adoption: 1) Text inherently lacks granularity for precise motion control. 2) Real-world applications require handling unseen motion patterns beyond training data. For instance, "walk clockwise" produces semi-circle or full-circle paths, both valid but requiring control, while excessive text descriptions confuse models and reduce accuracy. Novel motion patterns (e.g., "elderly walking" or "when paper accepted") remain challenging without training or fine-tuning samples. \par
To adress these issues, we propose to {\bf decouple any conditions and control them} during the generation process. We analyze that motion is a combination of heterogeneous yet related features, such as action, motion style, and trajectory. Similarly, a text-to-motion description comprises key verbs ({\it e.g.}, walk, jog, swim) and attributes ({\it e.g.}, angry, arms uplifted, and clockwise). We aim to develop an end-to-end application for controllable stylized text-to-motion generation, effectively addressing the proposed challenges, as illustrated in Fig. \ref{fig:ACMo_idea}. So, we design an \underline{A}ttribute \underline{C}ontrollable \underline{Mo}tion generation architecture called {\it ACMo} via plug-and-play modulars, which uses independent parameters to learn various conditional constraints. As shown in Tab. \ref{tab:ability}, it enables precise control in multimodal motion generation, including text, motion style, and trajectory. \par
\begin{table}[t]
	\caption{ Comparison of recent  state-of-the-art method on diverse text-to-motion tasks. Our work achieves fine-grained and user-friendly controllable motion generation.}
	\resizebox{\columnwidth}{!}{
		\begin{tabular}{lcccc}
			\toprule
			\multirow{2}{*}{Methods} & 
			\multicolumn{3}{c}{\bf Attribute} & \multirow{2}{*}{Reason}\\
			\cline{2-4} \rule{0pt}{10pt} 
			& {word-level} & {Style} & {Trajectory} \\ \hline
			
			Momask \cite{momask} & $\checkmark$ & $\times$ & $\times$  & $\times$ \\
			SmooDi \cite{smoodi} & $\times$ & $\checkmark$ & $\times$ & $\times$\\
			OmniControl \cite{omnicontrol} & $\times$ & $\times$ & $\checkmark$ & $\times$ \\
			ChatPose \cite{chatpose}& $\times$ & $\times$  & $\times$ & $\checkmark$  \\
			\hline
			Ours & $\checkmark$ & $\checkmark$ & $\checkmark$ & $\checkmark$\\
			\bottomrule
		\end{tabular}
	}
	\label{tab:ability}
\end{table}
Firstly, controllable generation relies on a high-performance text-to-motion pre-trained model \cite{smoodi, omnicontrol, motionlcm}. It bridges the semantic gap between language and motion at word-level \cite{momask, graphmotion} granularity. We propose to {\bf decouple text and motion} modules to achieve a framework called {\it Attribute Diffusion Model (ADM)}, adopting continuous diffusion \cite{mld} rather than discrete autoregressive approaches \cite{motiongpt,momask} for controlled generation. Three key innovations drive us:
1) We believe that decoupling using cross attention \cite{huang2024stablemofusion, sun2024alpha, li2022blip} is superior to feature concatenation \cite{mld}, which provides better alignment.
2) However, the effect of cross attention is diminished in an over-compressed variational autoencoder (VAE). We attribute this to the inadequate motion representations due to the high performance of some vector-quantized representations \cite{motiongpt,momask} in recent years. Therefore, we provide a more realistic motion VAE to preserve critical kinematic details. 3) The text encoder uses T5 \cite{motionlcm, ni2021sentence} instead of CLIP \cite{clip}, as CLIP is object-based and struggles with actions and attributes. Experimental validation on HumanML3D \cite{T2M} demonstrates that decoupling text-motion learning enhances performance for high-quality motion generation (FID reduced from 0.473 to 0.102), with inherent MultiModality benefits (2.614 exceeds 1.241).\par
Secondly, there is a precision paradox in text-motion alignment. Precise models are entrenched in the training data distribution, confined to narrow textual patterns, restricting their capacity for open-instruction generalization. To address this, we propose an {\it LLM Planner} that mediates this generalization gap through reasoning, bridging dataset-specific text and unseen motion descriptions. It functions as an independent module to {\bf decouple the text reasoning process}. Specifically, we implement three components: 1) Chain-of-Thought prompting \cite{liu2024plan, zhou2024avatargpt} for reasoning LLM ({\it e.g.} Deepseek-R1) with a high-frequency word knowledge, and 2) A zero-shot Motion {\underline{I}}nference {\underline{T}}ext dataset called {\it MotionIT} featuring 5K high-level goal-driven motion descriptions for evaluation LLM Planner. 3) we developed a new evaluation system using LLMs. Thus, the planner enables user-friendly text-to-motion generation. It serves as a bridge for user interactions with model instructions, enhancing the model’s understanding of the desired actions.\par
Lastly, we {\bf decouple styles and trajectories}, {\it i.e.} kinematics control signals, for controllable stylized generation. 1) Style: let the model 'see' the new {\it motion patterns} ({\it e.g.} motion style and new actions) and learn it. We attempts to simplify enabling motion prompts \cite{ipadapter} for text-to-motion diffusion models. Thus, we design a {\it Motion Adapter} with motion prompt input, which can specify the required motion generation. The core idea is to allow the model to retain the original pattern knowledge and learn new motion patterns through {\bf decoupling training}. We propose to reuse the decoupling modules of pre-trained model and fine-tune its corresponding modules to learn new motion patterns. This means that our motion prompt concretely generates semantics in the finetuning data. {\it This is a novel way to fine-tune new motion patterns, and very lightweight and fast training (15-20 minutes)}. We achieved great stylized motion generation on 100STYLE \cite{mason2022real}. 2) Trajectory: following \cite{omnicontrol, motionlcm}, we adopted controlnet for trajectory control. Our focus is on verifying the compatibility between the motion adapter and ControlNet to achieve multi-attribute control.\par
In sum, our work has the following main contributions: 1) To the best of our knowledge, this is the first work to propose decoupling any condition and separately implementing multi-attribute text-to-motion generation. 2) The Motion Adapter is a lightweight, efficient fine-tuning method for multimodal motion generation, which achieves fine-tuning state-of-the-art on 100STYLE. 3) We leveraged LLMs to enable user-friendly text-to-motion generation, capable of handling previously unseen text inputs. 4) The Attribute Diffusion Model achieves performance on par with the state-of-the-art latent diffusion model on HumanML3D.

%% file: sec/2_relatedwork.tex
\section{Related Work}
\indent \indent {\bf Motion Generation} generates 3D human motion sequences from multi-modal inputs, including image \cite{jiang2024motionchain,wei2025learning}, text \cite{T2M, mld, motiongpt, momask, mdm, remodiffuse}, action labels \cite{cai2018deep,action2motion,petrovich2021action}, audio \cite{li2021ai,gong2023tm2d}, control signals \cite{omnicontrol, motionlcm}, and incomplete pose sequences \cite{harvey2020robust,liu2022investigating}. In this paper, we focus on text and control signals, with further distinction made between the fine-grained \cite{motionclr, como, graphmotion} content in the text and the trajectory and style \cite{jang2022motion,aberman2020unpaired,smoodi} information in the control signals.\par
\begin{figure*}[htb]
	\begin{minipage}[b]{1.0\linewidth}
		\centering
		\centerline{\includegraphics[width=18cm]{fig/Method.pdf}}
		\medskip
	\end{minipage}
	\caption {The {\it ACMo} network architecture. Stage 1: {\it Attribute Diffusion Model} is trained by decoupling text and motion in a more powerful latent space. Stage 2: {\it Motion Adapter} finetunes new motion patterns and preserves the original knowledge. Stage 3: Trajectory control through Controlnet. Finally, the LLM Planner module inferences for text processing.}
	\label{fig:network architecture}
\end{figure*}
{\bf Text-to-Motion generation} has gained significant attention in recent years. The mainstream approaches now learn shared latent representations to directly convert text into motion \cite{action2motion}. T2M \cite{T2M} applied sequential VAEs \cite{kingma2013auto} to learn the probabilistic relationship between motion and text. {\it Diffusion models} \cite{ho2020denoising, song2020denoising} have significantly advanced this field, with MotionDiffuse \cite{motiondiffuse} introducing the first text-based diffusion model. MDM \cite{mdm} learned the mapping between text and motion in the original motion space. ReMoDiffuse \cite{remodiffuse} enhanced its performance by improving retrieval techniques.
MLD \cite{mld} proposed learning the text-to-motion mapping in latent space, reducing noise and redundancy in long sequences. Additionally, some methods using VQVAE \cite{van2017neural} and autoregressive models \cite{como,momask,motiongpt}, such as MoMask \cite{momask} with residual codebook quantization, highlight the importance of latent motion representation. Following the previous discussion, we further explore the text-to-motion performance of the latent diffusion model.\par
{\bf Attribute Motion Control} is implemented using a diffusion model, which has recently attracted attention due to its exceptional ability to control. ControlNet \cite{controlnet} enhances generative models by incorporating additional control inputs. Similarly, IP-Adapter \cite{ipadapter} improves domain transferability through lightweight adapter layers. In motion generation, OmniControl \cite{omnicontrol} used spatial guidance and motion ControlNet to align generated motion with input trajectory signals. MotionLCM \cite{motionlcm} introduced latent space consistency constraints to improve motion continuity. Our work focuses on seamlessly integrating trajectory control and motion style control.
{\it Motion style} is elusive and hard to define mathematically, making data-driven approaches suitable for control and transfer. For instance, Aberman et al. \cite{aberman2020unpaired} designed a two-branch pipeline to decouple and recombine motion styles, while Guo et al. \cite{guo2024generative} extract and infuse motion content and style in latent space. SmooDi \cite{smoodi} introduced style guidance and style ControlNet to direct motion, and MotionCLR \cite{motionclr} used an attention mechanism to edit motions without training. Compared to previous works, our approach conditions on motion style for multimodal control and uses lightweight adapters to refine the model.\par
{\bf LLM Planner for Motion Generation} has gained increasing interest in recent years due to advances in large language models (LLMs). MotionGPT \cite{motiongpt} finetuned T5 for unified tasks but lost text knowledge. ChatPose \cite{chatpose} was fine-tuned with reasoning generation but underperformed compared to specialized models. An AI agent \cite{lan2025visual,liu2024plan, como, zhou2024avatargpt} to guide the LLM planner and integrate high-performance generative models can address these issues. For example, Xiao et al. \cite{xiao2024unified} and Zhou et al. \cite{zhou2024avatargpt} used LLM as task planners for task decomposition. CoMo \cite{como} uses ChatGPT to get fine-grained text on different body parts. Inspired by this, we use the LLM to convert unseen text into training dataset-relevant text, achieveing reasoning generation.

%% file: sec/3_method.tex
\section{Method}
We aim to address controllable motion generation with multimodal inputs. To generate a 3D pose sequence $\hat{m}_{1:L}$ of length $L$, with a motion prompt $p_{1:N}$ of length $N$, text condition $c$, and trajectory signal $s \in \mathbb{R}^{L \times J \times 3}$, we designed a novel attention-based motion latent diffusion model, as shown in Fig. \ref{fig:network architecture}. It decouples text, style, and trajectory training across the network: first, training the text-to-motion Attribute Diffusion Model, then finetuning the Motion Adapter for stylized motion prompt inputs, and finally integrating Trajectory ControlNet for spatial signal control.
\subsection{Training: Attribute Diffusion Model}
\label{sec:adm}
Inspired by \cite{mld, motionclr, huang2024stablemofusion, momask}, we designed the Attribute Diffusion Model (ADM) in the latent space. This model leverages a transformer-based architecture \cite{mld, petrovich2021action}, consisting of a Motion Variational AutoEncoder, text embeddings, and the Attribute Latent Diffusion Model. \par
\par
{\bf Motion Variational AutoEncoder} consists of a motion encoder $\mathcal {E}$ and a motion decoder $\mathcal{D}$, trained solely on motion sequences $m_{1:L}\in \mathbb{R}^{L \times H}$ without text, where $H=263$ denotes the dimension of motion representation. We adopted a transformer-based architecture in the MLD \cite{mld} for Motion VAE, adjusting the network parameters to enhance the codec's performance. This VAE codec has lower computational overhead and produces more diverse motions. Specifically, the low-dimensional vectors are in motion latent space $z = \mathcal{E}(m)$, then reconstructed by decoder $\hat{m} = \mathcal{D}(z)$. The Mean Squared Error (MSE) loss and the Kullback-Leibler (KL) divergence are used as losses for reconstructing the training motion.
\par
{\bf Text Embedding} maps textual descriptions to the latent space. Most previous work \cite{mld, momask, como} relied on CLIP as a text encoder, but its focus on object-level text limits fine-grained property descriptions. Instead, we used T5 \cite{ni2021sentence} for text embedding $\tau$, achieving improved performance.
\par
{\bf Attribute Latent Diffusion Model} aims to explore a more powerful motion diffusion model in latent space. The diffusion probabilistic model \cite{ho2020denoising} transforms Gaussian noise into a data distribution $p(x)$. Its forward process is a Markov chain that gradually adds noise over $k$ steps, while the reverse process anneals the noise by modeling the conditional noise $\epsilon$. Specifically, the process initializes with a latent vector $z_0\in \mathbb{R}^{L\times D}$ from Motion VAE. Through forward diffusion, noise is systematically introduced across 
$k$ time steps, progressively transforming $z_0$ into a noisy latent state $z_k$. This formulation permits closed-form sampling at arbitrary step $k$ via:
\begin{equation}
	q (z_k |z_{k-1}) := \mathcal{N}(z_k;\sqrt{{\alpha}_k} h_0,(1-\alpha_k)I),
\end{equation}
where ${\alpha}_k\in(0,1)$ denotes a hyper-parameter for sampling.
The reverse diffusion step is expressed by the trained diffusion model $g$ and its parameters $\theta$ as follows:
\begin{equation}
	z_{k-1} = g_\theta(z_k, f_t), \quad k \in \{1, \dots, K\},
\end{equation}
where $f_t$ represents the conditional text feature at each diffusion step based on the predicted noise $\epsilon \sim \mathcal{N}(0, 1)$.\par 
Our denoiser $g_\theta$ (parameterized by $\theta$) builds upon MLD \cite{mld} and all layers with long skip connections \cite{bao2022all}. We designed the text-motion block to implement our denoiser using stacked self-attention and cross-attention.
{\it Self Attention} layers model the relationship between motions themselves without text condition \cite{motionclr,mld, mdm}:
\begin{equation}
z=Attn(q,k,v) = \sigma({{q}{k}^T}/{{\sqrt{d}}})v,
\label{eq:selfAttn}
\end{equation}
where $q$, $k$, and $v$ represent the query, key, and value, respectively, all corresponding to the motion modality, and $\sigma(\cdot)$ denotes the softmax function.
{\it Cross Attention} layers model the relationship between different decoupled multimodal data in latent space. In text-to-motion tasks, they model the correlation between text and motion sequences \cite{huang2024stablemofusion}. It can be formulated as:
\begin{equation}
	CrossAttn(q,k',v') = \sigma	({{q}{k'}^T}/{{\sqrt{d}}})v',
\label{eq:CrossAttn}
\end{equation}
where $q$ represents the text feature, and $k'$ and $v'$ are the motion latent vectors.\par

\subsection{LLM Planner}
LLM Planner serves as an independent plug-and-play module in our framework. The planner enhances text descriptions via zero-shot text inference. We tried different large language models in Sec. \ref{sec:llmAblation}, both easy to deploy and commercially available. Additional details about the prompt are provided in the supplementary.\par
Inspired by the proposed precision paradox, we made a wordbag as local knowledge. {\it The higher the frequency of words in the dataset, the better the alignment performance.} We constructed the proposed local knowledge base by relying on TF-IDF \cite{salton1988term} to calculate the importance of all texts in the dataset, and then selected the most important 512 words as local knowledge. This allows the model to map unseen text to text in the dataset. Finally, we have developed an evaluation system for assessing the results of LLM Planner by leveraging another LLM as an evaluator in Sec. \ref{sec:metrics}.
\par
\subsection{Fineture: Motion Prompt Adapter \& Trajectory ControlNet}
Inspired by \cite{motionclr, ipadapter, smoodi, omnicontrol}, we propose a lightweight finetuning method to learn new motion patterns. Our model is characterized by the fact that we reuse the decoupling module of ADM and finetune its corresponding modules to learn new motion patterns.\par
\begin{figure}[htbp]
	\centering
	\includegraphics[width=0.48\textwidth]{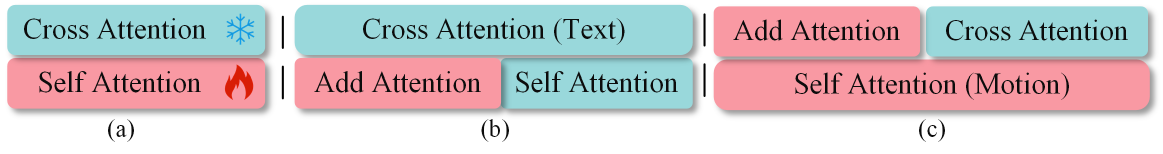}
	\caption{To achieve retaining text knowledge, we froze decoupled cross attention. Three fine-tuning methods are carried out, and the method (c) achieved the most efficient learning of motion patterns.}
	\label{fig:decoupleAttention}
\end{figure}

{\bf Decoupled Fineture.} Because different motions are new patterns with the same representation. We freeze and reuse motion VAE in the fine-tuning process, unlike other motion style training methods \cite{smoodi}. In Sec. \ref{sec:MotionStyleAndTraj}, we designed experiments to verify that the decoupling method (c) in Fig. \ref{fig:decoupleAttention} can efficiently learn the motion patterns with a small amount of data and training while retaining the text knowledge.\par
{\bf Decoupled Additional Text Cross Attention.} In order to preserve the original knowledge, we used a new cross attention to complement the new textual and visual mapping. We train the decoupled cross attention in Fig. \ref{fig:decoupleAttention} and then add it to the original cross attention \cite{ipadapter}. It can be formulated as:
\begin{equation}
	z^{new} = \sigma	({{q}{k}^T}/{{\sqrt{d}}})v+ \sigma({{q}{k'}^T}/{{\sqrt{d}}})v',
\end{equation}
where $ q $ represents the text feature, $ k' $ differs from $ k $ by $ k = zW $ and $ k' = zW' $, with $ W' $ being the weight matrix of the additional attention. The same applies to $ v $.\par
{\bf Self-to-Cross Attention.} Inspired by \cite{motionclr, ipadapter}, to achieve a function similar to motion style transfer, we propose to utilize motion prompts for multimodal stylized text-to-motion generation. With motion prompts, fine-grained and specific stylized results can be obtained within the specified text range. As shown in Stage 2 of Fig. \ref{fig:network architecture}, we added a linear layer and layer norm on Fig. \ref{fig:decoupleAttention} (c) to linearly transform the latent vectors. Self-to-cross attention refers to turning a self attention query in Eq. \ref{eq:selfAttn} into a motion prompt to achieve cross attention. This can be formulated with Eq. \ref{eq:CrossAttn}, where q is a motion prompt latent vector.\par
{\bf  Trajectory ControlNet.}  We build and only use trajectory  ControlNet to address the trajectory control, following \cite{motionlcm,omnicontrol}. Our core goal is to verify the compatibility of the trajectory controlnet with our motion adapter, so we set the hip controls for simplicity. We achieve this by copying denoiser $g$ in Stage 1 and adding it with a zero-initialized linear layer for eliminating random noise. Trajectory control is learning by training the trajectory encoder and ControlNet. Details are provided in the supplementary.

%% file: sec/4_experiments.tex
\section{Experiment}
\begin{table*}[t]
	\centering
	\caption{Comparison of text-conditional motion synthesis on HumanML3D. We evaluate each metric 20 times. "↑" indicates higher is better, "↓" indicates lower is better, and "→" indicates closer to real motion is optimal. All methods use the ground truth motion length and are sorted by FID. {\bf\textcolor{deepred}{Bold}} and {\it\textcolor{blue}{italic}} denote the best and second best results, respectively. (†) denotes a diffusion-based method.  MLD is our baseline for ablation. We ablated the effectiveness of the entire module with 9 layers, achieved optimal performance with 11 layers. The proposed ADM achieve {\it comparable FID and R Precision}, with {\bf MModality} being a notable advantage. The proposed method is of particular significance as it is not only an accurate improvement, but also inherently {\bf controllable} in latent space.}
	\resizebox{1.95\columnwidth}{!}{
		\begin{tabular}{lccccccc}
			\toprule
			\multirow{2}{*}{Methods} & 
			\multicolumn{3}{c}{R Precision↑} & \multirow{2}{*}{FID↓} & \multirow{2}{*}{MM Dist↓} &
			\multirow{2}{*}{Diversity→} &
			\multirow{2}{*}{MModality↑} \\
			\cline{2-4} \rule{0pt}{10pt} 
			& {Top 1} & {Top 2} & {Top 3} & &  & \\
			\hline
			\rule{0pt}{10pt}
			Real & 0.511$^{\pm .003}$ & 0.703$^{\pm .003}$ & 0.797$^{\pm .002}$ & 0.002$^{\pm .000}$ & 2.974$^{\pm .008}$ & 9.503$^{\pm .065}$ & - \\
			\hline
			T2M \cite{T2M} & 0.457$^{\pm .002}$ & 0.639$^{\pm .003}$ & 0.740$^{\pm .003}$ & 1.067$^{\pm .002}$ & 3.340$^{\pm .008}$ & 9.188$^{\pm .002}$ & 2.090$^{\pm .083}$ \\
			MotionDiffuse (†) \cite{motiondiffuse} & 0.491$^{\pm .001}$ & 0.681$^{\pm .001}$ & 0.782$^{\pm .001}$ & 0.630$^{\pm .001}$ & 3.113$^{\pm .001}$ & 9.410$^{\pm .049}$ & 1.553$^{\pm .042}$ \\
			MDM (†) \cite{mdm} & 0.418$^{\pm.005}$ & 0.604$^{\pm.005}$ & 0.707$^{\pm.004}$ & 0.489$^{\pm.025}$ & 3.630$^{\pm.023}$ & {\it \textcolor{blue}{9.450}$^{\pm.066}$} & {\bf \textcolor{deepred}{2.870}$^{\pm.111}$}  \\
			MotionLCM (†) \cite{motionlcm} & 0.502$^{\pm .003}$ & {\it \textcolor{blue}{0.698}$^{\pm .002}$}& {\it \textcolor{blue}{0.798}$^{\pm .002}$} & 0.304$^{\pm .012}$ & 3.012$^{\pm .007}$ & 9.607$^{\pm .066}$ & 2.259$^{\pm .092}$ \\
			CoMo \cite{como} & 0.502$^{\pm .002}$ & 0.692$^{\pm .007}$ & 0.790$^{\pm .002}$ & 0.262$^{\pm .004}$ & 3.032$^{\pm .015}$ & 9.936$^{\pm .066}$ & 1.013$^{\pm .046}$ \\
			MotionGPT \cite{motiongpt} & 0.492$^{\pm .003}$ & 0.681$^{\pm .003}$ & 0.778$^{\pm .002}$ & 0.232$^{\pm .008}$ & 3.096$^{\pm .008}$ & {\bf \textcolor{deepred}{9.528}$^{\pm .071}$} & 2.008$^{\pm .080}$ \\
			ReMoDiffuse (†) \cite{remodiffuse} & {\it \textcolor{blue}{0.510}$^{\pm .005}$} & {\it \textcolor{blue}{0.698}$^{\pm .006}$} & 0.795$^{\pm .004}$ & 0.103$^{\pm .004}$ & 2.974$^{\pm .016}$ & 9.018$^{\pm .075}$ & 1.795$^{\pm .043}$ \\
			MoMask \cite{momask} & {\bf \textcolor{deepred}{0.521}$^{\pm .002}$} & {\bf \textcolor{deepred}{0.713}$^{\pm .002}$} & {\bf \textcolor{deepred}{0.807}$^{\pm .002}$} & {\bf \textcolor{deepred}{0.045}$^{\pm .002}$} & {\bf \textcolor{deepred}{2.958}$^{\pm .008}$} & 9.648$^{\pm .066}$ & 1.241$^{\pm .040}$ \\
			\hline
			\rule{0pt}{10pt}
			MLD (†) \cite{mld} & 0.481$^{\pm .003}$ & 0.673$^{\pm .003}$ & 0.772$^{\pm .002}$ & 0.473$^{\pm .013}$ & 3.196$^{\pm .010}$ & 9.724$^{\pm .082}$ & 2.413$^{\pm .079}$ \\
			\hline
			\rule{0pt}{10pt}
			Ours (ADM)& 0.493$^{\pm .002}$ & {\it \textcolor{blue}{0.698}$^{\pm .003}$}& 0.795$^{\pm .002}$ & {\it \textcolor{blue}{0.102}$^{\pm .003}$} & {\it \textcolor{blue}{2.973}$^{\pm .006}$} & 9.749$^{\pm .082}$ & {\it \textcolor{blue}{2.614}$^{\pm .100}$}\\
			\hline
			\midrule
			+ VAE-7 & 0.473$^{\pm .004}$ & 0.667$^{\pm .003}$ & 0.769$^{\pm .003}$ & 0.305$^{\pm .011}$ & 3.159$^{\pm .009}$ & 9.663$^{\pm .072}$ & 2.460$^{\pm .104}$ \\
			+ VAE-7 \& T5 & 0.478$^{\pm .002}$ & 0.680$^{\pm .003}$ & 0.785$^{\pm .002}$ & 0.265$^{\pm .009}$ & 3.086$^{\pm .008}$ & 9.827$^{\pm .089}$ & 2.447$^{\pm .099}$ \\
			+ VAE-7 \& T5 \& CrossAttn & 0.486$^{\pm .003}$ & 0.687$^{\pm .003}$ & 0.791$^{\pm .003}$ & 0.122$^{\pm .005}$ & 3.044$^{\pm .011}$ & 9.670$^{\pm .070}$ & 2.532$^{\pm .095}$
			\\
			\hline
		\end{tabular}
	}
	\label{tab:humanml3d_comparison}
\end{table*}
\indent \indent {\bf Dataset.} First, we train and evaluate the proposed Attribute Diffusion Model on the popular text-to-motion dataset, {\it HumanML3D} \cite{T2M}. It collects 14,616 motion sequences and each motion was annotated with text from AMASS \cite{AMASS} and HumanAct12 \cite{guo2020action2motion} . Thanks to \cite{smoodi}, we use {\it 100STYLE} \cite{mason2022real} to finetune our Motion Adapter. Lastly, we present our evaluation dataset of the LLM Planner on MotionIT dataset (Sec. \ref{sec:MotionIT}).\par
{\bf Evaluation metrics} are followed as per \cite{mld,momask,motiongpt,smoodi,zhou2024avatargpt}.
\label{sec:metrics}
(1) Motion Quality: The Frechet Inception Distance (FID) compares feature distributions between real and generated motions via feature extractor \cite{T2M}. (2) Condition Matching: R-Precision (Top-1/2/3 accuracy) and Multi-modal Distance (MMDist) to assess text-motion alignment. (3) Diversity is measured by Diversity (DIV, feature variance) and MultiModality (MModality, diversity within same text input). (4) LLM Planner evaluation: Metrics such as Bleu, Rouge, and Cider assess the performance. Inspired by \cite{zhou2024avatargpt}, we propose a evaluation framework via LLM with differentiated prompts for Rule Consistency Score (RCS) and Plan Score (PS). $RCS \in \{0,1\}$ checks rule compliance, and $PS \in [1,5]$ rates planning rationality. (5) Average Response Time (ART) measures the LLM's average time per instruction.\par
{\bf Traning Detail.} Training loss and classifier-free diffusion guidance \cite{ho2022classifier} are constructed according to MLD \cite{mld}. The optimization objective of the proposed model is:
\begin{equation}
	L_{ADM} := \mathbb{E}_{g,t}	\Bigl[ ||\epsilon-g_\theta(z_t,t,c)||^2_2 \Bigr],
	\label{eq:LossADM}
\end{equation}
where $\epsilon$ denotes the noise generated in the forward process; $c$ is the conditions, {\it i.e.} texts $f_t$, motion prompt $f_m$ and trajectories $f_{\text{tra}}$; $t$ denotes the diffusion steps, we freeze the text encoder $\tau$ \cite{petrovich2022temos} and optimize the diffusion model $g$.\par
{\bf Implementation Details. }  Our denoiser $g_\theta$ in Sec. \ref{sec:adm} employs 11 layers and 4 attention heads, with GELU as the activation function. A frozen Sentence-T5-large serves as the text encoder. The latent representation has a shape of $z \in \mathbb{R}^{7,256}$.
All models are trained using the AdamW optimizer with a fixed learning rate of $10^{-4}$ and a mini-batch size of 128. Other key model configurations, such as the number of layers and guidance scale in the diffusion model, are analyzed in the ablation study (Sec. \ref{sec:admAblation}).  
Training the Attribute Diffusion Model takes approximately 48 hours, and VAE training 30 hours on 2$\times$A6000 GPUs. Training motion adapter takes {\it only 15 minutes} to train 100 epochs. Additional experimental details can be found in the supplementary materials.
\begin{figure*}[htbp]
	\centering
	\begin{tabularx}{\textwidth}{>{\centering\arraybackslash}X
		>{\centering\arraybackslash}X
		>{\centering\arraybackslash}X
		>{\centering\arraybackslash}X
		>{\centering\arraybackslash}X}
		\toprule
		Real & {\bf ADM (Ours)} & MLD & MotionLCM & MoMask\\
		\midrule
		\multicolumn{5}{c}{a man {\bf stands} up, {\bf walks} {\it{\textcolor{deepred}{clockwise} in a circle}}, then  {\bf sits} {\it \textcolor{deepred} {back down}}.} \\
		\includegraphics[width=0.8\linewidth]{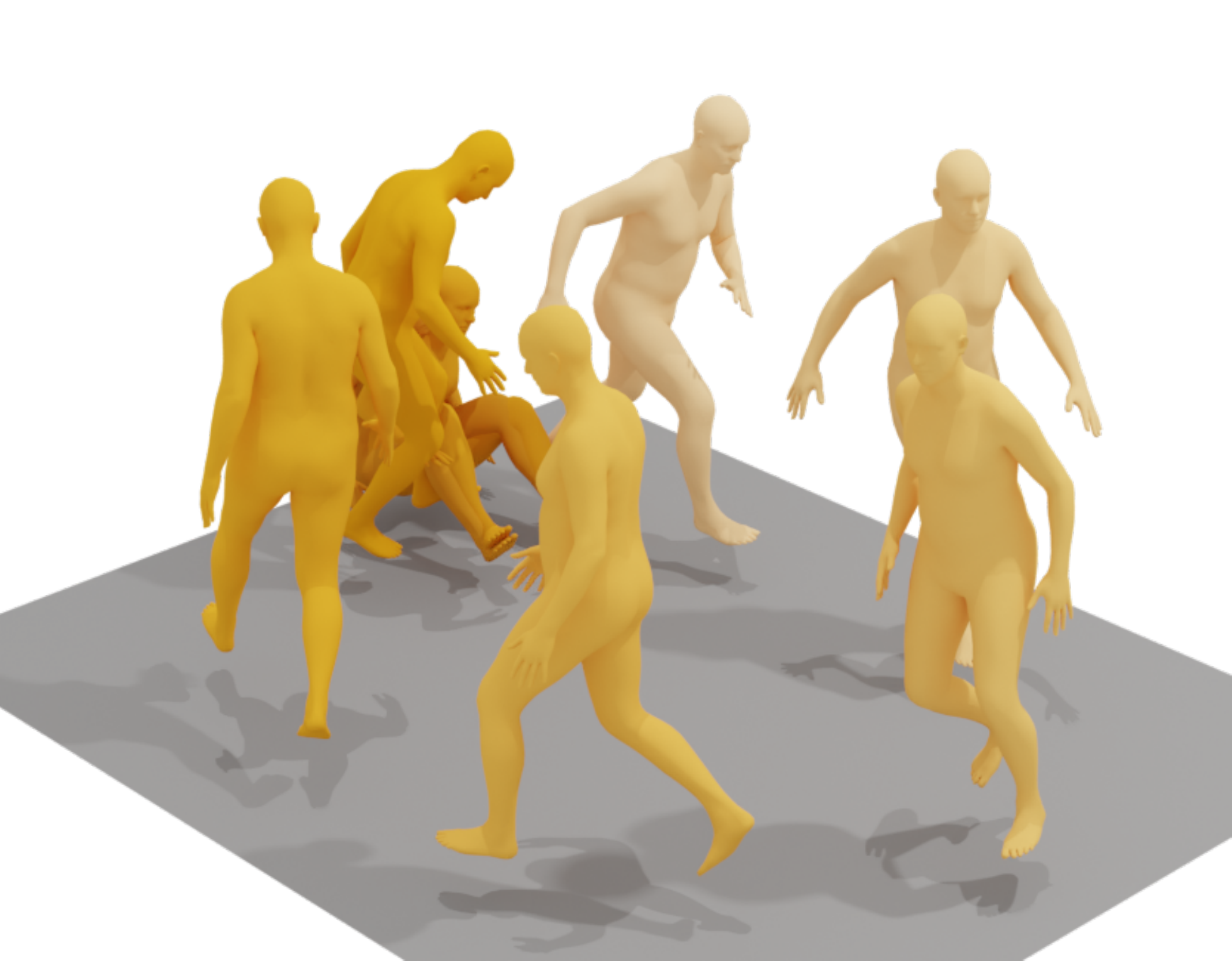} & 
		\includegraphics[width=0.7\linewidth]{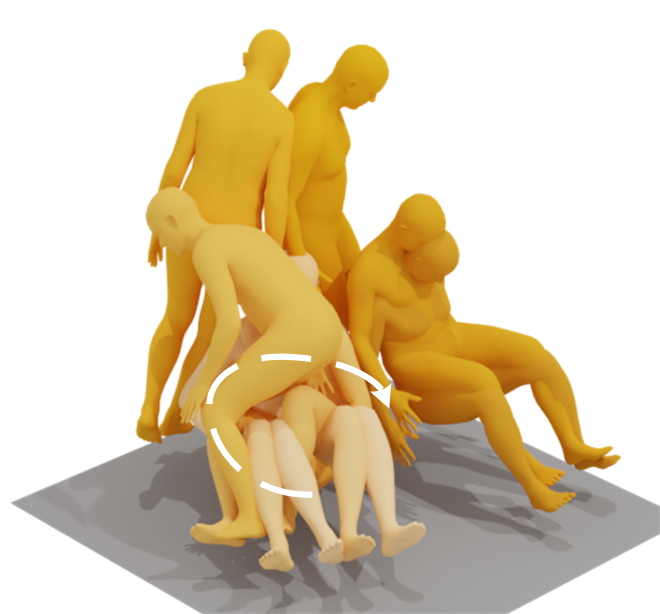} & 
		\includegraphics[width=0.7\linewidth]{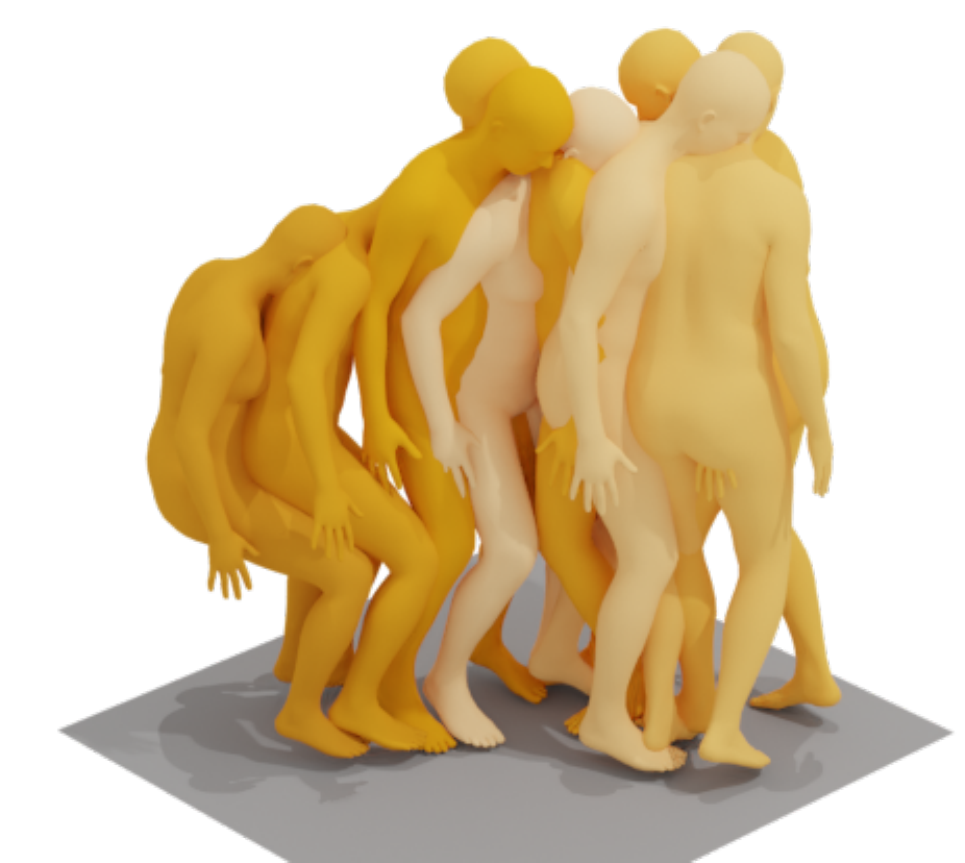} & 
		\includegraphics[width=0.8\linewidth]{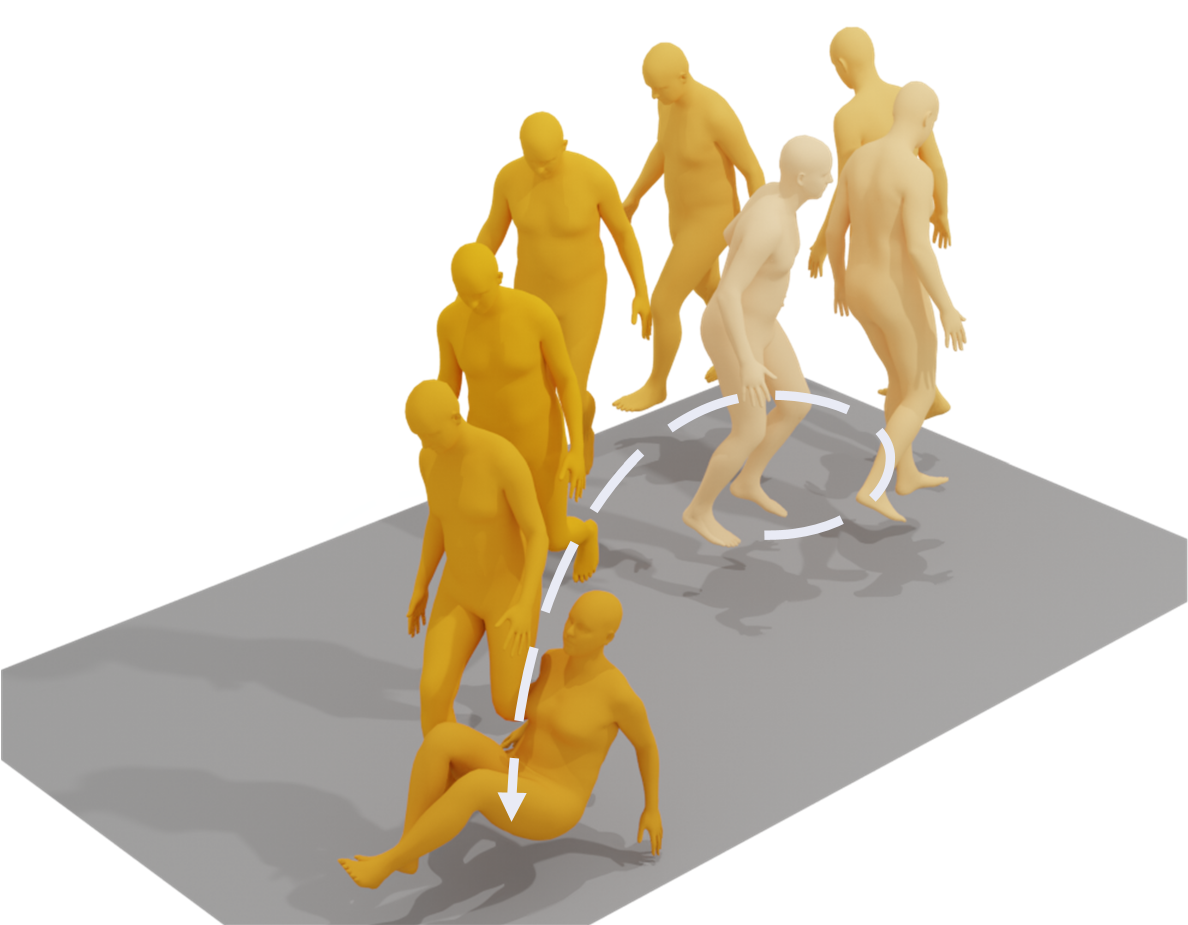} & 
		\includegraphics[width=0.8\linewidth]{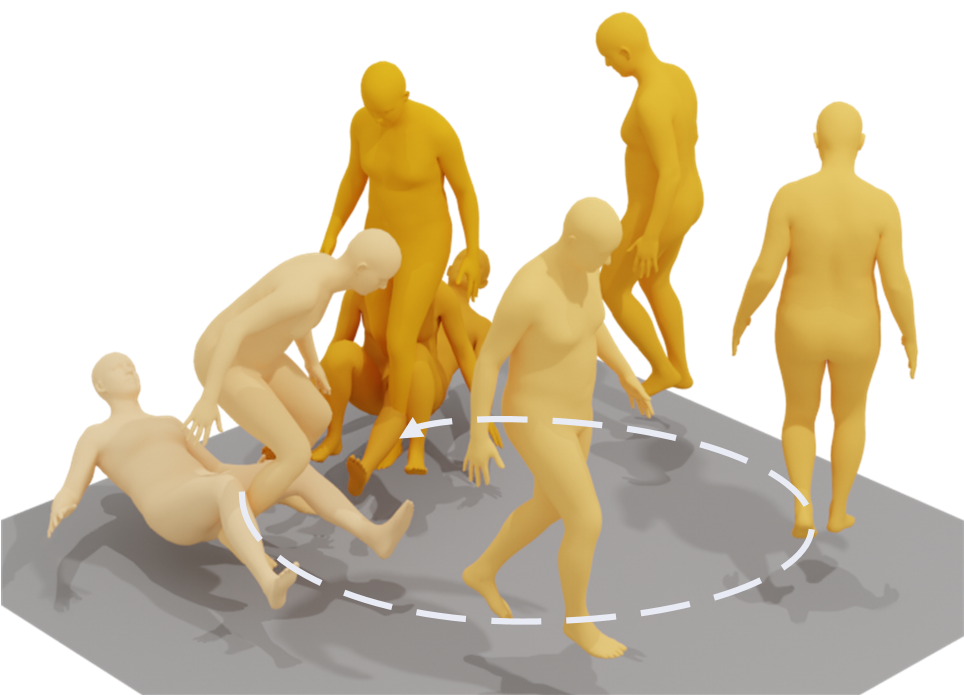}
		\\
		\hline
		\multicolumn{5}{c}{a man \textcolor{deepred}{\bf steps} {\it \textcolor{deepred}{forward vigorously}} and then {\bf kicks} something {\it low with his \textcolor{deepred}{right} leg}.}\\
		\includegraphics[width=0.8\linewidth]{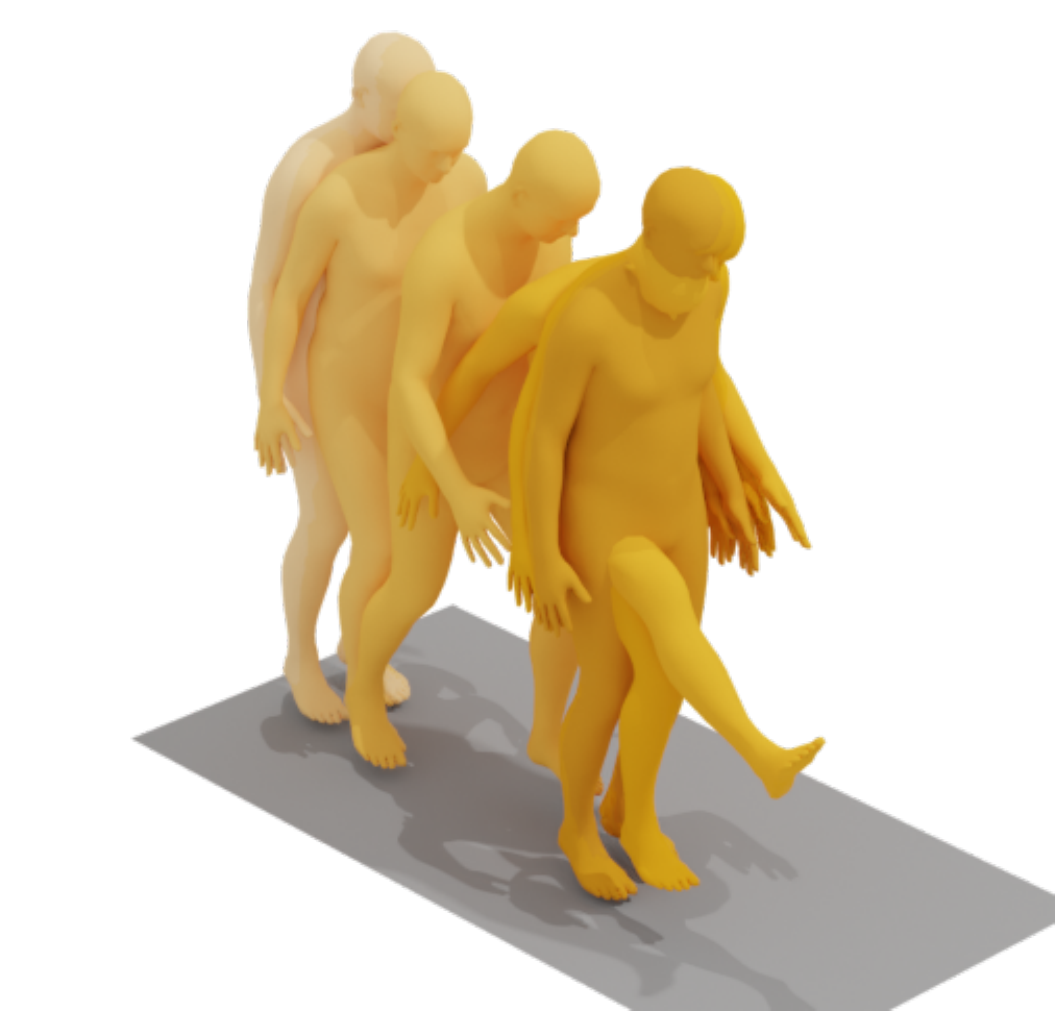} & \includegraphics[width=0.8\linewidth]{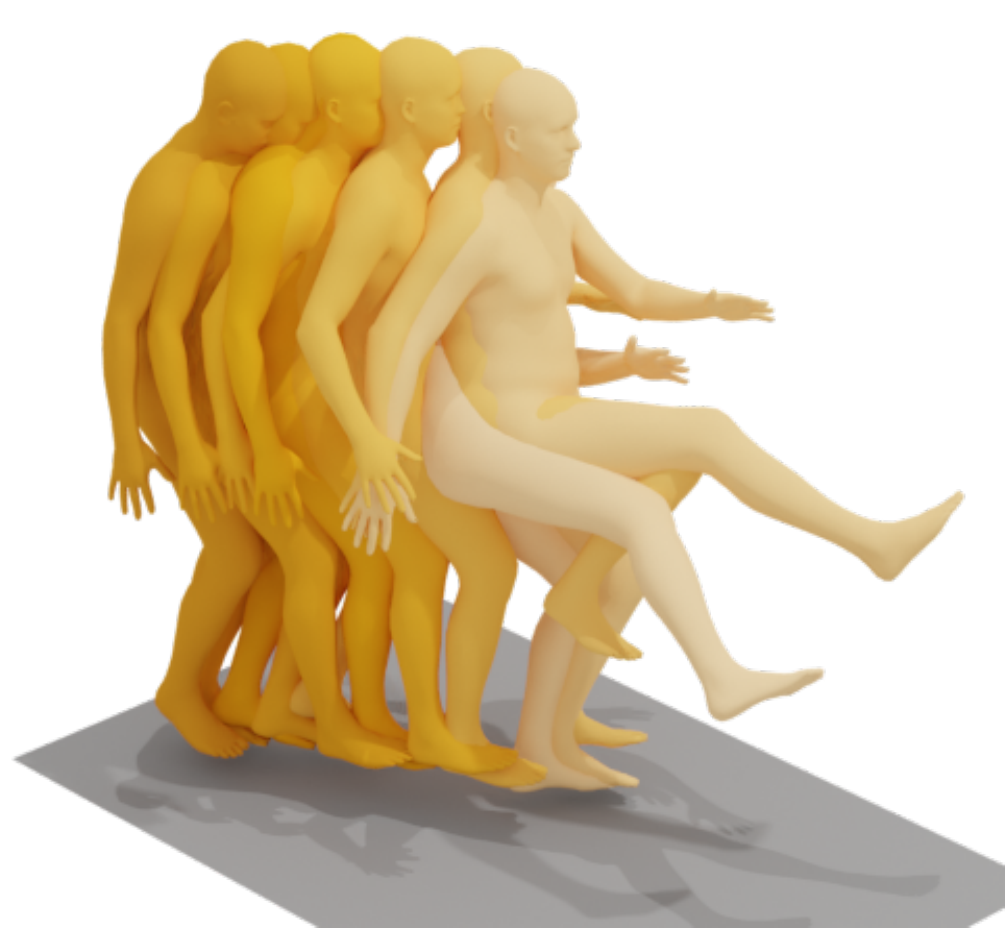} & 
		\includegraphics[width=0.7\linewidth]{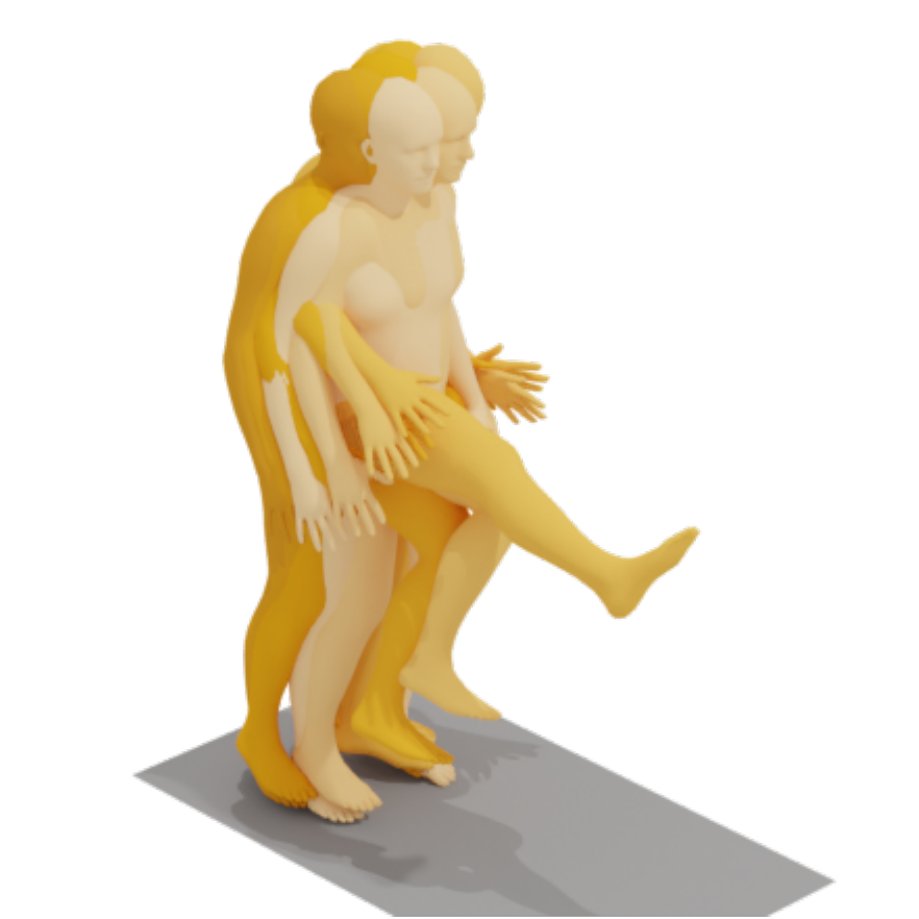} & 
		\includegraphics[width=0.8\linewidth]{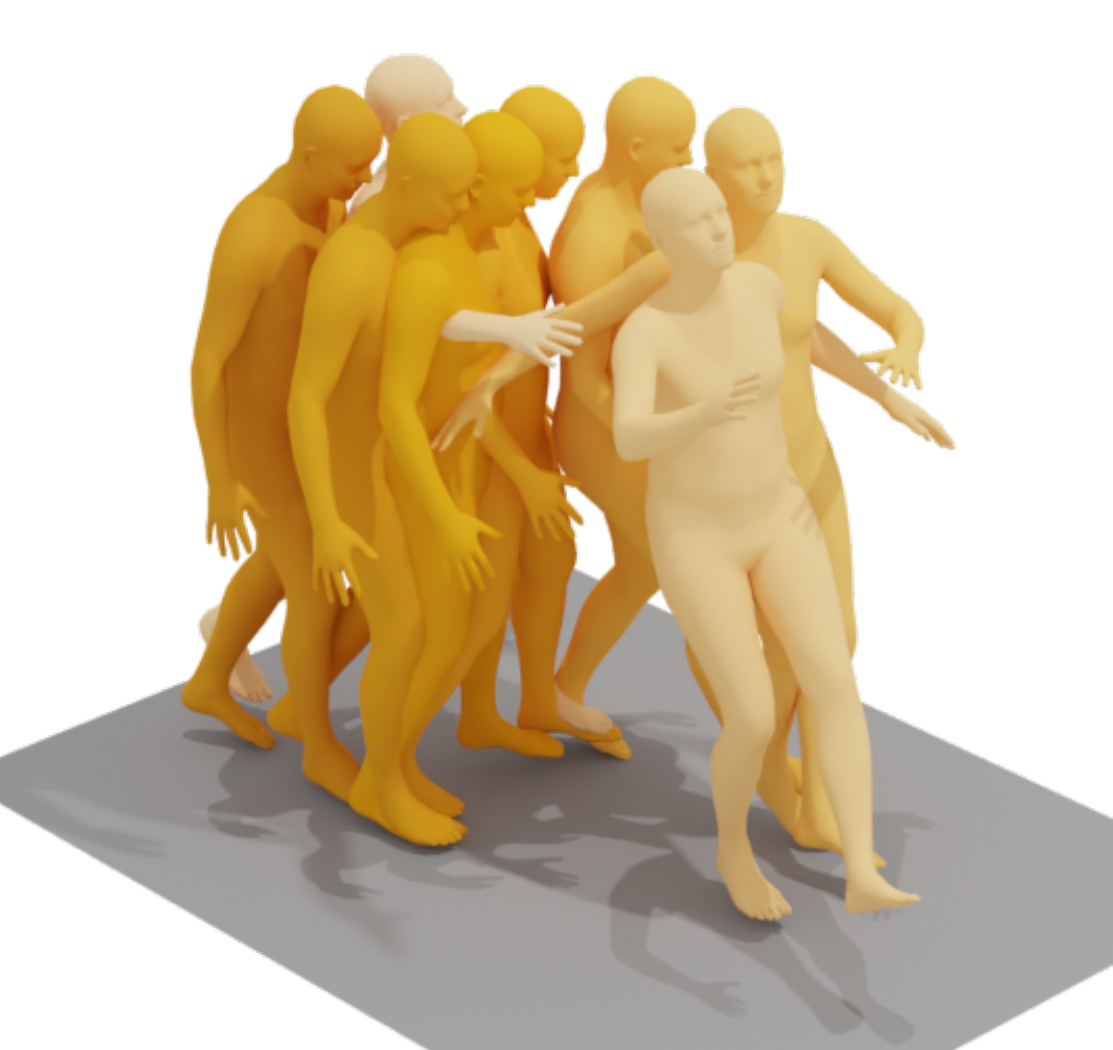} &
		\includegraphics[width=0.8\linewidth]{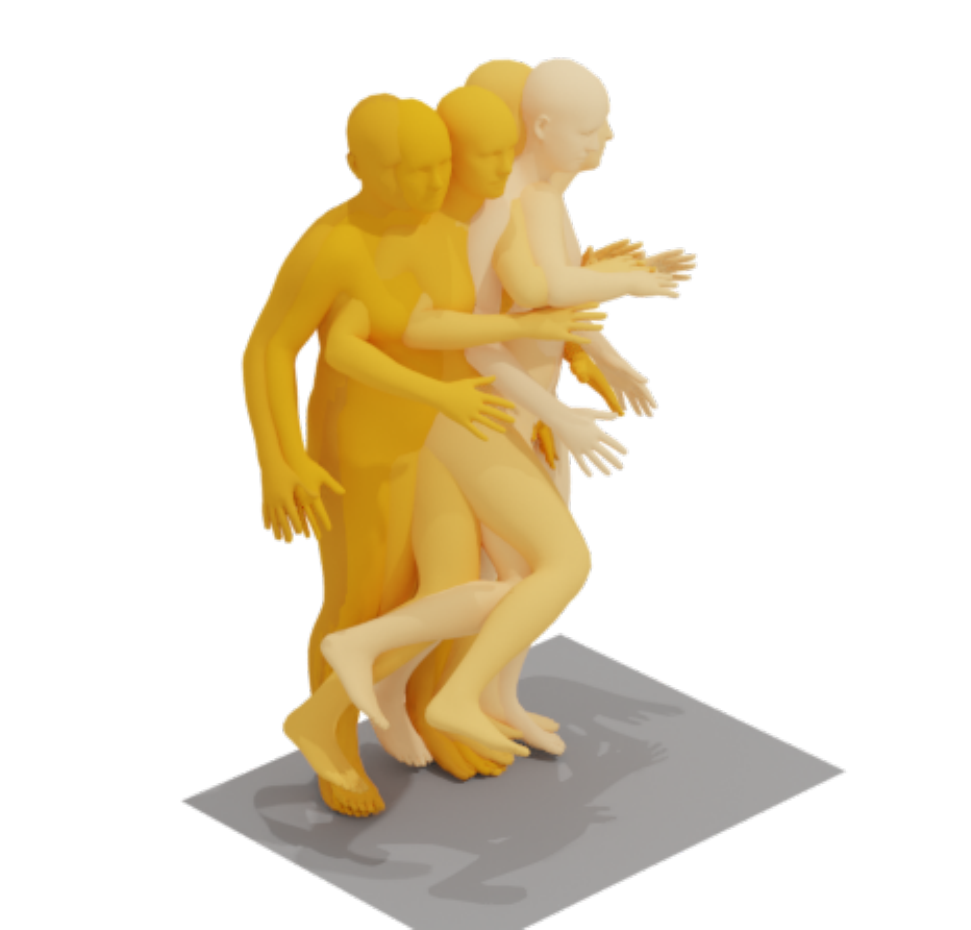} \\
		\bottomrule
	\end{tabularx}
	\caption{Visualization Comparison between the different methods given two distinct text descriptions from HumanML3D testset.  {\bf Bold} and {\it italic} denote the verb and attribute, respectively. This visualizes \textcolor{deepred}{subtle differences} at the word level, which illustrates the advantages of our approach.}
	\label{fig:VisualizationComparisonResults}
\end{figure*}
\subsection{Motion inference text dataset for LLM Planner} 
\label{sec:MotionIT}
Motion represents goal-driven human pose expressions. We created MotionIT to evaluate the LLM Planner on reasoning motion generation. It tests zero-shot inference capabilities for open-instruction requirements. The dataset includes everyday situations, emotional states, intentional goals, and time-sensitive actions, focusing on context-aware motion through scenario narratives and environmental dynamics.\par
Simple cases include "A person jogs to catch a bus pulling away from the stop." while complex ones involve reasoning about emotions or interactions, like "A person checks their bank balance, wondering if they can afford to take a financial risk." Moreover, we incorporate lengthy narrative sequences. 
While MotionIT presents challenging instructions, the LLM Planner offers a suboptimal but executable solution. \par
\subsection{Text-to-motion generation Experiment}
{\bf Comparison experiment} are evaluated as shown in Tab. \ref{tab:humanml3d_comparison}. The proposed ADM is compared the state-of-the-art methods on HumanML3D. Firstly, the improvement of our method over the baseline \cite{mld} is significant. Compared with all current diffusion model \cite{mdm,motiondiffuse,motionlcm,mld,remodiffuse} methods, our method achieves the best performance. Importantly, our method has a high MModality advantage while reducing the FID. We observed that this is also a characteristic of diffusion models and realistic VAEs. Although it is slightly inferior in performance compared to autoregressive models \cite{momask}, our method has unique controllability and style learning capabilities. In the supplementary materials regarding the comparison between VAE and existing methods, the VAE-7 we designed has achieved the best performance. {\bf Visualization} of text-to-motion can be seen in the Fig. \ref{fig:VisualizationComparisonResults}.\par
\begin{figure*}[htbp]
	\centering
	\begin{tabularx}{\textwidth}{>{\centering\arraybackslash}X
			>{\centering\arraybackslash}X
			|>{\centering\arraybackslash}X
			>{\centering\arraybackslash}X}
		\toprule
		ADM (Ours) & MoMask & ADM + LLM Planner & MoMask + LLM Planner \\
		\midrule
		\multicolumn{2}{c|}{A person {\bf \textcolor{deepred}{crawls}} {\it \textcolor{blue}{through a ditch}} to {\it \textcolor{blue}{retrieve a lost phone}}.} & \multicolumn{2}{c}{A person {\bf \textcolor{deepred}{crawls}} {\it \textcolor{blue}{forward}} on the ground {\it \textcolor{blue}{carefully}}.} \\
		\includegraphics[width=0.5\linewidth]{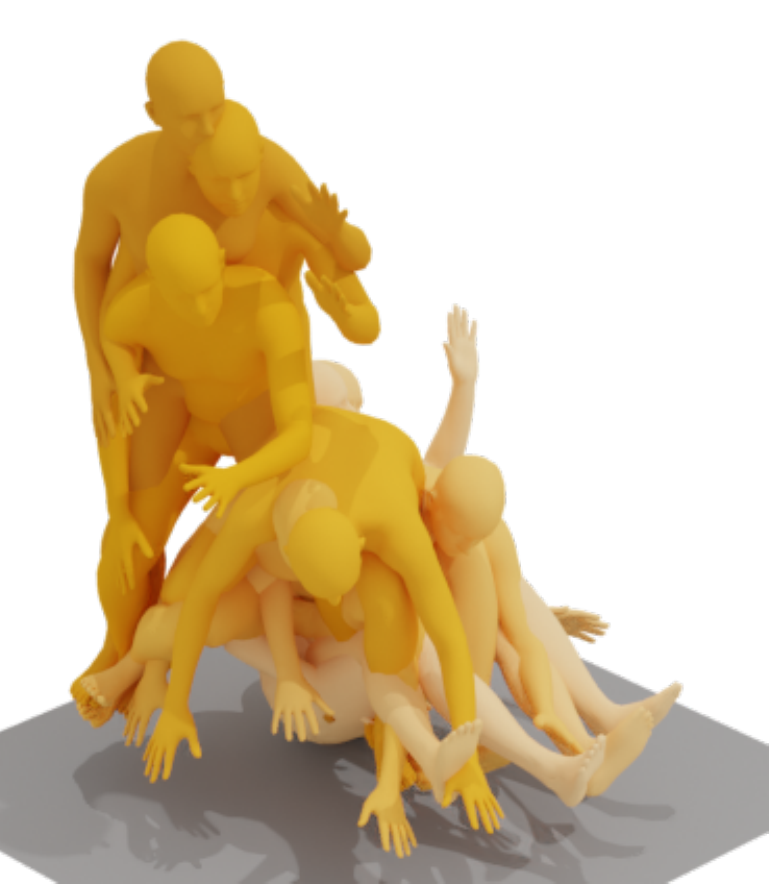} & \includegraphics[width=0.6\linewidth]{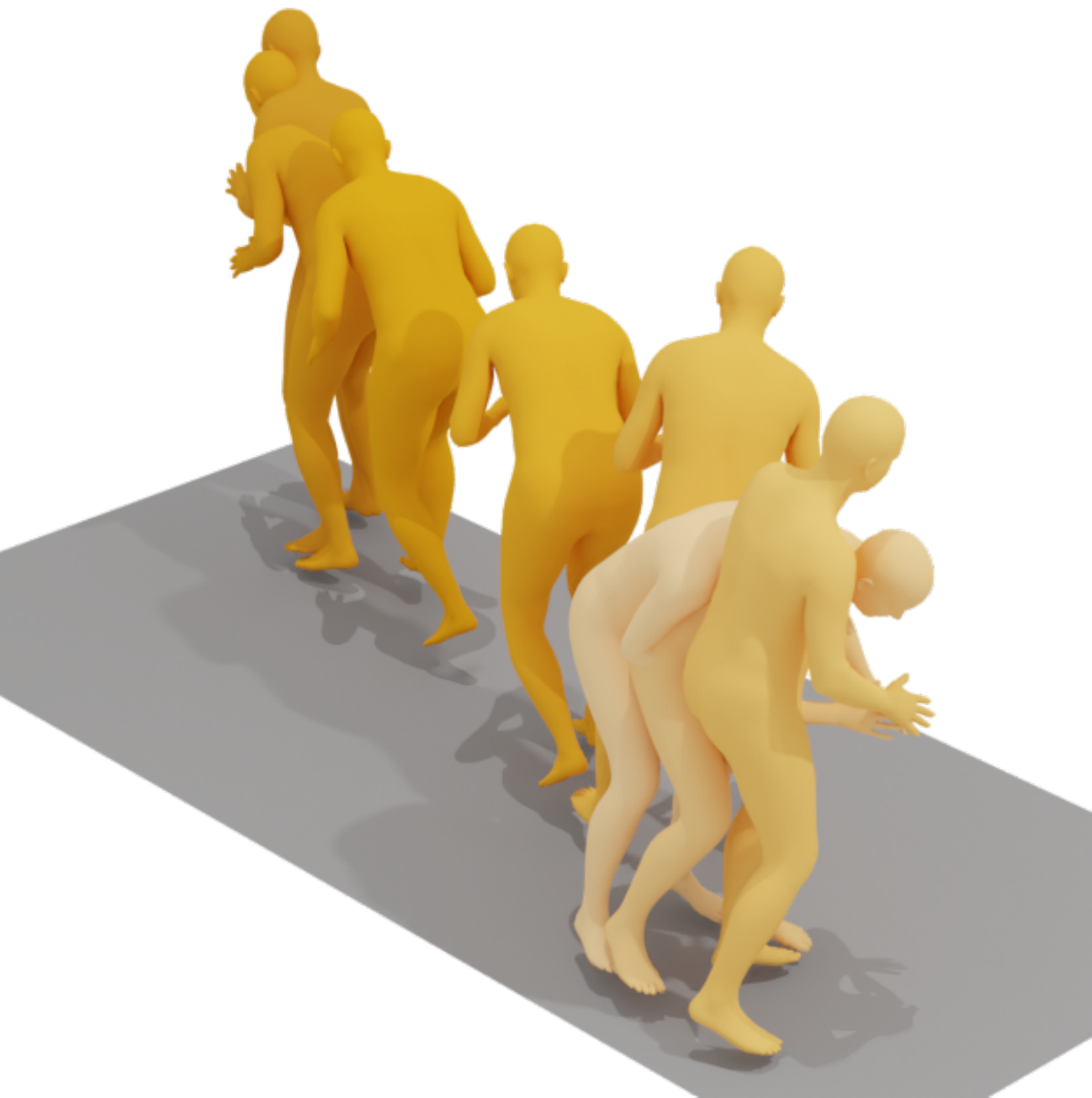} &
		\includegraphics[width=0.6\linewidth]{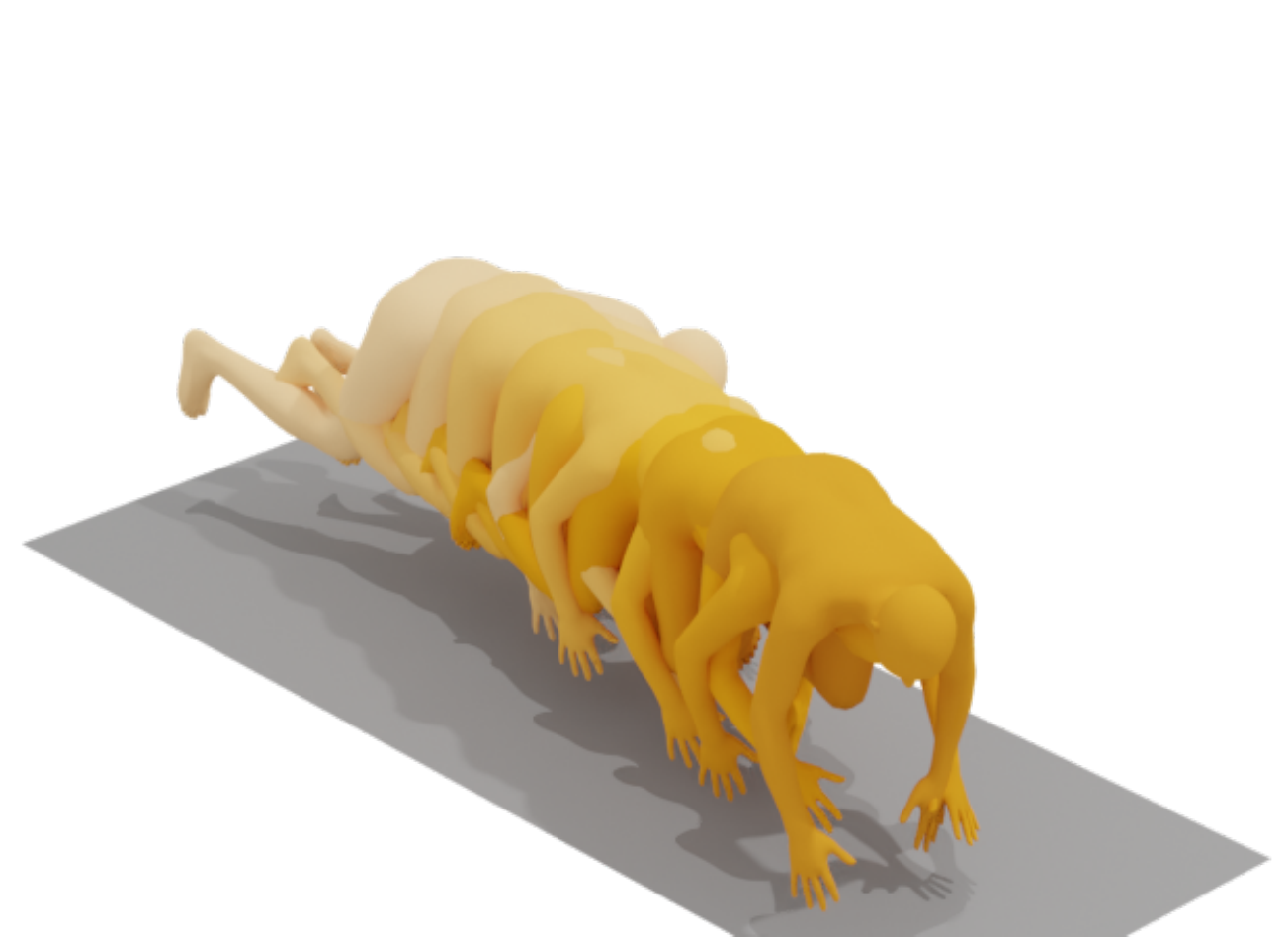} &
		\includegraphics[width=0.6\linewidth]{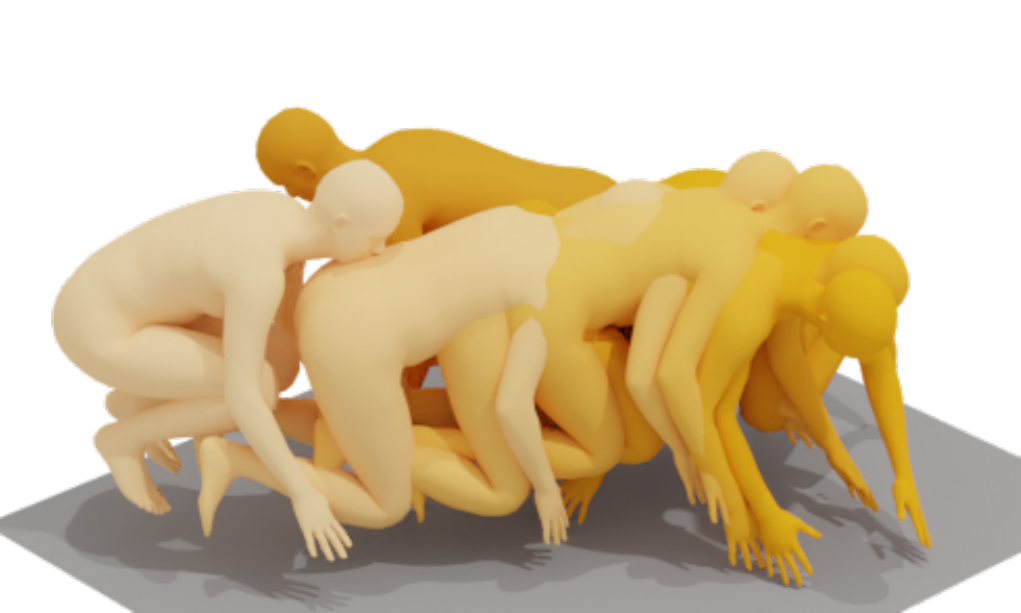} \\
		\bottomrule
	\end{tabularx}
	\caption{Visualization LLM Planner example. {\bf \textcolor{deepred}{Bold}} and {\it \textcolor{blue}{italics}} represent key verb and the transformation of the inference context, respectively. The model struggles with understanding context and gets disturbed ({\it e.g.}, 'retrieve' as a goal not a generated action) without LLM planner. The LLM understands user instructions and leverages world and local knowledge to convert into dataset text ({\it e.g.}, 'through a ditch' $->$ 'forward' and 'retrieve a lost phone' $->$ 'carefully'), enabling effective diffusion model generation.}
	\label{fig:VisualizationLLMPlannerResults}
\end{figure*}
{\bf Ablation Studies.}
\label{sec:admAblation}
{\it Component Analysis} is presented in Tab. \ref{tab:humanml3d_comparison}, which includes the ablation effects of three main components. This clearly highlights the significant performance improvements achieved through our decoupled optimization of text and motion modalities. A more realistic VAE ({\it i.e.} VAE-7) and a higher-performing diffusion model can achieve better alignment effects. 
{\it Cross Attention} added significantly improved performance, particularly in the powerful latent space. Therefore, we conducted ablation experiments in Tab. \ref{tab:Ablation_Cross_Attention}. Obviously, our two designs have been improved compared with no cross attention. We have verified the two designs: Stack Layers which learn motion first and then align it with text, and Blocks which learn motion relationships and align it while learning motion, which shows that our Block design is more efficient. To further understand its impact, we conducted an ablation study on its parameters, including the number of layers and guidance scale. We optimize the model for complex latent space capabilities by increasing the number of model layers. By adjusting the train-free hyperparameters, we can obtain models with different characteristics. In conclusion, the ablation experiments demonstrate the effectiveness of cross attention in latent space.\par
\begin{table}[htbp]
	\centering
	\caption{Ablation of Cross Attention layers on HumanML3D. GS represents guidance scale, which is a train-free hyperparameter. The number of network layers in our experimental design and GS is 9 and 11, respectively.}
	\resizebox{\columnwidth}{!}{
		\begin{tabular}{lccccccc}
			\toprule
			\multirow{2}{*}{Methods} & 
			R Precision & \multirow{2}{*}{FID↓} & \multirow{2}{*}{MM Dist↓} &
			\multirow{2}{*}{Diversity→} &
			\multirow{2}{*}{MModality↑} \\ 
			& {Top 1↑}&  \\ \hline
			\rule{0pt}{10pt}
			Real & 0.511$^{\pm .003}$ & 0.002$^{\pm .000}$ & 2.974$^{\pm .008}$ & 9.503$^{\pm .065}$ & - \\
			\hline
			w/o Attn. &  0.478$^{\pm.002}$ & 
			0.265$^{\pm .009}$ & 3.086$^{\pm .008}$ & 9.827$^{\pm .089}$ & 2.447$^{\pm .099}$\\
			Stack layer & 0.480$^{\pm .002}$ & 0.149$^{\pm .005}$ & 3.080$^{\pm .007}$ & 9.737$^{\pm .075}$ & 2.605$^{\pm .010}$\\
			Block & 0.486$^{\pm .003}$ & 0.122$^{\pm .005}$ & 3.044$^{\pm .011}$ & 9.670$^{\pm .070}$ & 2.532$^{\pm .095}$\\
			\hline
			Block (5 layers) & 0.438$^{\pm .003}$ & 0.242$^{\pm .006}$ & 3.279$^{\pm .008}$ & 9.726$^{\pm .088}$ &
			{\textcolor{red}{2.790}}$^{\pm .113}$\\
			Block (7 layers) & 0.471$^{\pm .002}$ & 0.153$^{\pm .006}$ & 3.119$^{\pm .007}$ & 9.687$^{\pm .087}$ & 2.674$^{\pm .098}$\\
			Block (9 layers) & 0.486$^{\pm .003}$ & 0.122$^{\pm .005}$ & 3.044$^{\pm .011}$ &  {\textcolor{red}{9.670}}$^{\pm .070}$ & 2.532$^{\pm .095}$\\
			Block (11 layers) & 0.492$^{\pm .002}$ & 0.113$^{\pm .005}$ & 3.034$^{\pm .006}$ & 9.785$^{\pm .085}$ &  2.566$^{\pm .095}$\\
			Block (13 layers) & 0.478$^{\pm .002}$ & 0.109$^{\pm .004}$ & 3.077$^{\pm .007}$ & 9.813$^{\pm .077}$ &  2.549$^{\pm .081}$\\
			\hline
			GS = 6.5 & 0.493$^{\pm .002}$ &  {\textcolor{red} {0.102}}$^{\pm .003}$ & {\textcolor{red}{2.973}}$^{\pm .006}$ & 9.749$^{\pm .082}$ & 2.614$^{\pm .100}$\\
			GS = 7 & 0.489$^{\pm .002}$ &  0.107$^{\pm .003}$ & 3.050$^{\pm .006}$ & 9.739$^{\pm .075}$ & 2.517$^{\pm .078}$\\
			GS = 7.5 & 0.492$^{\pm .002}$ & 0.113$^{\pm .005}$ & 3.034$^{\pm .006}$ & 9.785$^{\pm .085}$ &  2.566$^{\pm .095}$\\
			GS = 8 & 0.485$^{\pm .003}$ & 0.103$^{\pm .004}$ & 3.064$^{\pm .007}$ & 9.726$^{\pm .073}$ & 2.554$^{\pm .080}$\\
			GS = 8.5 & {\textcolor{red} {0.497}}$^{\pm .003}$ & 0.121$^{\pm .005}$ & 3.017$^{\pm .007}$ & 9.795$^{\pm .090}$ & 2.503$^{\pm .093}$\\
			\hline
		\end{tabular}
	}
	\label{tab:Ablation_Cross_Attention}
\end{table}
It is worth noting that our experimental results are inconsistent with, or even contrary to, the results of MLD \cite{mld}. MLD believes that the insufficient performance of VAE-7 is due to insufficient data. However, we proposed that the stronger Motion VAE requires a more powerful latent diffusion model to achieve better text alignment. This is achieved through cross attention and the addition of  layers. Therefore, cross attention will perform worse in an insufficient latent space (VAE-1), and VAE-7 will perform worse in the absence of a more powerful diffusion model.\par 
\subsection{Reasoning motion generation Experiment}
We conducted a visualization experiment, as shown in Fig. \ref{fig:VisualizationLLMPlannerResults}. It can be seen that through the LLM Planner, unseen zero-shot texts can be converted into dataset descriptions, thereby facilitating user-friendly motion generation.\par

{\bf LLM Planner Ablation.} In Tab. \ref{tab:Ablation_LLMPlanner}, we conduct experiments directly using prompt engineering on the Llama, Qwen, and Deepseek models. In our experiments, we found that Deepseek-r1 performs best, effectively enabling the diffusion model to understand dataset-specific  text and complete inference generation.   However, its response time for each instruction is slow due to the reasoning chain process.   Meanwhile, other current large models can also perform this task effectively. We recommend using Deepseek-V3 or Qwen2.5-max as an LLM Planner for tradeoff.\par
\label{sec:llmAblation}
\begin{figure*}[htbp]
	\centering
	\begin{tabularx}{\textwidth}{>{\centering\arraybackslash}m{2.2cm}
			>{\centering\arraybackslash}X|
			>{\centering\arraybackslash}X
			>{\centering\arraybackslash}X
			>{\centering\arraybackslash}X
			>{\centering\arraybackslash}X}
		\toprule
		\multirow{2}{*}{Text} & Motion Prompt ({\it 'Teaport'}) & \multirow{2}{*}{ADM} & w/ finetuning on 100STYLE & Motion Prompt ({\it 'Old'} person) & \multirow{2}{*}{Trajectory}\\
		{\it a person walks forward slowly in a straight line} & 
		\includegraphics[width=0.9\linewidth,valign=c]{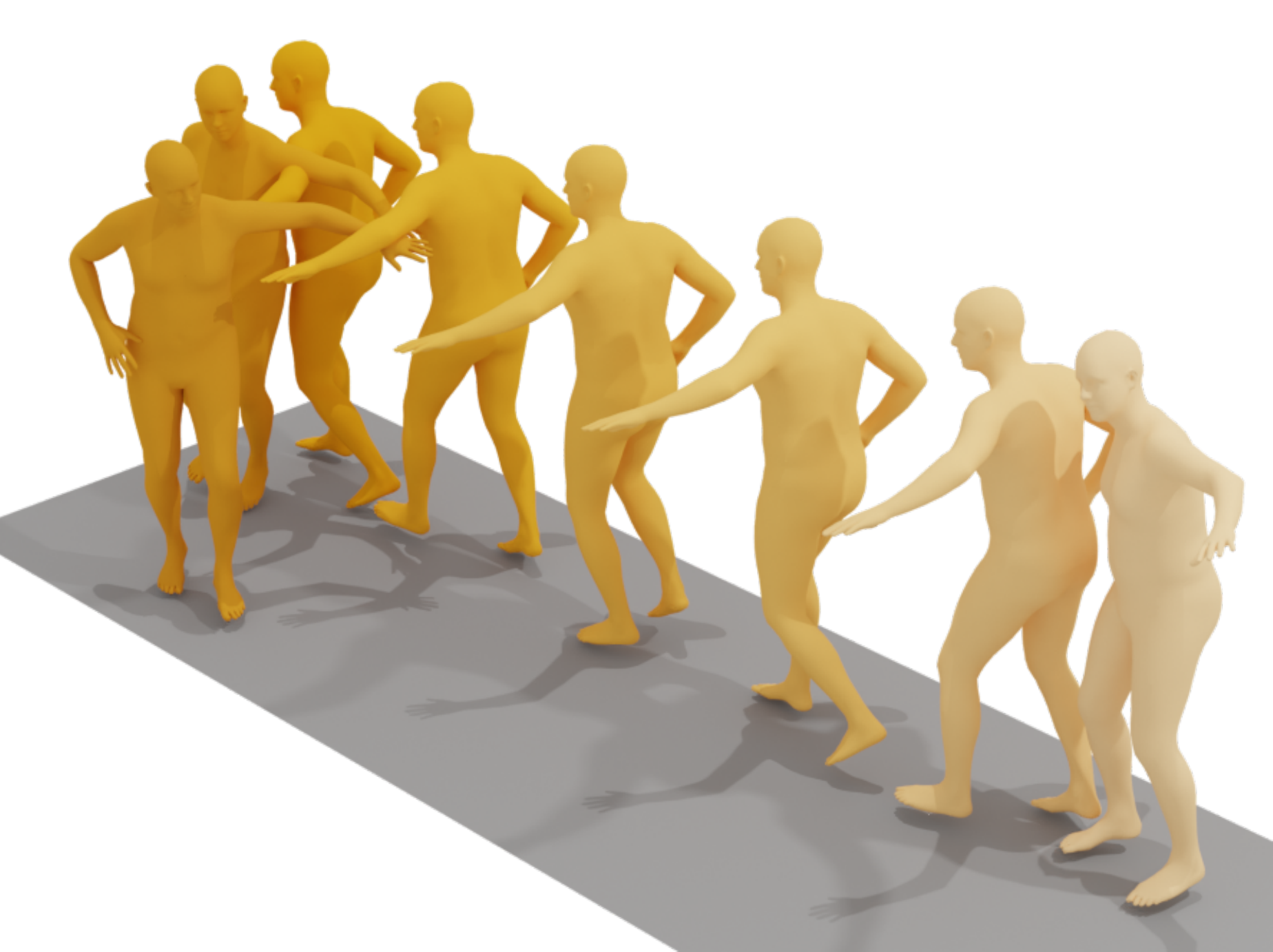}
		&
		\multicolumn{2}{c}{\it 'a person walks wobbly down an incline'} & \includegraphics[width=0.9\linewidth,valign=c]{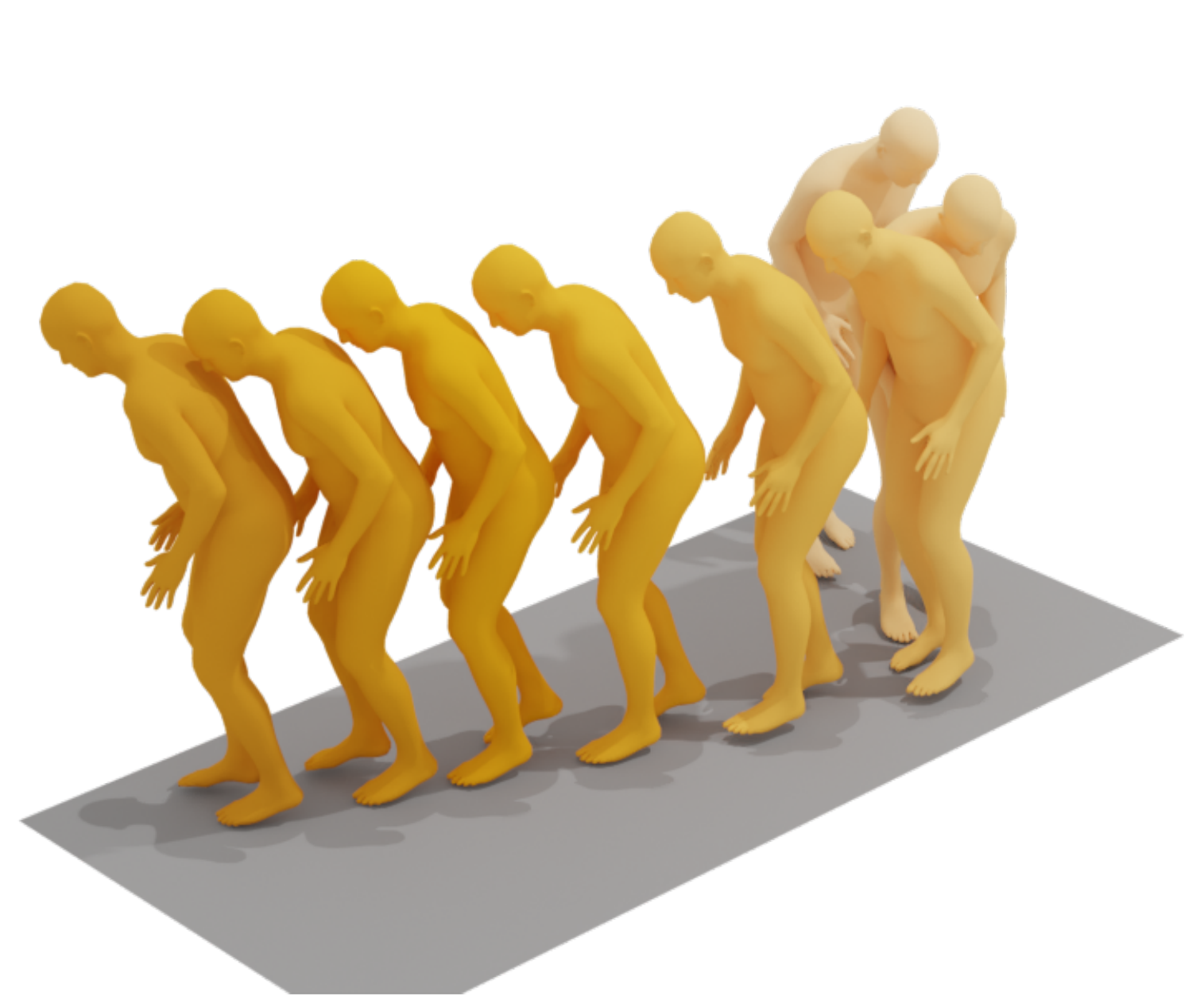} &
		\includegraphics[width=0.9\linewidth,valign=c]{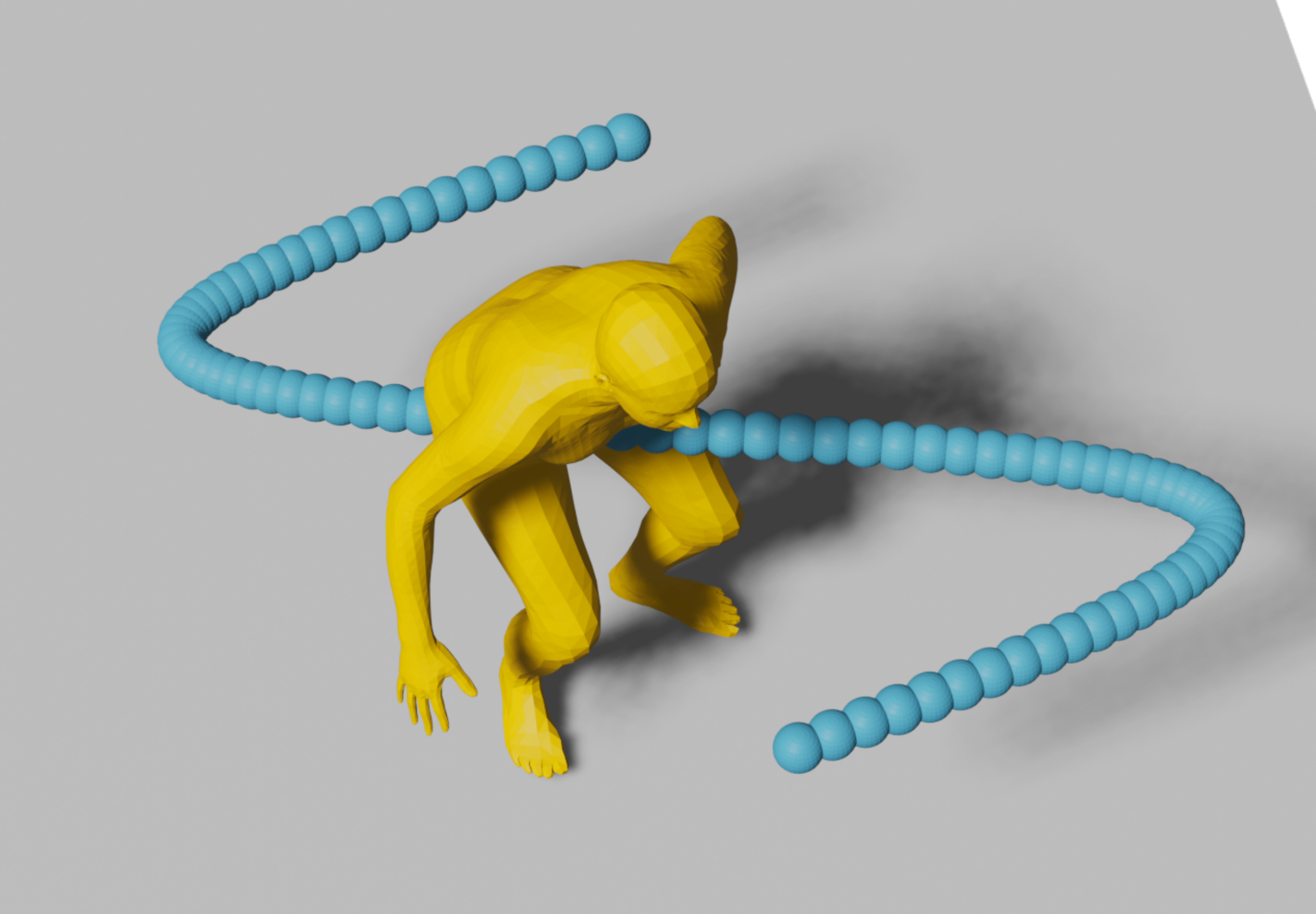} \\
		\cline{3-6}
		{\it while holding something heavy on their shoulder} &
		\includegraphics[width=0.9\linewidth,valign=c]{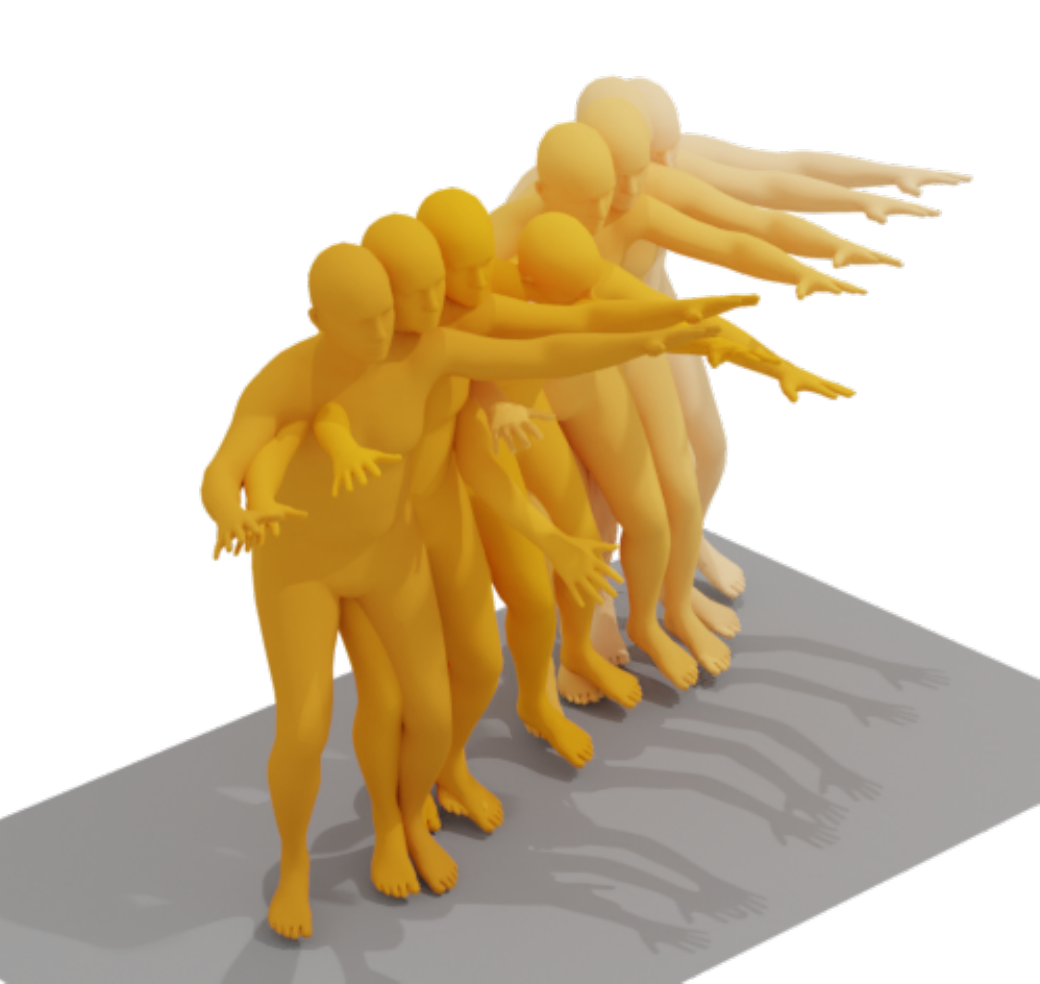} &
		\includegraphics[width=0.9\linewidth,valign=c]{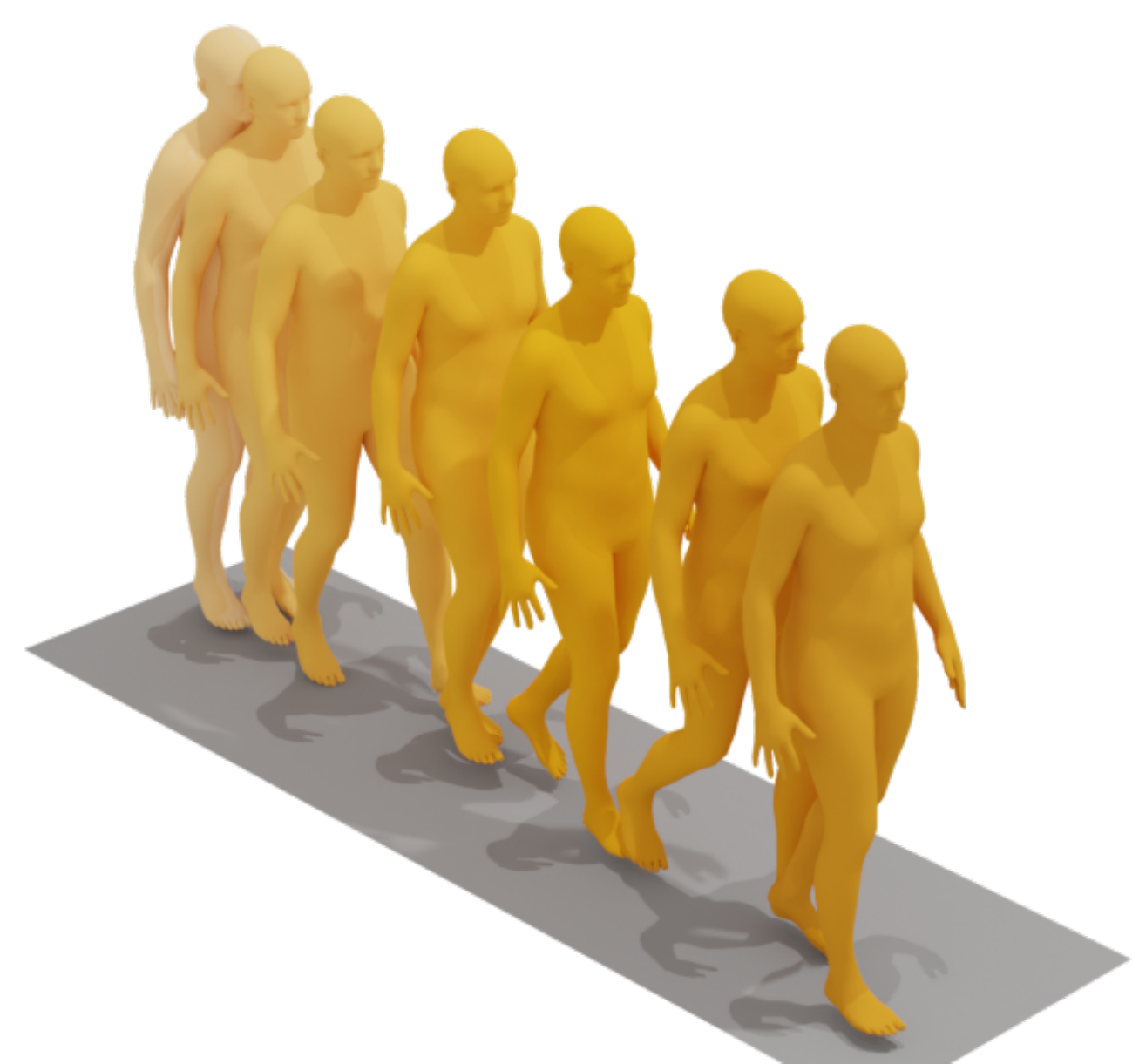}
		&
		\includegraphics[width=0.9\linewidth,valign=c]{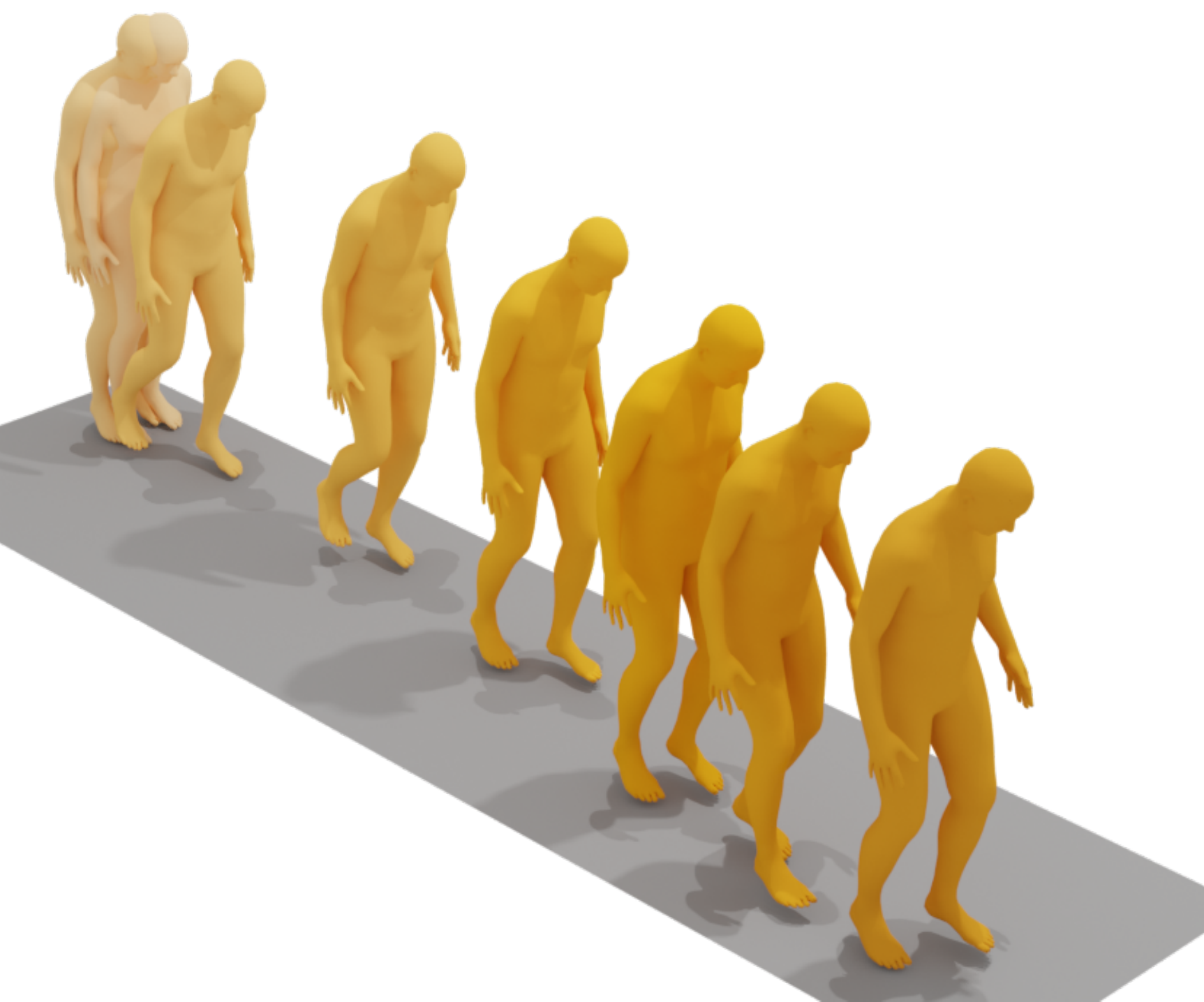}
		&
		\includegraphics[width=0.95\linewidth,valign=c]{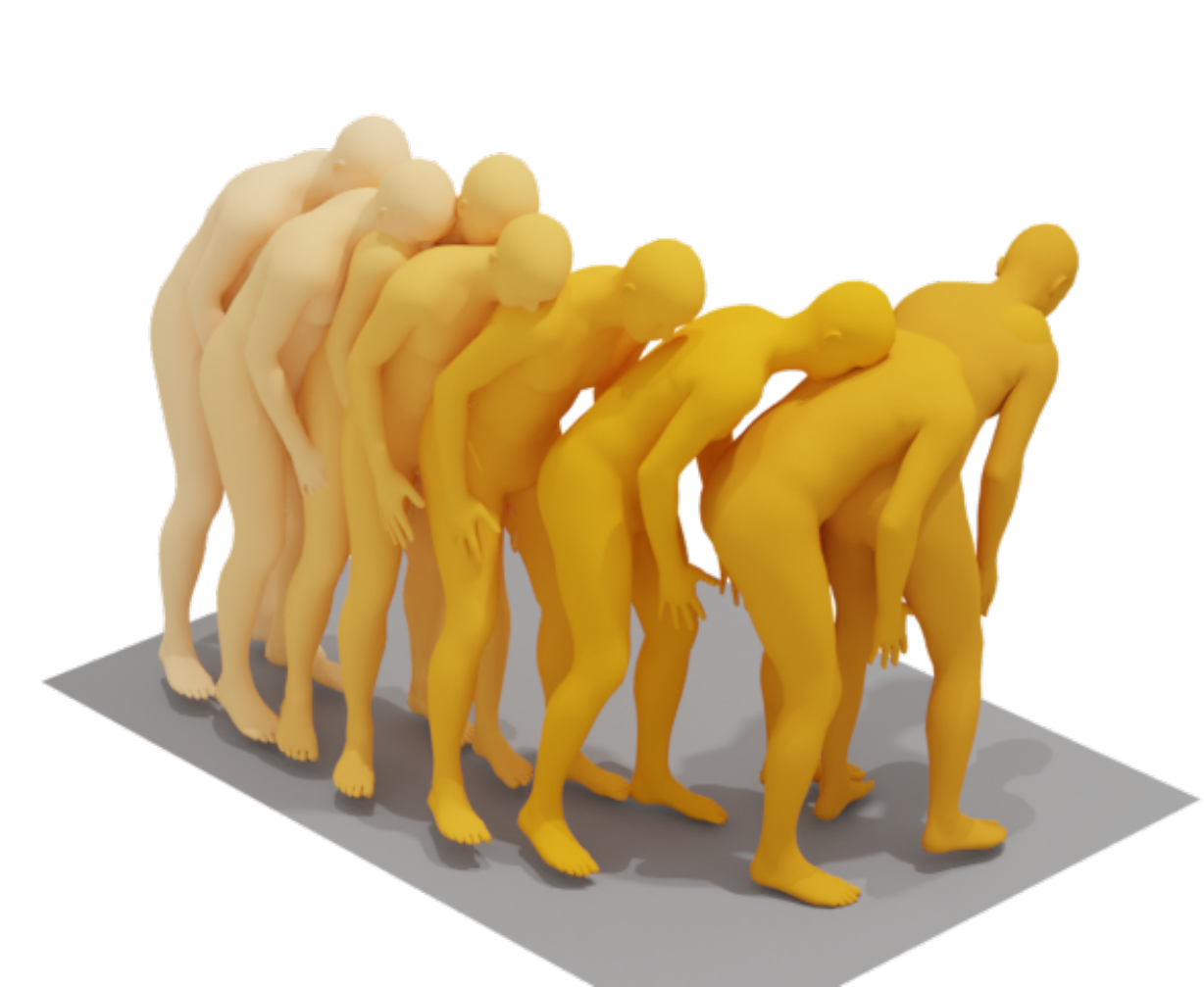}
		& \includegraphics[width=0.8\linewidth,valign=c]{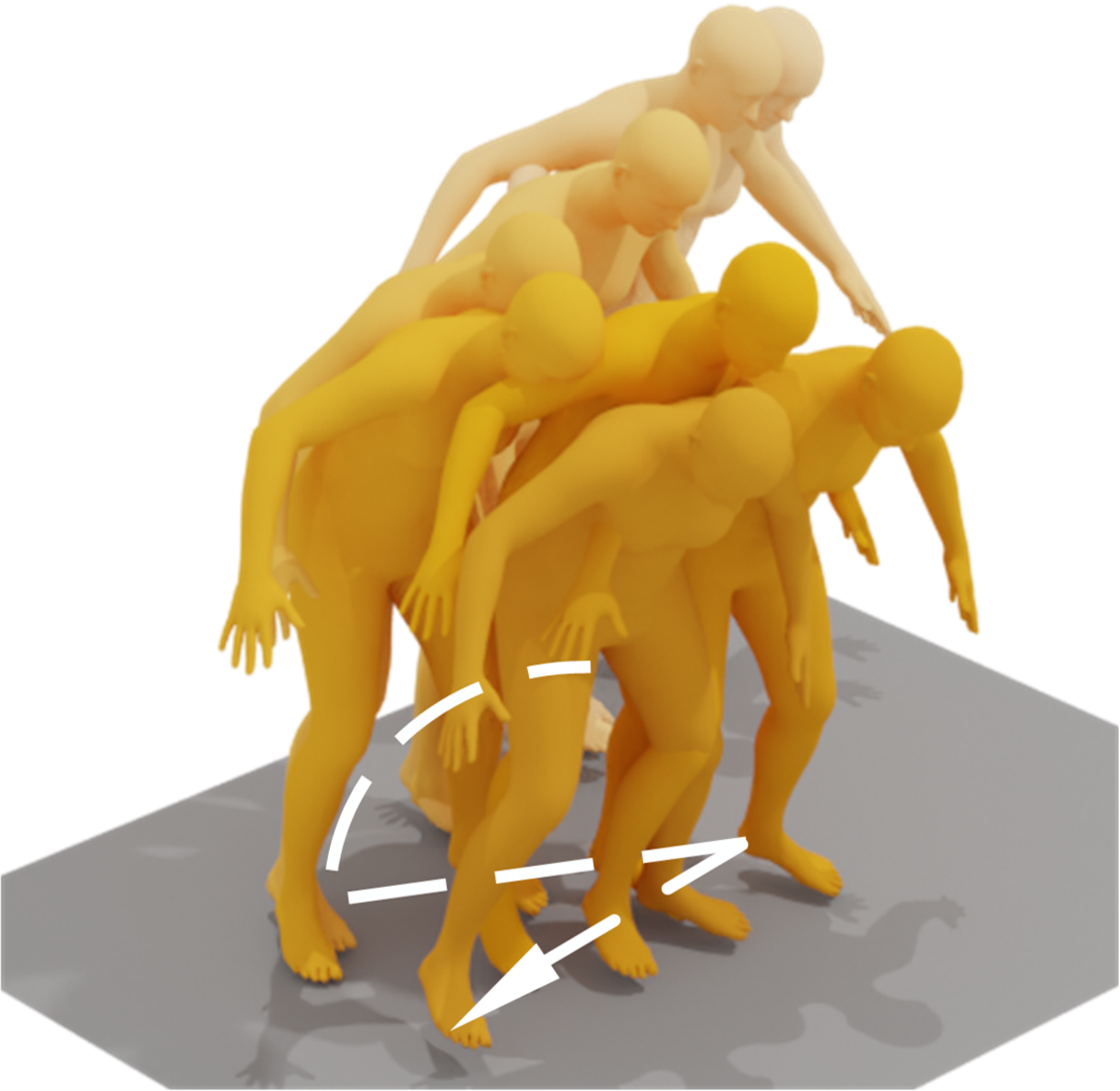}\\
		\bottomrule
	\end{tabularx}
	\caption{Visual ablation experiment of motion prompt and trajectory. The proposed Motion Adapter facilitates stylized generation by motion prompts to specific semantics, as shown on the left. On the right is a visualization example of multiple attributes without enhanced text desciption, we finetune then add a motion prompt, and subsequently add a track. More visualizations in supplemental material.}
	\label{fig:VisualizationMotionPromptTraj}
\end{figure*}

\begin{table}[htbp]
	\centering
	\caption{Ablation of LLM Planner on MotionIT. We used original instruction in MotionIT as ground truth for BLEU, ROUGE, and CIDEr. A DeepSeek-R1-Distill-Qwen-32B acts as a judge to calculate the PS and RCS. We take samples and repeate the tests on the samples three times.}
	\resizebox{\columnwidth}{!}{
		\begin{tabular}{l | ccc | ccc }
			\toprule
			Method & BLEU@4↑ & ROUGE↑ & CIDEr↑ & {\bf PS↓} & {\bf RCS(\%)↑} & {\bf ART↓} \\
			\hline
			Deepseek-R1  & 0.04 & 0.27 & 0.13 & 3.37$^{\pm .04}$ &  {\textcolor{red}{75}}$^{\pm 2}$ &63.04s \\
			Deepseek-V3  & 0.08 & 0.44 & 0.62 & 2.16$^{\pm .04}$& 59$^{\pm 1}$ & 2.49s\\
			Qwen2.5-max  & 0.14 & 0.52 & 1.21 & {\textcolor{red}{1.55}}$^{\pm .04}$ & 55$^{\pm 1}$ & 856ms\\
			llama-3.2-3B & 0.04 & 0.27 & 0.17 & 3.42$^{\pm .06}$ & 40$^{\pm 1}$ & 13.96ms\\
			llama-3.2-1B & 0.06 & 0.32 & 0.33 & 4.50$^{\pm .03}$ & 37$^{\pm 1}$ &  {\textcolor{red} {13.57ms}}\\
			\hline
		\end{tabular}
	}
	
	\label{tab:Ablation_LLMPlanner}
\end{table}

\subsection{Motion Style and Trajectory Experiment}
\label{sec:MotionStyleAndTraj}
To ensure a fair comparison and eliminate the influence of data factors, we used the model without the LLM Planner for attribute experiments. Except for Fig. \ref{fig:ACMo_idea}, additional visualization examples are provided in Fig. \ref{fig:VisualizationMotionPromptTraj}. This shows the effect of the proposed multiple attribute controls. We use the {\it motion prompt} to specify the motion we wish, allowing the model to understand the motion to be generated under the text category.\par
\begin{table}[htbp]
	\centering
	\caption{Comparison of text-to-motion generation on 100STYLE via finetuning. {\it Param.} represents the sum of the number of frozen parameters and the number of {\it trainable} parameters. Text data annotated via MotionGPT \cite{motiongpt} lacks adequate 'Real' motion description. The FID metric measures the difference in data distribution between the generated results and 100STYLE motions, which is important. Finetuning training within 100 epochs.}
	\resizebox{\columnwidth}{!}{
		\begin{tabular}{clccccc}
			\toprule
			\multirow{2}{*}{Dataset} & 
			\multirow{2}{*}{Methods} & 
			R Precision & \multirow{2}{*}{\bf FID↓} & \multirow{2}{*}{MM Dist↓} &
			\multirow{2}{*}{Diversity→} &
			\multirow{2}{*}{Param. (M)} \\ 
			& & {Top 3↑}&  \\ \midrule
			\multirow{3}{*}{w/o finetune} & Real (100STYLE) & 0.317$^{\pm .004}$ & - & 5.687$^{\pm .026}$ & 6.492$^{\pm .066}$ & - \\
			\cline{2-7}
			& MLD & 0.687$^{\pm .007}$ & 5.956$^{\pm .114}$ & 3.610$^{\pm .029}$ & 9.102$^{\pm .071}$ & 8.3 \\
			& ADM & 0.766$^{\pm .006}$ & 5.935$^{\pm .109}$ & 3.109$^{\pm .019}$ & 8.960$^{\pm .066}$ & 20.3 \\
			\midrule
			\multirow{9}{*}{100STYLE} & Smoodi \cite{smoodi} & 0.571 & 1.609 & 4.477 & 9.235 & 8.3{\it+19.2} \\
			& ADM (Full Model) & 0.539$^{\pm .008}$ & {\textcolor{red}{1.223}}$^{\pm .055}$ & 4.527$^{\pm .028}$  & 7.468$^{\pm .083}$ & {\it+20.3} \\
			& Motion Adapter & {\textcolor{red}{0.597}}$^{\pm .007}$ & 1.271$^{\pm .055}$ & {\textcolor{red}{4.217}}$^{\pm .024}$ & {\textcolor{red}{7.361}}$^{\pm .060}$ & 10.6{\it+19.2} \\
			\cline{2-7} \noalign{\vskip 3pt} 
			\cline{2-7} \noalign{\vskip 3pt} 
			& Add on Self Attn. & 0.591$^{\pm .009}$ & 2.095$^{\pm .124}$ & 4.099$^{\pm .033}$ & 7.830$^{\pm .088}$ & 20.1{\it+ 9.7}\\
			& Only train Self Attn. & 0.597$^{\pm .008}$ & 1.800$^{\pm .066}$ & 4.120$^{\pm .030}$ & 7.897$^{\pm .077}$ & 11{\it+ 9.7}\\
			& (epoch=1000) & 0.487$^{\pm .007}$ & 0.887$^{\pm .066}$ & 4.841$^{\pm .036}$ & 7.205$^{\pm .069}$ & 11{\it+ 9.7}\\
			& + Add on Text Attn. & 0.576$^{\pm .007}$ & 1.486$^{\pm .060}$ & 4.264$^{\pm .035}$ & 7.679$^{\pm .094}$ & 10.6{\it+19.2} \\
			& + Self-to-Cross Attn. & 0.597$^{\pm .007}$ & 1.271$^{\pm .055}$ & 4.217$^{\pm .024}$ & 7.361$^{\pm .060}$ & 10.6{\it+19.2} \\
			\bottomrule
		\end{tabular}
	}
	\label{tab:Comparsion_motion_Adapter}
\end{table}
{\bf Motion Adapter Ablation.} 
We attempted to achieve fine-tuning with new motion patterns, as shown in Tab. \ref{tab:Comparsion_motion_Adapter}. Overfitting is prevented within 100 epochs. The FID of the ground truth compared to itself is less than 1e-9 and can be considered negligible.
Firstly, since our text auto-annotations come from MotionGPT 
\cite{motiongpt}, this is not a high-quality labeled and may cause confusion with previous data, thus reducing R Precision. Secondly, we hope to learn new motion patterns while retain the original training knowledge. So, it is obviously that our method reduces the value of FID while retaining the R precision. This is a significant advantage of our decoupling training, our method don't forget the original text memory. At the same time, our method has very few training parameters and fast training.  In contrast, the parameters of ControlNet in Smoodi relies on the denoiser of replicated pretrain model. Despite the ADM having more parameters than MLD, the motion adapter requires fine-tuning only a minimal set of parameters to achieve superior performance, highlighting its efficiency. Our method solves three issues: 1) MLD’s weak text-to-motion ability, 2) the 10x longer training time compared to MLD in Smoodi \cite{smoodi} and 3) maintain knowledge then learn new patterns. Finally, the proposed approach enables rapid fine-tuning to allow the model to 'see' new motion patterns.\par

%% file: sec/5_conclusion.tex
\section{Discussion and Conclusion}
\indent \indent {\bf Discussion.}
Generating motion trajectories from text and fine-tuning new actions via video will be our future work. Meanwhile, existing metrics are inadequate, such as the absence of quantitative evaluation for multi-attribute tasks and cases where R Precision exceeds real motion data. Finally, we believe that the proposed motion adapter and decoupling any conditions can provide insights for the generation of images, videos, and 3D models. \par
{\bf Limitations.} We acknowledge certain limitations in ACMo. The text-to-motion latent diffusion model offers ease of control but is less precise than state-of-the-art. Additionally, while the proposed motion adapter is lightweight, our pipeline increases model size and is slower than other compact models. Additional multi-attribute limitations and bad cases are discussed in the supplementary materials.\par
{\bf Conclusion.} We present {\it ACMo}, an effective architecture for multimodal, multi-attribute controllable motion generation. It uses the {\it Attribute Diffusion Model} for text-to-motion generation, introduces the {\it Motion Adapter} for stylized motion prompts, and integrates trajectory control. The introduced {\it LLM Planner} can map unseen attributes into dataset text. Extensive experiments and ablation studies across multiple datasets validate their effectiveness.